\documentclass[acmlarge]{acmart}
\usepackage{latexsym}
\usepackage{url}
\usepackage{amsmath}
\usepackage{amssymb}
\usepackage{hyperref}

\usepackage{graphicx}

\newcommand{\REVISE}[1]{#1}

\newcommand{\compactpara}[1]{\noindent\textbf{#1}}

\newcommand{\vs}{\textit{vs.}}
\newcommand{\eg}{\textit{e.g.}}

\newcommand{\ftlink}[1]{}

\newcommand\BibTeX{B{\sc ib}\TeX}

\AtBeginDocument{%
  \providecommand\BibTeX{{%
    \normalfont B\kern-0.5em{\scshape i\kern-0.25em b}\kern-0.8em\TeX}}}

\setcopyright{iw3c2w3g}
\copyrightyear{2018}
\acmYear{2020}
\acmDOI{10.1145/1122445.1122456}

\acmJournal{CSUR}
\acmVolume{37}
\acmNumber{4}
\acmArticle{111}
\acmMonth{8}

\begin{document}

\title{A Comprehensive Survey of Multilingual Neural Machine Translation}

\author{Raj Dabre}
\authornote{All authors contributed equally to this research.}
\email{prajdabre@gmail.com}
\orcid{1234-5678-9012}
\affiliation{%
  \institution{National Institute of Information and Communications Technology (NICT), Kyoto, Japan}
}
\author{Chenhui Chu}
\authornotemark[1]
\email{chu@ids.osaka-u.ac.jp}
\affiliation{%
  \institution{Osaka University, Osaka, Japan}
}

\author{Anoop Kunchukuttan}
\authornotemark[1]
\affiliation{%
  \institution{Microsoft AI and Research, Hyderabad, India}
  }
\email{anoop.kunchukuttan@gmail.com}

\begin{abstract}
  We present a survey on multilingual neural machine translation (MNMT), which has gained a lot of traction in the recent years. MNMT has been useful in improving translation quality as a result of translation knowledge transfer \REVISE{(transfer learning)}. MNMT is more promising and interesting than its statistical machine translation counterpart because end-to-end modeling and distributed representations open new avenues for research on machine translation. Many approaches have been proposed in order to exploit multilingual parallel corpora for improving translation quality. However, the lack of a comprehensive survey makes it difficult to determine which approaches are promising and hence deserve further exploration. In this paper, we present an in-depth survey of existing literature on MNMT. We first categorize various approaches based on their central use-case and then further categorize them based on resource scenarios, underlying modeling principles, \REVISE{core-issues and challenges}. Wherever possible we address the strengths and weaknesses of several techniques by comparing them with each other. We also discuss the future directions that MNMT research might take. This paper is aimed towards both, beginners and experts in NMT. We hope this paper will serve as a starting point as well as a source of new ideas for researchers and engineers interested in MNMT.
\end{abstract}






\maketitle


\section{Introduction}

Neural machine translation (NMT) 
\cite{DBLP:journals/corr/ChoMBB14,DBLP:journals/corr/SutskeverVL14,bahdanau15} has become the dominant paradigm for MT in academic research as well as commercial use \cite{DBLP:journals/corr/WuSCLNMKCGMKSJL16}. NMT has shown state-of-the-art performance for many language pairs \cite{bojar-EtAl:2017:WMT1,bojar2018wmtfindings}. Its success can be mainly attributed to the use of distributed representations of language, enabling end-to-end training of \REVISE{an} MT system. Unlike \REVISE{classical} statistical machine translation (SMT) systems \cite{koehn-EtAl:2007:PosterDemo}, separate lossy components like word aligners, translation rule extractors and other feature extractors are not required. 
The dominant NMT approach is the \textit{Embed - Encode - Attend - Decode} paradigm. Recurrent neural network (RNN) \cite{bahdanau15}, convolutional neural network (CNN) \cite{gehring17} and self-attention/feed-forward network (SA/FFN) \cite{NIPS2017_7181} architectures are popular approaches based on this paradigm. For a more detailed exposition of NMT, we refer readers to some prominent tutorials \cite{neubig17nmt,koehn17nmt}.


While initial research on NMT started with building translation systems between two languages, researchers discovered that the NMT framework can naturally incorporate multiple languages. Hence, there has been a massive increase in work on MT systems that involve more than two languages \cite{dong15,firat16,N16-1004,ijcai2017-555,johnson17,P17-1176,DBLP:conf/aaai/ChenLL18,D18-1103}. We refer to NMT systems handling translation between more than one language pair as \textit{multilingual NMT} (MNMT) systems. The ultimate goal of MNMT research is to develop one model for translation between as many languages as possible by effective use of available linguistic resources.

\begin{figure*}
    \centering
    \includegraphics[width=\textwidth]{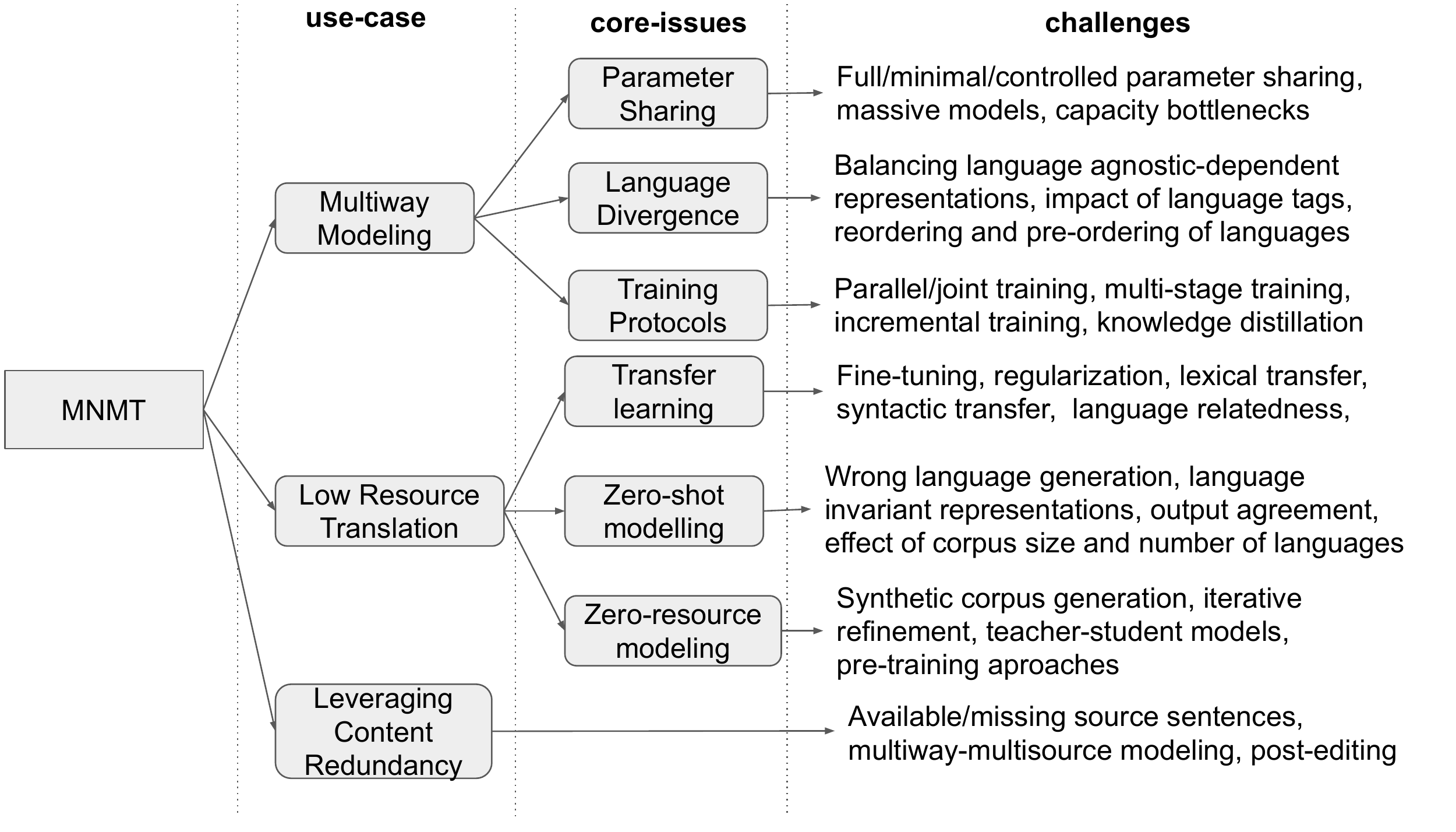}
        \caption{\REVISE{MNMT research categorized according to use-cases, core-issues and the challenges involved. Note that the focus is on use-cases and an approach for one use-case can be naturally adapted to another use-case. Note that the lower two use-cases can face similar core issues as the the uppermost use-case.}}
        \label{fig:overview}
\end{figure*}

MNMT systems are desirable because training models with data from many language pairs might help a resource-poor language acquire extra knowledge; from the other languages \cite{N16-1004}. Moreover, MNMT systems tend to generalize\footnote{\REVISE{Using additional languages can help in word sense disambiguation which can help improve translation quality.}} better due to exposure to diverse languages, leading to improved translation quality compared to bilingual NMT systems. This particular phenomenon is known as translation knowledge transfer \cite{Pan:2010:STL:1850483.1850545}. For the rest of the paper we will use the terms ``knowledge transfer,'' ``transfer learning'' and ``translation knowledge transfer'' interchangeably. 
Knowledge transfer has been strongly observed for translation between low-resource languages, which have scarce parallel corpora or other linguistic resources but have benefited from data in other languages \cite{DBLP:conf/emnlp/ZophYMK16:original}. Knowledge transfer also has been observed between high-resource languages, where MNMT systems outperform bilingual NMT systems \cite{johnson17}. Multilingual training is also known to be a source of regularization during training which further improves generalization. In addition, MNMT systems have the potential to be relatively compact,\footnote{Relatively modest increase in model parameters despite an increase in the number of languages as a result of parameter sharing.} because a single model handles translations between multiple language pairs \cite{johnson17}. This can reduce the deployment footprint, which is crucial for constrained environments like mobile phones or IoT devices. It can also simplify the large-scale deployment of MT systems. 

There are multiple scenarios where MNMT has been put to use based on available resources and use-cases. The following are the major scenarios where MNMT has been explored in the literature. (See Figure \ref{fig:overview} for an overview): 

\vspace{2 mm}
\compactpara{Multiway Translation.} The goal is constructing a single NMT system for one-to-many \cite{dong15}, many-to-one \cite{lee17} or many-to-many \cite{firat16} translation using parallel corpora for more than one language pair. \REVISE{In this scenario we make a very generic assumption that parallel corpora for a number of languages is available. Here, the ultimate objective is to incorporate a number of languages into a single model.}
\vspace{2 mm}

\compactpara{Low Resource Translation.} \REVISE{Little to no parallel corpora exist for most language pairs in the world. Multiple studies have explored using \textit{assisting} languages to improve translation between low-resource language pairs. These multilingual NMT approaches for low-resource MT address two broad scenarios: (a) a high-resource language pair (\eg Spanish-English) is available to assist a low-resource language pair (\eg Catalan-English). Transfer learning is typically used in this scenario \cite{DBLP:conf/emnlp/ZophYMK16:original}.  (b) no direct parallel corpus for the low-resource pair, but languages share a parallel corpus with one or more \textit{pivot} language(s). }

\vspace{2 mm}

\compactpara{Multi-Source Translation.} Documents that have been translated into more than one language might, in the future, be required to be translated into another language. In this scenario, existing multilingual redundancy on the source side can be exploited for multi-source translation \cite{N16-1004}. \REVISE{Multilingual redundancy can help in better disambiguation of content to be translated, leading to an improvement in translation quality.}

\vspace{2 mm}
\REVISE{We believe that the biggest benefit of doing MNMT research by exploring these scenarios is that we might gain insights into and an answer to an important question in natural language processing: }
\begin{description}
\item[Q.] \REVISE{\emph{How can we leverage multilingual data effectively in order to learn distributions across multiple languages so as to improve MT (NLP) performance across all languages?}
}
\end{description}

\REVISE{This question can be decomposed into a number of smaller questions and in this paper we try to answer two of them as follows:}
\begin{description}
\item[Q1.] \REVISE{\emph{Is it possible to have a one-model-for-all-languages solution to MT (NLP) applications?}}
\item[Q2.] \REVISE{\emph{Can shared multilingual distributed representations help MT (NLP) for low-resource languages?}}
\end{description}

\vspace{2 mm}
Given these benefits, scenarios and the tremendous increase in the work on MNMT in recent years, we write this survey paper on MNMT to systematically organize the work in this area. To the best of our knowledge, no such comprehensive survey on MNMT exists. Our goal is to shed light on various MNMT scenarios, fundamental questions in MNMT, basic principles, architectures, and datasets for MNMT systems.
The remainder of this paper is structured as follows:  
We present a systematic categorization of different approaches to MNMT in each of the above mentioned scenarios to help understand the array of design choices available while building MNMT systems (Sections \ref{sec:multiway}, \ref{sec:low}, \ref{sec:unseen}, and \ref{sec:multisource}). \REVISE{The top-level organization of the survey is use-case scenario based: multiway NMT (Section \ref{sec:multiway}), low-resource NMT  (Sections \ref{sec:low}  and \ref{sec:unseen}) and multisource NMT (Section \ref{sec:multisource}). Although zero-shot/zero-resource is a special case of low-resource NMT, we dedicate a separate section (\ref{sec:unseen}) given its growing importance and interest. For each scenario, we address the challenges, considerations and multilingual NMT-based solutions. Note that a model or technique proposed for one scenario may be used in another scenario. We have mentioned proposed solutions in the context of the scenario in which they have been discussed in the literature. Their applicability to other scenarios may be subject to further investigation.} 
We put the work in MNMT into a historical perspective with respect to multilingual MT in older MT paradigms (Section \ref{sec:history}). We also describe popular multilingual datasets and the shared tasks that focus on multilingual NMT (Section \ref{sec:data}). In addition, we compare MNMT with domain adaptation for NMT, which tackles the problem of improving low-resource in-domain translation (Section \ref{sec:domain}).
Finally, we share our opinions on future research directions in MNMT (Section \ref{sec:future}) and conclude this paper (Section \ref{sec:conlusion}).

\section{Neural Machine Translation}
\label{sec:nmt}
 \begin{figure}[t]
          \includegraphics[width=0.75\hsize]{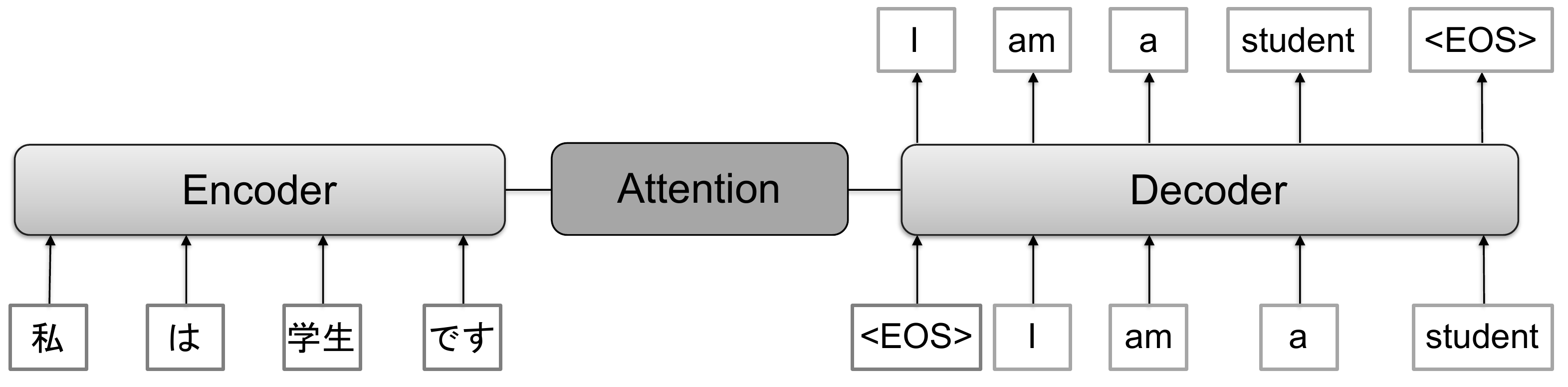}
     \caption{\label{fig:rnnsearch} \REVISE{A standard NMT model based on the encode-attend-decode modeling approach.}}
 \end{figure}
\REVISE{Given a parallel corpus $\mathbf{C}$ consisting of a set of parallel sentence pairs $(\mathbf{x, y})$, the training objective for NMT is maximize the log-likelihood $\mathcal{L}$ w.r.t $\mathbf{\theta}$:}
\begin{equation}
\label{loss}
\REVISE{\mathcal{L}_\theta = \sum_{(\mathbf{x,y})\in\mathbf{C}}\log p(\mathbf{y}|\mathbf{x}; \mathbf{\theta}).}
\end{equation}
\REVISE{where $\mathbf{x} = \{x_1, ..., x_n\}$ is an input sentence, $\mathbf{y} = \{y_1, ..., y_m\}$ is its translation, and $\mathbf{\theta}$ is a set of parameters to be learned. The probability of a target sentence given the source sentence is:}
\begin{equation*}
\REVISE{p(\mathbf{y}|\mathbf{x};\mathbf{\theta}) = \prod_{j=1}^{m} p(y_j|y_{<j}, \mathbf{x};\mathbf{\theta}),}
\end{equation*}
\REVISE{where $m$ is the number of words in $\mathbf{y}$, $y_j$ is the current generated word, and $y_{<j}$ are the previously generated words. At inference time, beam search is typically used to find the translation that maximizes the above probability.} 

\REVISE{The most commonly used NMT approach is the \textit{Embed - Encode - Attend - Decode} paradigm.
Figure~\ref{fig:rnnsearch} shows an overview of this paradigm. The encoder first converts words in the source sentence into word embeddings. These word embeddings are then processed by neural layers and converted to representations that capture contextual  information about these words. We call these contextual representations as the \textit{encoder representations}. 
The decoder uses an attention mechanism, the encoder representations, and previously generated words to generate what we call the decoder representations (states) which in turn are used to generate the next target word. They encoder and decoder can be RNN \cite{bahdanau15}, CNN \cite{gehring17} or self-attention and feed-forward \cite{NIPS2017_7181} layers. Among these, the self-attention layers are the most widely used. It is a common practice to stack multiple layers which leads to an improvement in translation quality. The attention mechanism is calculated cross the decoder and encoder as:}
\begin{equation}
\label{attention_score}
\REVISE{e_{ji}=a(\mathbf{s_{j-1}}, \mathbf{h_i}),}
\end{equation}

\begin{equation*}
\REVISE{a_{ji}=\frac{\mathrm{exp}(e_{ji})}{\sum_{k=1}^{m}\mathrm{exp}(e_{ki})}, }
\end{equation*}
\REVISE{where $e_{ji}$ is an alignment score, $a$ is an alignment model that scores the match level of the inputs around position $i$ and the output at position $j$, $\mathbf{s_{j-1}}$ is the decoder hidden state of the previous generated word, $\mathbf{h_i}$ is the encoder hidden state at position $i$.
The calculated attention vector is then used to weight the encoder hidden states to obtain a context vector as:}
\begin{eqnarray*}
\REVISE{\mathbf{c_j}=\sum_{i=1}^{n}{a_{ji}}\mathbf{h_i}}
\end{eqnarray*}
\REVISE{This context vector, is fed to the decoder along with the previously generated word and its hidden state to produce a representation for generating the current word. 
An decoder hidden state for the current word $\mathbf{s_j}$ is computed by:}
\begin{equation*}
\REVISE{\mathbf{s_j}=g(\mathbf{s_{j-1}}, \mathbf{y_{j-1}}, \mathbf{c_j}),}
\end{equation*}
\REVISE{where $g$ is an activation decoder function, $\mathbf{s_{j-1}}$ is the previous decoder hidden state, $\mathbf{y_{j-1}}$ is the embedding of the previous word.
The current decoder hidden state $\mathbf{s_j}$, the previous word embedding and the context vector are fed 
to a feedforward layer $f$ and a softmax layer to compute a score for generating a target word as output:}
\begin{eqnarray*}
\REVISE{P(y_j|y_{<j}, \mathbf{x}) = {\rm softmax} (f(\mathbf{s_j}, \mathbf{y_{j-1}}, \mathbf{c_j})).}
\end{eqnarray*}

\vspace{2 mm}

\noindent \REVISE{ \compactpara{Training NMT models.} The parallel corpus that is used to train the NMT model is first subjected to pre-processing where it is sufficiently cleaned to remove noisy training examples. A vocabulary of the $N$ most frequent words is then created and the remaining words are treated as unknown words mapped to a single token designated by ``UNK.'' To overcome the problem of unknown words, the most common practice involves subword tokenization using methods such as byte-pair encoding (BPE) \cite{DBLP:journals/corr/SennrichHB15}, word-piece model (WPM) \cite{conf/icassp/SchusterN12} or sentence-piece model (SPM) \cite{kudo-richardson-2018-sentencepiece}. This enables the use of an open vocabulary. In order to train a NMT model, we typically minimize the cross-entropy\footnote{This also implies maximizing the likelihood of the predicted sequence of words in the target language.} (loss) between the predicted target words and the actual target words in the reference. This loss minimization is an optimization problem and gradient descent methods \cite{DBLP:journals/corr/Ruder16} such as SGD, ADAM, ADAGRAD and so on can be used. ADAM is popular in MT due to its ability to quickly train models but suffers from the inability to sufficiently converge. On the other hand, SGD is known to converge better but requires long training times. It is common practice to design a learning schedule that combines several optimizers in order to train a model with with high performance. Training is either done for a large number of iterations or till the model converges sufficiently. Typically, a model is said to converge when its evaluation on a development set does not change by a significant amount over several iterations. We refer researchers to previous works on NMT regarding this topic. Another consideration during training is the tuning of hyperparameters such as learning rate, hidden dimension size, number of layers and so on. Researchers often train a wide variety of models, an approach known as grid search, and choose a model with the best performance. This model is then used for translation. People who are familiar with deep learning might notice that this basic model can be naturally extended to involve multiple language pairs.}

\section{Multiway NMT}
\label{sec:multiway}

\REVISE{
\begin{figure*}
    \centering
    \includegraphics[width=\textwidth]{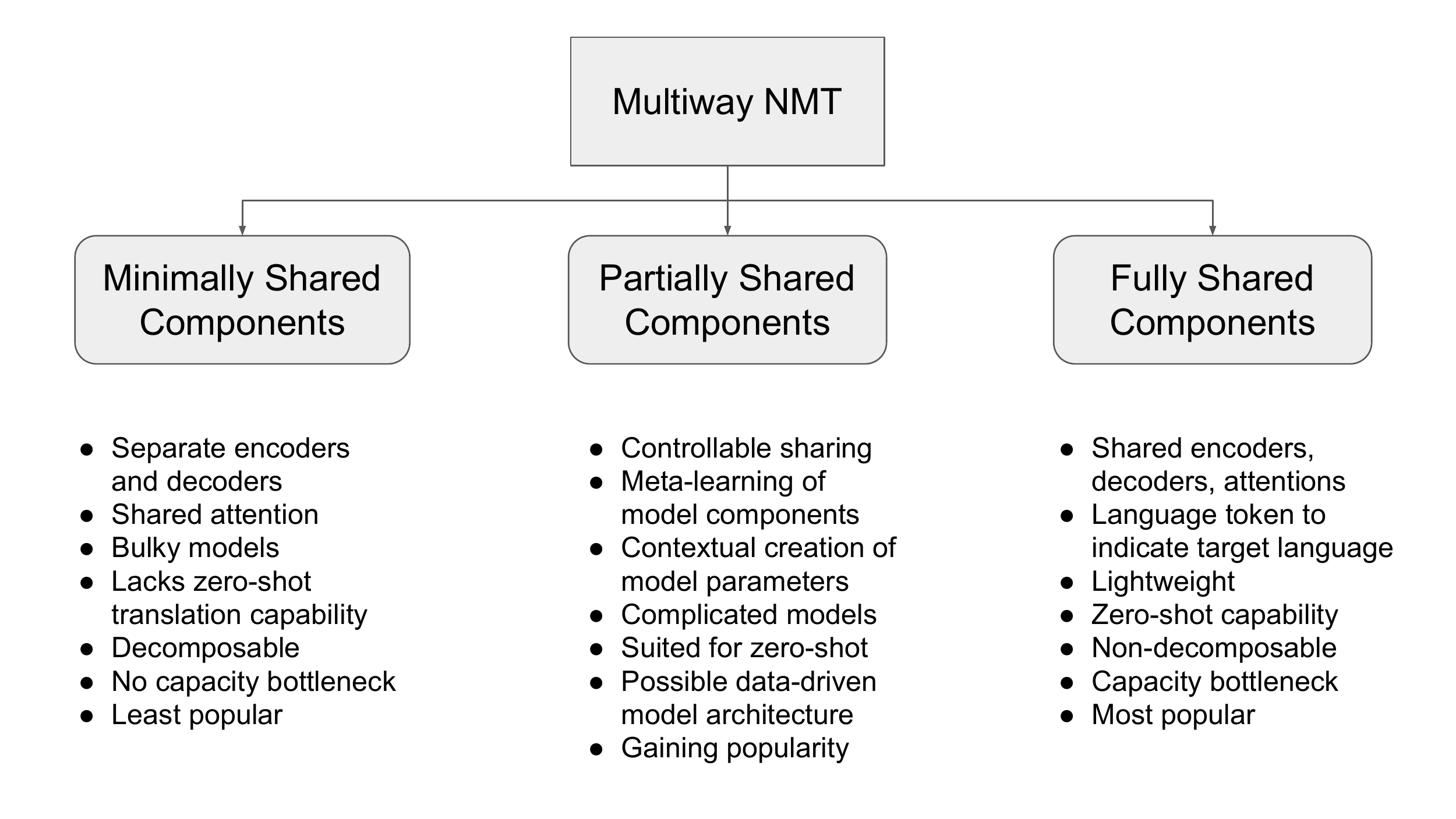}
        \caption{\REVISE{An overview of multiway NMT from the perspective of the level of sharing and the features and limitations each sharing approach. All types of MNMT models have to deal with complex training issues ranging from batching to language grouping to knowledge distillation. Additionally, it is important to address language divergence and finding the right balance of language-specific and language-agnostic representations.}}
        \label{fig:mnmtoverview}
\end{figure*}
}

\begin{figure*}
    \centering
    \includegraphics[width=\textwidth]{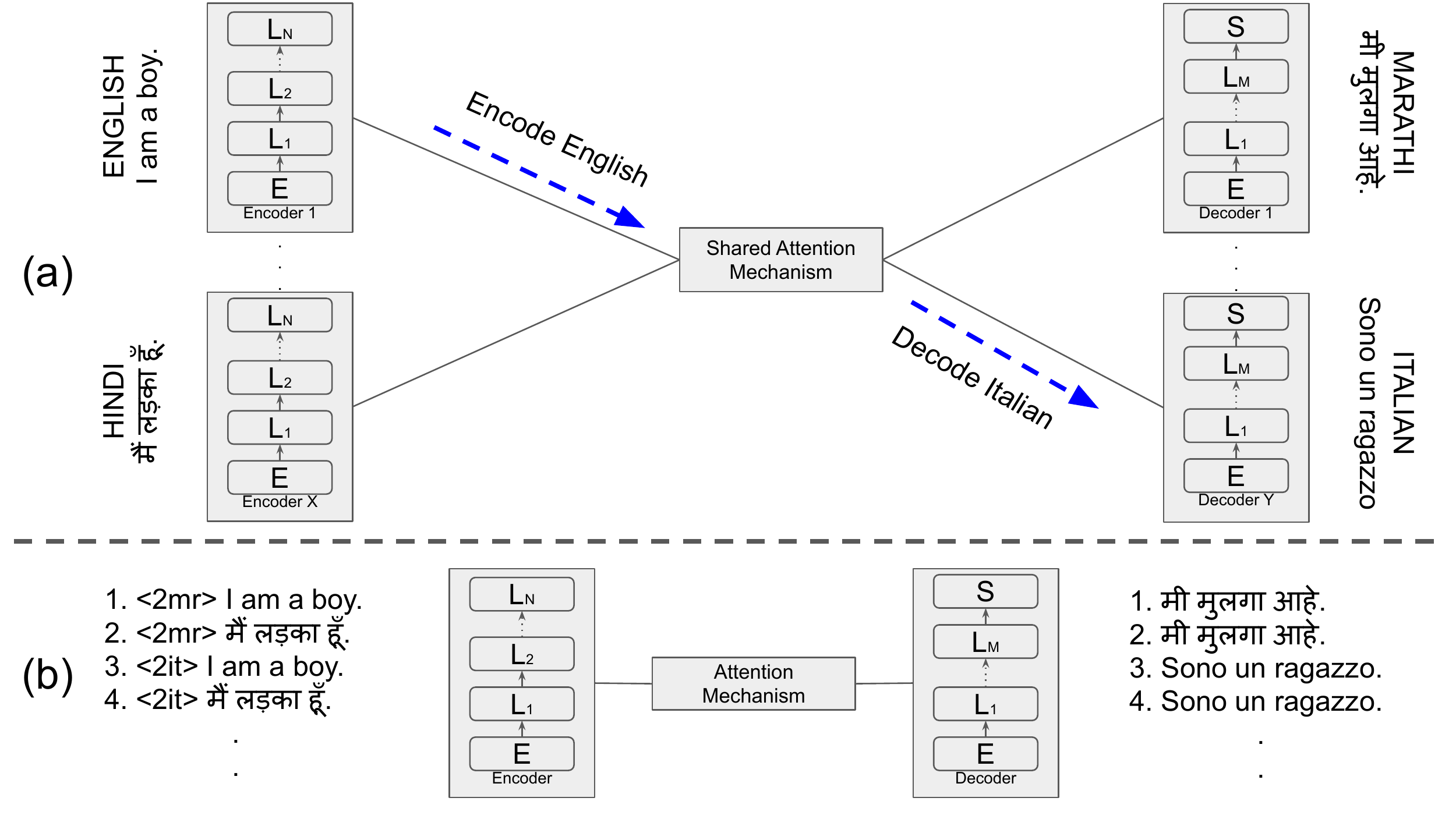}
        \caption{\REVISE{Part (a) of the figure shows a X sources (X encoders) and Y targets (Y decoders) MNMT model. The encoders contain N layers ($L^{1}$ to $L^{N}$) and decoders contain M layers (($L^{1}$ to $L^{M}$)) that can be stacked. Each layer can be recurrent, convolutional or feed-forward. The decoders contain a softmax layer (S) to predict target language words. A single attention mechanism is shared among all encoders and decoders. In the figure, the blue arrows indicate the flow of processing to translate an English sentence into its Italian translation. Given the vast number of components, they can be initialized by pre-trained models such as BERT in order to perform transfer learning. Post-training, this model can be decomposed into individual bilingual models. Part (b) of the figure shows a fully shared MNMT model for all language pairs. The ``language tag'' trick where a token like ``$<2xx>$'' is prefixed to each source sentence to indicate the desired target language. All existing MNMT models are adaptations of these two fundamental architectures.  Depending on the language pairs involved, the size of the data, the languages of focus and deployment conditions, one should consider choosing the components that should be shared.}}
        \label{fig:mlnmtoverview}
\end{figure*}

\REVISE{The primary goal of MNMT is a model that can support translation between more than one language pair. We use the term \textit{multiway} NMT models to denote such models.} Formally, a single model can support translation for $l$ language pairs $(src_l,tgt_l) \in \mathbf{L}$ ~ $(l=1 \textrm{ to } L)$, where $\mathbf{L} \subset S \times T$, and $S, T$ are sets of $X$ source and  $Y$ target languages respectively. $S$ and $T$ need not be mutually exclusive. Parallel corpora are available for \REVISE{\textbf{all}} of these $l$ language pairs as $C(src_l)$ and $C(tgt_l)$. Note that our objective in this specific scenario is to train a translation system between all language pairs. Refer to Figure~\ref{fig:mlnmtoverview} for the two prototypical MNMT approaches with minimal and complete sharing of components. Most existing works are variations of these models. Particularly, one-to-many \cite{dong15},  many-to-one \cite{lee17} and many-to-many \cite{firat16} NMT models are specific instances of this general framework. \REVISE{The training objective for multiway NMT is maximization of the mean translation likelihood across all language pairs: }

\begin{equation*}
\REVISE{\mathcal{L}_\theta = \frac{1}{L}\sum_{l=1}^{L} \mathcal{L}^{C(src_l), C(tgt_l)} (\theta),}
\end{equation*}
\REVISE{which can be calculated in the same way as Equation (\ref{loss}).}

Multiway NMT systems are of great interest since it is believed that transfer learning between languages can take place which will help improve the overall translation quality for many translation directions \cite{DBLP:conf/emnlp/ZophYMK16:original} and at the same time enable translations between language pairs with no data \cite{johnson17}. Analyzing multiway NMT systems could also provide an understanding of the relationship between languages from a statistical and linguistic point of view \cite{Dabre-MTS2017,D18-1103}. 

Multiway translation systems follow the standard encode-attend-decode paradigm in popular NMT systems. However, the architecture is adapted to support multiple languages. \REVISE{This involves addressing issues related to vocabularies and associated embeddings, stackable layers (RNN/CNN/Feed-Forward), parameter sharing, training protocols and language divergence.} We address each issue in this section. Refer to Figure~\ref{fig:mnmtoverview} for an overview of the multiway NMT paradigm.


\subsection{\REVISE{Parameter Sharing}}
\label{sec:proto}

There are a wide range of architectural choices in the design of MNMT models. The choices are primarily defined by the degree of parameter sharing among various supported languages.
primarily defined by the degree of parameter sharing among various supported languages. 

\compactpara{Minimal Parameter Sharing}. 
\citet{firat16} proposed a model comprised of separate embeddings, encoders and decoders for each language which all shared a single attention mechanism. \REVISE{Additionally there are two shared components: a layer for all encoders for initializing the initial decoder state by using the final encoder state and an affine layer for all decoders to project the final decoder state before computing softmax. However, the focus is on the shared attention layer as it has to bear most of the burden of connecting all source and target languages.} \REVISE{Figure~\ref{fig:mlnmtoverview}-(a) depicts a simplified view of this model. Different from the attention score calculated for a single language pair in Equation (\ref{attention_score}), the attention score in \citet{firat16} is calculated from multiple encoders and decoders as:}
\begin{equation*}
\label{attention}
\REVISE{e_{ji}^{vu}=a(s_{j-1}^{v}, h_i^{u}),}
\end{equation*}
\REVISE{where $u$ is the u-th encoder and $v$ is the v-th decoder.}
By sharing attention across languages, it was hypothesized that transfer learning could take place and such a model was able to outperform bilingual models, especially when the target language was English. However, this model has a large number of parameters. Nevertheless, the number of parameters only grows linearly with the number of languages, while it grows quadratically for bilingual systems spanning all the language pairs in the multiway system. However when \citet{johnson17} showed that such bulky models are not required, this particular approach for MNMT gradually lost popularity and there is very little influential research that uses this model. Instead, current research efforts are focused on a middle-ground where the amount of sharing is controlled. 

\vspace{2 mm}

\compactpara{Complete Parameter Sharing}. \citet{johnson17} proposed a highly compact model where all languages share the same embeddings, encoder, decoder, and attention mechanism. Typically, a common vocabulary across all languages is first generated using a subword-level encoding mechanism such as \REVISE{byte-pair encoding (BPE) \cite{DBLP:journals/corr/SennrichHB15}, word-piece model (WPM) \cite{conf/icassp/SchusterN12} or sentence-piece model (SPM) \cite{kudo-richardson-2018-sentencepiece}.} Thereafter, all corpora are concatenated\footnote{It is common to oversample smaller corpora so that all language pairs are equally represented in the model regardless of the size of the corpora for those pairs.} and the input sequences are prefixed with a special token (called the \textit{language tag}) to indicate the target language (see Figure~\ref{fig:mlnmtoverview}-(b)). This enables the decoder to correctly generate the target language despite all target languages sharing the same decoder parameters.  Note that the embedding and softmax layers are shared across all languages and \citet{ha16} proposed a similar model, but they maintained separate vocabularies for each language. While this might help in faster inference due to smaller softmax layers, the possibility of cognate sharing is lower, especially for linguistically close languages sharing a common script. This architecture is particularly useful for related languages, because they have high degree of lexical and syntactic similarity \cite{sachan18}. \REVISE{There is no conclusive proof of whether sharing, unification or separation of vocabularies provides better performance. An empirical analysis will be worthwhile. A few techniques through which lexical similarity can be further leveraged are}: 
\begin{itemize}
    \item representing all languages in a common script using script conversion \cite{dabre18wat,lee17} or transliteration (\citet{nakov2009improved} for multilingual SMT).
    \item using a common subword-vocabulary across all languages \textit{e.g.} character \cite{lee17} and BPE \cite{nguyen17}. 
    \item representing words by  both character encoding and a latent embedding space shared by all languages \cite{wang-ICLR2019}.    
\end{itemize}

Concerning the choice of non-embedding or softmax layers, \citet{rikters18} and \citet{lakew18} have compared RNN, CNN and the self-attention based architectures for MNMT. They show that self-attention based architectures outperform the other architectures in many cases. The most common hypothesis is that self-attention enables random access to all words in a sentence which leads to better word and sentence representations. 

The complete sharing approach treats the NMT system as a \textit{black box} to train a multilingual system. The model has maximum simplicity and has minimal parameter size as all languages share the same parameters; and achieves comparable/better results w.r.t. bilingual systems. \REVISE{Work on \compactpara{massively multilingual NMT} \cite{wild,aharoni19,bapna-firat-2019-simple} pushes the idea of completely shared models for all language pairs to its limits. \citet{wild,aharoni19} trained a single model for 59 and 103 languages and explore a wide range of model configurations focusing on data selection,\footnote{In particular they focused on the effects of temperature based data sampling on the multilingual models.} corpora balancing, vocabulary, deep stacking, training and inferencing approaches. While massively multilingual models have a strong positive impact on low-resource languages, they tend to benefit translation into English a lot more than from English. Furthermore, the gains in translation quality tend to taper off when using more than 50 languages. These works focus on translation performance between language pairs for which there was no training data but we will discuss this separately in Section~\ref{sec:unseen}. However, a massively multilingual system also runs into \compactpara{capacity bottlenecks} \cite{aharoni19,Siddhant-AAAI2020} where not all translation directions show improved performance despite a massive amount of data being fed to a model with a massive number of parameters. Future research should focus on better addressing the bottleneck issues, deeply stacked model training issues and designing new stackable layers which can handle a wide variety and a large number of languages. We strongly recommend readers to read the paper by \citet{wild} which itself is a survey on multilingual NMT on web-scale data.} From the perspective of understanding the working of such models, a toolkit for visualization and inspection of multilingual models should be extremely valuable.




\vspace{2 mm}
\compactpara{Controlled Parameter Sharing}
In between the extremities of parameter sharing exemplified by the above mentioned models, lies an array of choices. The degree of parameter sharing can be controlled at various layers of the MNMT system. A major factor driving the degree of parameter sharing is the divergence between the languages involved \cite{sachan18}.  

Sharing encoders among multiple languages is very effective and is widely used \cite{lee17,sachan18}. Keeping decoders separate is important because the burden of generation is mainly on the decoder. Therefore, the job of the encoder is relatively simpler which means that sharing an encoder between many languages leads to better parameter utilization. On the other hand, the decoder and its attention mechanism, should be as robust as possible. \citet{blackwood18} explored target language, source language and pair specific attention parameters. They showed that target language specific attention performs better than other attention sharing configurations, thus highlighting that designing a strong decoder is extremely important. For self-attention based NMT models, \citet{sachan18} explored various parameter sharing strategies. They showed that sharing the decoder self-attention and encoder-decoder cross-attention  parameters are useful for linguistically dissimilar languages. By sharing self- and cross-attention mechanisms in the decoder, the decoder most likely learns target language representations which are better aligned with source language representations. \REVISE{\citet{wang-etal-2019-compact} further proposed a mechanism to generate a universal representation instead of separate encoders and decoders to maximize parameter sharing. They also used language-sensitive embedding, attention, and discriminator for different languages. This helps control the amount of sharing in an indirect fashion. \citet{bapna-firat-2019-simple} also extend a fully shared model with language pair specific adaptor layers which are fine-tuned for those pairs. After training a fully-shared model, additional adaptor layers are inserted into the model and only those layers are fine-tuned which requires significantly lesser computation cost. This does lead to an increase in the number of parameters but it is modest compared to a minimally shared system.}

Fixing sharing configurations prior to training is ill-advised because sharing one set of parameters might be optimal for one language pair but not another. To this end, \citet{Zaremoodi-ACL2018} proposed a routing network to dynamically control parameter sharing where the parts to be shared depend on the parallel corpora used for training. On the other hand, \citet{platanios18} learned the degree of parameter sharing from the training data. This is achieved by defining the language specific model parameters as a function of global parameters and language embeddings. A base set of parameters is transformed into another set of parameters for a specific language using linear projections. If $\theta$ is the set of base parameters then the parameters for a specific language pair $src_{i}$ and $tgt_{j}$ can be obtained via a linear transformation $F(\theta)$. This linear transformation involves learning some additional parameters for projection but this involves far fewer parameters compared to modeling complete sets of parameters for each new language and thus this approach is quite attractive. It will be interesting to determine whether using non-linear projections is better than linear ones. This approach reduces the number of language specific parameters (only language embeddings), while still allowing each language to have its own unique parameters for different network layers. In fact, the number of parameters is only a small multiple of the compact model (the multiplication factor accounts for the language embedding size) \cite{johnson17}, but the language embeddings can directly impact the model parameters instead of the weak influence that language tags have.

Designing the right sharing strategy is important to maintaining a balance between model compactness and translation accuracy. It will be interesting to see more approaches that use the training data itself to enable a model to increase or decrease its own complexity or capacity. Reinforcement learning and genetic algorithms applied to neural architecture search (NAS) \cite{45826} can be one of the ways to achieve this.

\subsection{Addressing Language Divergence}
A central task in MNMT is alignment of representations of words and sentence across languages so that divergence in languages between bridged and the model can handle many languages. This involves the study and understanding of the representations learned by multilingual models and using this understanding to further improve modeling choices. The remainder of this subsection discusses these issues related to multilingual representations.

\vspace{2 mm}
\noindent \REVISE{\compactpara{The nature of multilingual representations.}
Since MNMT systems share network components across languages, they induce a relationship among representations of sentences across languages. Understanding the nature of these multilingual representations can help get insights into the working of multilingual models. Some works that visualize multilingual model embeddings suggest that the encoder learns similar representations for similar sentences across languages \cite{johnson17,Dabre-MTS2017}. Since these  visualizations are done in very low dimensions (2-3 dimensions), they might not convey the right picture about the language-invariance of multilingual representations. \citet{kudugunta-etal-2019-investigating} do a systematic study of representations generated from a massively, multilingual system using SVCCA \cite{raghu2017svcca}, a framework for for comparing representations across different languages, models and layers. Their study brings out the following observations on the nature of multilingual embeddings from a compact MNMT system:} 
\REVISE{
\begin{itemize}
\item While encoder representations for similar sentences are similar across languages, there is a fine-grained clustering based on language similarity. This explains why transfer learning works better with related languages (as discussed in Section \ref{sec:low}).
\item The boundary between the encoder and decoder is blurry and the source language representations  depend on the target language and vice-versa.
\item Representation similarity varies across layers. The invariance increases in higher layers on the encoder side. On the other hand, the invariance decreases in higher layers on the decoder side. This is expected since the decoder is sensitive to the target language to be generated. The decoder has to achieve the right balance between language agnostic and language aware representations. 
\end{itemize}
}

\REVISE{Language invariant representations seem like a nice abstraction, drawing comparisons to interlingual representations \cite{johnson17}. It has been a major goal of most MNMT research, as discussed in the subsequent sections. They have been shown to be beneficial for building compact models and transfer learning. Given these empirical observations from multiple sources, we think that language invariant representations in appropriate parts of the multilingual model is a desirable property.}

\vspace{2 mm}
\compactpara{Encoder Representation.} There are two issues that might make encoder representations language-dependent. Parallel sentences, from different source languages, can have different number of tokens. Hence, the decoder's attention mechanism sees a variable number of encoder representations for equivalent sentences across languages. To overcome this, an attention bridge network generates a fixed number of contextual representations that are input to the attention network \cite{lu18,vazquez18}. \REVISE{By minimizing the diversity of representations, the decoder's task is simplified and it becomes better at language generation. The choice of a single encoder for all languages is also promoted by \citet{hokamp-etal-2019-evaluating} who opt for language speific decoders.} \citet{rudramurthy19} pointed out that the sentence representations generated by the encoder are dependent on the word order of the language and are hence language specific. They focused on reordering input sentences in order to reduce the divergence caused due to different word orders in order to improve the quality of transfer learning.



\vspace{2 mm}
\compactpara{Decoder Representation.} The divergence in decoder representation needs to be addressed when multiple target languages are involved. This is a challenging scenario because a decoder should generate representations that help it generate meaningful and fluent sentences in each target language. \REVISE{Therefore a balance between learning language invariant representations and being able to generate language specific translations is crucial. If several parallel corpora for different language pairs are simply concatenated and then fed to a standard NMT model then the NMT model might end up generating a mixed language translation as a result of vocabulary leakage due to language invariant representations. The \textit{language tag} trick has been very effective in preventing vocabulary leakage \cite{johnson17} because it enables the decoder to clearly distinguish between different languages.\footnote{Our own experiments in extremely low resource scenarios show that it is impossible to completely avoid vocabulary leakage, especially when the same word is present in different languages. Employing a special bias vector helps mitigate this issue.} Further, \citet{blackwood18} added the language tag to the beginning as well as end of sequence to enhance its effect on the sentence representations learned by a left-to-right encoder. This shows that dedicating a few parameters to learn language tokens can help a decoder maintain a balance between language-agnostic and language distinct features. \citet{hokamp-etal-2019-evaluating} showed that more often than not, using separate decoders and attention mechanisms give better results as compared to a shared decoder and attention mechanism. This work implies that the best way to handle language divergence would be to use a shared encoder for source languages and different decoders for target languages. We expect that the balance between language agnostic and language specific representations should depend on the language pairs. \citet{tan-etal-2019-multilingual,PraR:2018} are some of the works that cluster languages into language families and train separate MNMT models per family. Language families can be decided by using linguistic knowledge\footnote{\url{https://en.wikipedia.org/wiki/List_of_language_families}} \cite{PraR:2018} or by using embedding similarities where the embeddings are obtained from a multilingual word2vec model \cite{tan-etal-2019-multilingual}. Comparing language family specific models and mixed-language family models shows that the former models outperform the latter models. In the future, when training a model on a large number of languages, researchers could consider different decoders for different language families and each decoder in turn can use the language token trick to generate specific languages in the language family assigned to the decoder.}

\vspace{2 mm}
\noindent \REVISE{\compactpara{Impact of Language Tag.} There are some works that explore the effects of the \textit{language tag} trick on the shared decoder, it's language agnostic (or lack thereof) representations and the final translation quality. 
\citet{wang18} explored multiple methods for supporting multiple target languages: (a) target language tag at beginning of the decoder, (b) target language dependent positional embeddings, and (c) divide hidden units of each decoder layer into shared and language-dependent ones. Each of these methods provide gains over \citet{johnson17}, and combining all methods gave the best results. \citet{hokamp-etal-2019-evaluating} showed that in a shared decoder setting, using a task-specific (language pair to be translated) embedding works better than using language tokens. We expect that this is because learning task specific embeddings needs more parameters and help the decoder learn better features to distinguish between tasks. In the related sub-area of pre-training using MASS \cite{song-etal-2019-code}, mBERT/XLM \cite{lample2019crosslingual} it was shown that using language or task tags is sufficient for distinguishing between the languages used for pre-training large models. Given that massive pre-training is relatively new, further exploration is much needed.}

\subsection{Training Protocols}

\REVISE{Apart from model architecture and language divergence issues is MNMT model training which requires sophisticated methods}. The core of all methods is the minimization of the negative log-likelihood of the translation for all language pairs. Conceptually, the negative log-likelihood of the MNMT model is the average of the negative log-likelihoods of multiple bilingual models. As we have mentioned before, minimizing the negative log-likelihood can be done using one or more gradient descent based optimization algorithms \cite{DBLP:journals/corr/Ruder16} such as SGD, ADAM, ADAGRAD and so on. There are two main types of training approaches: single stage or parallel or joint training and sequential or multi-stage training.
\vspace{2 mm}

\noindent \REVISE{\compactpara{Single Stage Parallel/Joint Training.} We simply pre-process and concatenate the parallel corpora for all language pairs and then feed them to the model batch-wise. For models with separate encoders and decoders, each batch consists of sentence pairs for a specific language pair \cite{firat16} whereas for fully shared models, a single batch can contain sentence pairs from multiple language pairs \cite{lee17,johnson17}. As some language pairs might have more data than other languages, the model may be biased to translate these pairs better. To avoid this, sentence pairs from different language pairs are sampled to maintain a healthy balance. The most common way of maintaining balance is to oversample smaller datasets to match the sizes of the largest datasets but \citet{wild} worked on temperature based sampling and showed its cost-benefit analysis.} 

\vspace{2 mm}
\noindent \REVISE{\compactpara{Multi-stage Training.} The fundamental principle behind this type of training is that joint training of a model from scratch suffers from capacity bottlenecks on top of the increased model complexity. Therefore, training a model in multiple stages should help simplify the learning. On the other hand, not all language pairs may be available when the initial MNMT model is trained. It will be expensive to re-train the multilingual model from scratch when a language pair has to be added. A practical concern with training MNMT in an {\bf incremental} fashion is dealing with vocabulary. Capacity bottlenecks can be mostly addressed through knowledge distillation whereas incremental inclusion of languages can be addressed by dynamically upgrading the model's capacity before training on newer data.} 

\paragraph{Knowledge Distillation} Knowledge distillation was originally proposed by \citet{HinVin15Distilling} for tasks that do not involve generating sequences such as image classification. The underlying idea is to train a large model with many layers and then distill it's knowledge into a small model with fewer layers by training the small model on the softmax generated by the large model instead of the actual labels. This approach does not work well for sequence generation and hence \citet{kim-rush-2016-sequence} proposed sequence distillation where they simply translate the training data using a pre-trained model. They then use this pseduo-data to train smaller models. These smaller models are able to learn faster and better because of the reduced burden for decision making. Following this, \citet{tan18} trained bilingual models for all language pairs involved and then these bilingual models are used as \textit{teacher models} to train a single \textit{student model} for all language pairs. The student model is trained using a linear interpolation of the standard likelihood loss as well as distillation loss that captures the distance between the output distributions of the student and teacher models. The distillation loss is applied for a language pair only if the teacher model shows better translation accuracy than the student model on the validation set. This approach shows better results than joint training of a black-box model, but training time increases significantly because bilingual models also have to be trained. 

\REVISE{\paragraph{Incremental Training} These approaches aim to decrease the cost of incorporating new languages or data in multilingual models by avoiding expensive retraining. Some works alter a pre-trained model's capacity to accommodate new languages. \citet{DBLP:journals/corr/abs-1811-01137} update the vocabulary of the parent model with the low-resource language pair's vocabulary before transferring parameters.} Embeddings of words that are common between the low- and high-resource languages are left untouched and randomly initialized embeddings may be used for as yet unseen words. A simpler solution would be to consider a universal romanised script so that incorporating a new language can be a matter of resuming training or fine-tuning. \REVISE{Work by \citet{escolano19} focuses on first training bilingual models and then gradually increasing their capacity to include more languages. To address capacity bottlenecks, \citet{bapna-firat-2019-simple} propose expanding the capacities of pre-trained MNMT models (especially those trained on massive amounts of multilingual data) using tiny feed-forward components which they call adaptors. For each language pair in a pre-trained (multilingual) model, they add adaptors at each layer  and fine-tune them on parallel corpora for that direction. These modular adaptor layers can be considered as experts that specialize for specific language pairs and can be added incrementally. They showed that this approach can help boost the performance of massively multilingual models trained by \citet{wild,aharoni19}. Note that this model can be used to incorporate new data, but new language pairs cannot be added.}

\vspace{2 mm}
A major criticism of all these approaches is that the MNMT models are trained in the same way as a regular NMT model. Most researchers tend to treat all language pairs treated equally, with the exception of oversampling smaller corpora to match the sizes of the larger corpora, and tend to ignore the fact that NMT might be able to handle some language pairs better than others. \REVISE{ However, there are works that propose that focus on scaling learning rates or gradients differently for high-resource and low resource language pairs \cite{Jean2019AdaptiveSF}. On the other hand, \citet{kiperwasser-ballesteros-2018-scheduled} proposed a multi-task learning model for learning syntax and translation where they showed different effects of their model for high-resource and low resource language pairs. Therefore, we suppose that the MNMT models in the vast majority of papers are sub-optimally trained even if they improve the translation quality for low-resource languages. In hindsight, this particular aspect of MNMT training deserves more attention.} 

\REVISE{Although multiway MNMT modeling has been thoroughly explored there are still a number of open questions, especially the bottleneck capacity and parameter sharing protocols. The next section will pay special attention of MNMT in resource constrained scenarios.}

\section{MNMT for Low-resource Language Pairs}
\label{sec:low}

\begin{figure*}
    \centering
    \includegraphics[width=\textwidth]{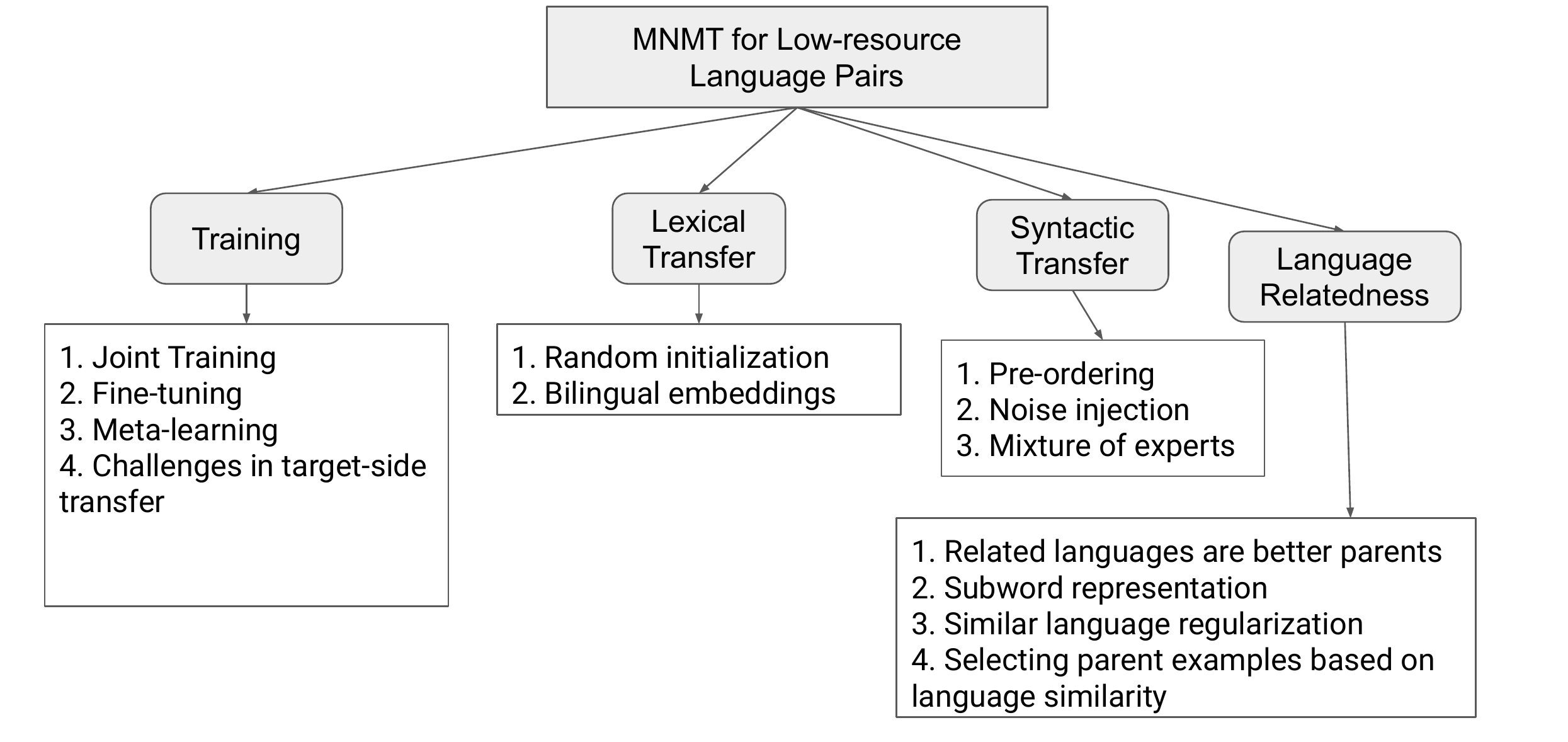}
        \caption{An overview of MNMT for low-resource languages pairs depicting various aspects of proposed solutions.}
        \label{fig:low_overview}
\end{figure*}

Many language pairs have a limited amount of parallel corpora, that is insufficient for training high-quality NMT systems. While data augmentation strategies like back-translation \cite{sennrich-haddow-birch:2016:P16-11} and self-learning \cite{he2020revisiting} can improve translation quality via regularization and domain adaptation, parallel training signals are limited. Can MNMT provide additional parallel training signals from a high-resource language pair (\eg Spanish-English) to improve low-resource MT (\eg Catalan-English) ? Since MNMT systems strive to represent multiple languages in the same vector space, it should be possible to utilize data from high-resource language pairs to improve translation of low-resource language pairs. Such \textit{transfer learning} \cite{Pan:2010:STL:1850483.1850545} approaches have been widely explored in the literature. The high-resource language (model) is often referred to as a \textit{parent language (model)} whereas the low-resource language (model) is known as a \textit{child language (model)}. The related literature has focussed on four aspects of transfer learning: (1) training procedures to ensure maximum transfer, (2) addressing lexical divergence between the parent and child languages, (3) dealing with the differences in syntax between the parent and child languages, and (4) effect and utilization of language relatedness.
The remainder of this section discusses these aspects. Figure \ref{fig:low_overview} depicts these aspects.

\subsection{Training}
 Most studies have explored transfer learning on the source-side: the high-resource and low-resource language pairs share the same target language. The simplest approach is \textbf{jointly training} both language pairs \cite{johnson17}. However, the final model may not be optimally tuned for the child language pair, whose performance we are interested in. 

A better approach is to \textbf{fine-tune} the parent model with data from the child language pair. \citet{DBLP:conf/emnlp/ZophYMK16:original} showed that such transfer learning can benefit low-resource language pairs. First, they trained a \textit{parent model} on a high-resource language pair. The \textit{child model} is initialized with the parent's parameters wherever possible and trained on the small parallel corpus for the low-resource pair. They also studied the effect of fine-tuning only a subset of the child model's parameters (source and target embeddings, RNN layers and attention). They  observed that fine-tuning all parameters except the input and output embeddings was optimal in one setting. However, further experiments are required for drawing strong conclusions. 

Training the parent model to optimality may not be the best objective for child tasks. It may be preferable that parent model parameters are amenable to fast adaptation/fine-tuning on child tasks. Learning such a parent model is referred to as \textbf{meta-learning} and can generalize well to child tasks. \citet{gu18b} used the model-agnostic meta-learning (MAML) framework \cite{finn17} to learn  appropriate parameter initialization from the parent pair(s) by taking the child pair into consideration. The meta-learning based approach significantly outperform simple fine-tuning. They also show that having more parent language pairs (jointly trained) also improves performance on the child language pairs. 

Transfer learning on the target-side has been more challenging than transfer learning on the source-side. Distinct target languages require target language specific representations, while transfer learning prefers target language invariant representations. The success of transfer learning relies on achieving the right balance between these factors. \citet{johnson17} show that joint training does not provide any significant benefit. Fine-tuning is beneficial in very low-resource scenarios \cite{dabre-etal-2019-exploiting}, but gains may be limited due to catastropic forgetting. \citet{dabre-etal-2019-exploiting} show that a multi-step fine-tuning process is beneficial when multiple target languages are involved. They do not focus on language divergence during their multilingual multi-stage tuning but show that the size of helping data matters. From their work, it will be worthwhile to consider involving multiple medium sized (few hundreds of thousands of lines) helping corpora involving a variety of languages. This multilingual multi-stage transfer learning scenario requires further research.

\subsection{Lexical Transfer} \citet{DBLP:conf/emnlp/ZophYMK16:original} randomly initialized the word embeddings of the child source language, because those could not be transferred from the parent. However, this approach does not map the embeddings of similar words across the source languages \textit{apriori}. \citet{gu18} improved on this simple initialization by mapping pre-trained monolingual word embeddings of the parent and child sources to a common vector space. This mapping is learned via the orthogonal Procrustes method \cite{schonemann1966generalized} using a bilingual dictionaries between the sources and the target language \cite{he2020revisiting}. \citet{kim19} proposed a variant of this approach where the parent model is first trained and monolingual word-embeddings of the child source are mapped to the parent source's embeddings prior to fine-tuning. While \citet{gu18} require the child and parent sources to be mapped while training the parent model, the mapping in \citet{kim19}'s model can be trained after the parent model has been trained. 

\subsection{Syntactic Transfer} 
It is not always possible to have parent and child languages from the same language family and hence blindly fine-tuning a parent model on the child language data might not take into account the syntactic divergence between them. Although it is important to address this issue there are surprisingly few works that address this issue. \citet{rudramurthy19} showed that reducing the {\it word order divergence} between source languages by \textbf{pre-ordering} the parent sentences to watch child word order is beneficial in extremely low-resource scenarios. We expect the future work will focus on syntactic divergence between languages via NMT models that handle reordering implicitly rather than rely on pre-ordering. \citet{kim19} take a different approach to mitigate syntactic divergence. They train the parent encoder with \textbf{noisy source data} introduced via probabilistic insertion and deletion of words as well as permutation of word pairs. This ensures that the encoder is not over-optimized for the parent source language syntax. \citet{gu18} proposed to achieve better \textbf{transfer of syntax-sensitive contextual representations} from parents using a mixture of language experts network.



\subsection{Language Relatedness} 
Results from the above-mentioned approaches show that, in many cases, transfer learning benefits the child language pair even if the parent and child languages are not related. These benefits are also seen even when the languages have different writing systems. \REVISE{These are interesting findings and can be attributed to the ability of the NMT models to learn cross-lingual representations as discussed in previous sections.} It is natural to ask if language relatedness can impact the effectiveness of lexical and syntactic transfer. Learning cross-lingual embeddings, used for lexical transfer, is difficult for distant languages \cite{sogaard2018limitations,pires-etal-2019-multilingual}.

\citet{DBLP:conf/emnlp/ZophYMK16:original} and \citet{Y17-1038} empirically showed that a related parent language benefits the child language more than an unrelated parent. 
\citet{Maimaiti:2019:MTL:3327969.3314945} further showed that using multiple highly related high-resource language pairs and applying fine tuning in multiple rounds can improve translation performance more, compared to only using one high-resource language pair for transfer learning.
\citet{kocmi-bojar:2018:WMT} presented a contradictory result in the case of Estonian (related Finnish \vs unrelated Czech parent) and suggest that size of the parent is more important; but most of the literature suggests that language relatedness is beneficial. It is probably easier to overcome language divergence when the languages are related. 

Further, language relatedness can be explicitly utilized to improve transfer learning. Language relatedness is typically exploited by using shared BPE vocabulary and BPE embeddings between the parent and child languages \cite{nguyen17}. \citet{Maimaiti:2019:MTL:3327969.3314945} used a unified transliteration scheme at the character level. This utilizes the \textbf{lexical similarity} between the languages and shows significant improvements in translation quality. \citet{D18-1103} used \textbf{``similar language regularization''} to prevent overfitting when rapidly adapting a pre-trained, massively multilingual NMT model (universal model) for low-resource languages. While fine-tuning the universal model for a low-resource pair, overfitting is avoided by using a subset of the training data for a related high-resource pair along with the low-resource pair. \citet{DBLP:journals/corr/Chaudhary19} used this approach to translate 1,095 languages to English. Further, not all parallel data from the parent task may be useful in improving the child task. \citet{wang19} proposed \textbf{selection of sentence pairs from the parent task} based on the similarity of the parent's source sentences to the child's source sentences. The significant gains from simple methods described point to the value of utilizing language relatedness. Further methods should be explored to create language invariant representations specifically designed for related languages.


While low-resource translation is hard in itself, an even more extreme scenario where no direct data exists between language pairs of interest. The next section discusses literature related to  this scenario.


 

\section{MNMT for Unseen Language Pairs} \label{sec:unseen}

\begin{figure*}
    \centering
    \includegraphics[width=\textwidth]{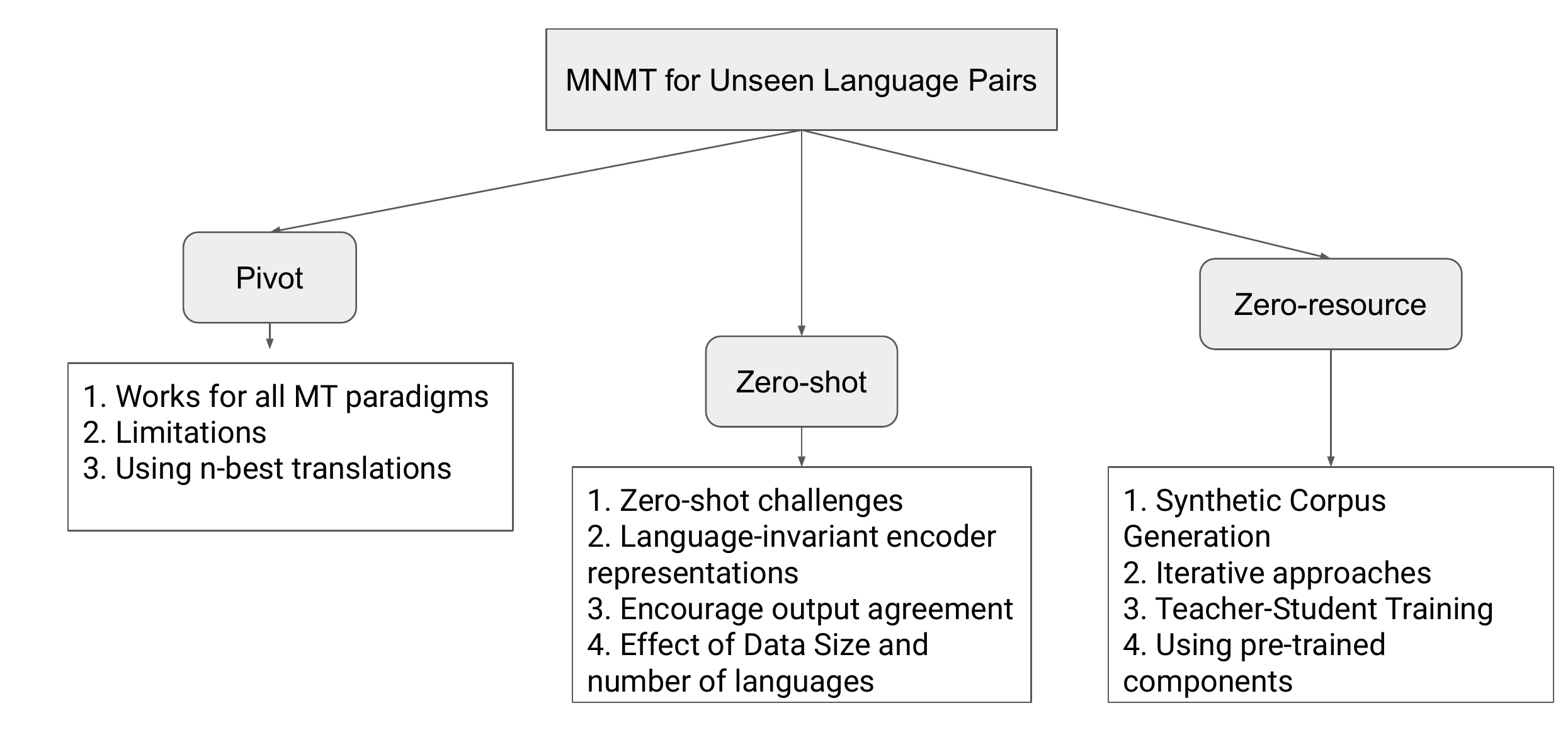}
        \caption{An overview of MNMT for unseen languages pairs. There are three broad-approaches: pivot, zero-shot and zero-resource approaches.}
        \label{fig:unseen_overview}
\end{figure*}

Providing machine translation solutions for arbitrary language pairs remains a challenge since little to no parallel corpora exists for most language pairs.\footnote{N-way translations of the Bible may be amongst the rare source of parallel corpora across arbitrary language pairs, but it is not sufficient for training a general-purpose MT system.} Unsupervised NMT \cite{lample2018unsupervised} has shown promise and generated some interested in recent years, but their quality remains way behind supervised NMT systems for most language pairs.

Can we do better than unsupervised NMT by utilizing multilingual translation corpora? A key observation is: even if two languages do not have a parallel corpus, they are likely to share a parallel corpus with a third language (called the \textit{pivot} language). In many cases, English is likely to be the pivot language given its widespread global usage. Pivot language based translation is a type of multilingual MT involving corpora for two language pairs: source-pivot and pivot-target parallel corpora, and has been widely explored for translation between unseen languages. In addition to simple pivot translation, zero-shot and zero-resource MNMT approaches have been proposed for translation between unseen language pairs. These approaches are described in the reminder of this section. Figure \ref{fig:unseen_overview} give an overview of the major approaches and issues for translation between unseen language pairs.

\subsection{Pivot Translation}

The simplest approach to pivot translation is building independent source-pivot (S-P) and pivot-target (P-T) MT systems. At test time, the source sentence cascades through the S-P and P-T systems to generate the target sentence. This simple process has two limitations: (a) translation errors compound in a pipeline, (b) decoding time is doubled since inference has be run twice. Sometimes, more than one pivot may be required to translate between the source and pivot language. Increased pipeline length exacerbates the above mentioned problems. The quality of the source-pivot translation is a bottleneck to the system. A variant of this approach extracts n-best translations from the S-P system. For each pivot translation, the P-T system can generate m-best translations. The $n\times m$ translation candidates can be re-ranked using scores from both systems and external features to generate the final translation. This approach improves the robustness of the pipeline to translation errors. Note that this pivot method is agnostic to the underlying translation technology and can be applied to SMT \cite{utiyama2007}, RBMT \cite{wuandwang2009} or NMT \cite{lakew-etal-2017-zeroshot} systems. Note that the pivoting can be done using an MNMT system too \cite{lakew-etal-2017-zeroshot,gu-etal-2019-improved} (as opposed to pivoting via bilingual systems).

\subsection{Zero-shot Translation}

Multilingual NMT models offer an appealing possibility. Even if the MNMT system has not been trained for the unseen language pair, \citet{johnson17} showed that the system is able to generate reasonable target language translations for the source sentence. Along with the source sentence, the desired output language's language tag is provided as input. This is sufficient for the system to generate output in the target language. Note that the MNMT system was exposed to zero bilingual resources between the source and target languages during training and encountered the unseen language pair only at test-time. Hence, this translation scenario is referred to as \textit{zero-shot translation}. 

The appeal of zero-shot translation is two-fold: \begin{itemize}
    \item Translation between any arbitrary language pair can be done in a single decoding step, without any pivoting, irrespective of the number of implicit pivots required to bridge the source and target language in the multilingual training corpus. The MNMT system can be conceived as an \textit{implicit pivoting} system.
    \item Given a multilingual parallel corpus spanning N languages, only a single multilingual NMT model is required to translate between $N\times (N-1)$  languages. 
\end{itemize}

\vspace{2 mm}
\compactpara{Limitations of Zero-shot Translation.} It has been argued that training an MNMT system incorporating multiple languages could benefit zero-shot translation due to better inter-lingual representations and elimination of cascading errors \cite{johnson17}. The simple zero-shot system described above, though promising, belies these expectations and its performance is generally lower than the pivot translation system \cite{johnson17,arivazhagan2018missing,gu-etal-2019-improved,pham-etal-2019-improving}. 

Some researchers have analyzed zero-shot translation to understand its underperformance. The following reasons have been suggested as limitations of zero-shot translation: 
\paragraph{Spurious Correlations between input and output language} During training, the network is not exposed to the unseen pairs. In its quest to capture all correlations in the training data, the model learns associations between the input representations and the target language for the observed language pairs. At test time, irrespective of the target language tag, the network will tend to output a language it has already observed with the source language in the training data \cite{gu-etal-2019-improved}. \citet{arivazhagan2018missing} showed that translation quality is closer to pivot systems if evaluation is restricted to sentences where the correct target language is generated.
\paragraph{Language variant encoder representations} The encoder representations generated by the model for equivalent source and pivot languages are not similar. Due to this discrepancy, the output generated by the decoder will be different for the source and pivot representations \cite{arivazhagan2018missing,kudugunta-etal-2019-investigating}. 

\vspace{2 mm}
\compactpara{Improving Zero-shot Translation.} The following methods have been suggested for generating language invariant encoder representations, while still achieving zero-shot translation: 

\paragraph{Minimize divergence between encoder representations} During training, additional objectives ensure that the source and pivot encoder representations are similar. \citet{arivazhagan2018missing} suggested an unsupervised  approach to align the source and pivot vector spaces by minimizing a domain adversarial loss \cite{ganin2016domain} - a discriminator is trained to distinguish between different encoder languages using representations from an adversarial encoder. Since S-P parallel corpora is also available, supervised loss terms which penalize divergence in source and pivot representations for each sentence pair have also be  explored. Different loss functions like cosine distance \cite{arivazhagan2018missing}, Euclidean distance \cite{pham-etal-2019-improving}, correlation distance \cite{saha2016correlational} have been shown to be beneficial in reducing the source/pivot divergence. \citet{ji2020cross} proposed to use pre-trained cross-lingual encoders trained using multilingual MLM, XLM and BRLM objectives  to obtain language-invariant encoder representations. 

\paragraph{Encourage output agreement} \citet{al-shedivat-parikh-2019-consistency} incorporated additional terms in the training objective to encourage source and pivot representations of parallel sentences to generate similar output sentences (synthetic) in an auxiliary language (possibly an unseen pair). This also avoids spurious correlations since the network learns to generate unseen pairs. Similar considerations motivated \citet{pham-etal-2019-improving} to add a pivot auto-encoding task in addition to the source-pivot translation task. They incorporate additional loss terms that encourage the attention-context vectors as well as decoder output representations to agree while generating the same pivot output. 

\paragraph{Effect of corpus size and number of languages} \citet{aharoni19} suggested that the zero-shot performance of multilingual NMT system increases with the number of languages incorporated in the model. It is not clear if the approaches mentioned above to address zero-shot NMT limitations can scale to a large number of languages. \citet{arivazhagan2018missing} showed that cosine distance based alignment can be scaled to a small set of languages.  Some studies suggest that zero-shot translation works reasonably well only when the multilingual parallel corpora is large \cite{Giulia-MTS2017,DBLP:journals/corr/abs-1811-01389}.


\paragraph{Addressing wrong language generation} To address the problem of generation of words in the wrong language,  \citet{DBLP:journals/corr/abs-1711-07893} proposed to filter the output of the softmax forcing the model to translate into the desired language. This method is rather effective despite its simplicity. 

\vspace{2mm}
The zeroshot approaches discussed above can complement multiway NMT systems described in Section \ref{sec:multiway} to support translation between the language pairs the model has not observed during training. Further work is needed to establish if these methods can scale to massively multilingual models.

\subsection{Zero-resource Translation}

The zero-shot approaches discussed in the previous section seek to avoid any training specific to unseen language pairs. Their goal is to enable training massively multilingual NMT systems that can perform reasonably well in zero-shot directions, without incurring any substantial overhead during training for all unseen language heads. When there is a case for optimizing the translation quality of an unseen pair, the training process can also consider objectives specific to the language pair of interest or tune the system specifically for the language pair of interest. Such approaches are referred to as \textit{zero-resource} approaches. Their training objectives and regimen customized for the unseen language pair of interest distinguish them from \textit{zero-shot} approaches. Note that zero-resource approaches do not use any true source-target parallel corpus. 

The following approaches have been explored for zero-resource translation.

\vspace{2 mm}
\compactpara{Synthetic Corpus Generation.} The pivot side of the P-T parallel corpus  is back-translated to the source language. The back-translation can be achieved either though zero-shot translation or pivot translation, creating a synthetic S-T parallel corpus. A S-T translation model can be trained on this synthetic parallel corpus. Adding the synthetic corpus to the multilingual corpus helps alleviate the spurious correlation problem. Some works have shown that this approach can outperform the pivot translation approach \cite{lakew-etal-2017-zeroshot,gu-etal-2019-improved}. The source is synthetic, hence there is be a difference between the training and test scenarios. Further, synthetic parallel corpus can be generated from monolingual pivot data too \cite{currey-heafield-2019-zero}. 

\vspace{2 mm}
\compactpara{Iterative approaches.} The S-T and T-S systems can be trained iteratively such that the two directions reinforce each other \cite{lakew-etal-2017-zeroshot}. \citet{DBLP:journals/corr/abs-1805-10338} jointly trained both the models incorporating language modelling and reconstruction objectives via reinforcement learning. The LM objective ensures grammatical correctness of translated sentences while the reconstruction objective ensures translation quality. The major shortcoming of iterative approaches is that they usually do not yield improvements after the first 2-3 iterations and are extremely time-consuming.  

\vspace{2 mm}
\compactpara{Teacher-Student Training.} \citet{P17-1176} assumed that the source and pivot sentences of the S-P parallel corpus will generate similar probability distributions for translating into a third language (target). They build the S-T model (`student') without any parallel S-T corpus by training the model to follow a P-T model (`teacher'). They propose two approaches: sentence-level mode approximation and word-level KL divergence. Both approaches outperform the pivot baseline with the latter showing better performance. A shortcoming of the first method is reliance on S-T parallel corpus with synthetic target, whereas the second approach learns to follow a soft distribution. 

\vspace{2 mm}
\compactpara{Combining Pre-trained encoders and decoders.}
\citet{kim-etal-2019-pivot} combined S-P encoder with P-T decoder to create the S-T model. They improved the simple initialization using some fine-tuning objectives and/or source-pivot adaptation to ensure that source and pivot representations are aligned. Whether this can be done via a fully shared MNMT model or not is unclear.

\vspace{2 mm}
Most the methods discussed in this section are either just competitive with or slightly better than simple pivot translation. This opens opportunities in understanding the challenges in unseen language translation and exploring solutions.

\section{Multi-Source NMT}
\label{sec:multisource}


\begin{table}[t]
\centering
\resizebox{\textwidth}{!}{%
\begin{tabular}{|c|c|c|c|c|c|}
\hline
\multicolumn{2}{|c|}{\textbf{Multi-Source Approach}} & \begin{tabular}[c]{@{}c@{}}\textbf{N-way data}\\ \textbf{needed}\end{tabular} & \textbf{Solutions} & \textbf{Concerns} & \textbf{Benefits} \\ \hline
\textbf{1} & \textbf{Vanilla} & Yes & Multi or Shared Encoder model & \begin{tabular}[c]{@{}c@{}}Long training times\\ Bulky Models\end{tabular} & Better for expanding N-way corpora \\ \hline
\textbf{2} & \textbf{Ensembling} & No & Ensemble multiple bilingual models & \begin{tabular}[c]{@{}c@{}}Learning ensemble functions\\ need small N-way corpora\end{tabular} & Reuse existing bilingual models \\ \hline
\textbf{3} & \textbf{Synthetic data} & No & Generate missing source sentences & Infeasible for real time translation & Applicable for post-editing \\ \hline
\end{tabular}%
}
\caption{An overview of multi-source NMT approaches based on the availability of N-way data, the training concerns and benefits concerns.}
\label{tab:msoverview}
\end{table}
If a source sentence has already been translated into multiple languages then these sentences can be used together to improve the translation into the target language. This technique is known as multi-source MT \cite{och2001statistical}. The underlying principle is to leverage redundancy in terms of source side linguistic phenomena expressed in multiple languages.

\vspace{2 mm}

\noindent \REVISE{\compactpara{Why Multi-source MT?} At first, it may seem unrealistic to assume the availability of the same sentence in multiple languages but we would like to point to parliamentary proceedings in the European Union (EU) or India. In the EU there are more than 10 official languages and in India there are more than 22 official languages. The EU parliamentary proceedings are maintained in multiple languages especially when they are concerning issues affecting multiple member nations. Therefore, it is extremely common for human interpreters to simultaneously interpret a speaker's language into several other languages. In such a situation, instead of employing a large number of interpreters for each language, it is possible to employ interpreters for a subset of languages and then leverage multi-source machine translation of these subsets of languages which can give much better translation quality as compared to singe-source machine translation. Furthermore, these improved translations can be edited by translators and added to the data-set that is used to train multi-source MT systems.}  Thus, it will be possible to create N-lingual (N $>$ 3) corpora such as Europarl \cite{koehn2005epc} and UN \cite{ZIEMSKI16.1195}.
Refer to Table~\ref{tab:msoverview} for a simplified overview of the multi-source NMT paradigm. There are three possible resource/application scenarios where multi-source NMT can be used.


\vspace{2 mm}
\compactpara{Multi-Source Available.}
\REVISE{Most studies assume that the same sentence is available in multiple languages although this need not be true. However, whenever multi-source sentences are available it is crucial to leverage them.} Just like multiway NMT models, a multi-source NMT model can consist of multiple encoders or a single encoder. \citet{N16-1004} showed that a multi-source NMT model using {\it separate encoders and attention networks} for each source language outperforms single source models. In their model, each encoder generates representations for different source language sentences and the decoder has a separate attention mechanism per source language, \REVISE{and the attention weight in Equation (\ref{attention_score}) is calculated as:}
\begin{equation*}
\label{attention_multisrc}
\REVISE{e_{ji}^{src_l}=a(s_{j-1}, h_i^{src_l}),}
\end{equation*}
\REVISE{where $src_l$ indicates a source language. The separate attentions are concatenated and then used in the decoder. This leads to an increase in the hidden layer size of the decoder and hence the number of decoder parameters. This is often acceptable unless large number of source languages are to be used. Although not explored anywhere, we suppose that a linear transformation can be used to down-project the concatenation of the context vectors and thus prevent the large hidden sizes in the decoder.}
\REVISE{A simpler approach concatenated multiple {\it source sentences} and fed them to a standard NMT model \cite{Dabre-MTS2017}, with performance comparable to \citet{N16-1004}. This model was obviously inspired by the success of fully shared MNMT models \cite{johnson17}. A single encoder is responsible for encoding a long multilingual sentence.\footnote{Note that the order of the input sentences should be the same during training and testing.}}
Interestingly, this model could automatically identify the boundaries between different source languages and simplify the training process for multi-source NMT. \citet{Dabre-MTS2017} also showed that it is better to use linguistically similar source languages, especially in low-resource scenarios. \REVISE{Both studies showed that the attention mechanisms tend to prefer some languages over the other. Especially, linguistically distant languages are practically ignored with computing context for decoding. It is possible to speed up encoding by parallellizing the encoders.} 

{\it Ensembling} of individual source-target models is another beneficial  approach, for which \citet{C16-1133} proposed several methods with different degrees of parameterization. They proposed to learn an ensembling function to combine the softmaxes of several bilingual models. This approach requires a smaller N-lingual corpus but training an ensembling function can be costly in terms of parameters. On the other hand, \citet{Dabre-MTS2017} have shown that it is possible to perform naive ensembling\footnote{This means uniform avergaging of softmaxes of different models as opposed to weighted averaging.} and still get reasonable improvements in translation quality. \REVISE{Note that the ensembling method used by \citet{Dabre-MTS2017} is the same as the late averaging technique proposed by \citet{firat16b}. Although the black-box method by \citet{Dabre-MTS2017} is significantly simpler and more elegant than the method in \cite{N16-1004}, most works on multi-source NMT use the latter method.}

\vspace{2 mm}
\compactpara{Missing Source Sentences.}
There can be missing source sentences in multi-source corpora and during translation. \citet{W18-2711} extended \cite{N16-1004} by representing each ``missing'' source language with a {\it dummy token}. The NMT model manages to adapt to missing sentences and thus manages to give improved translations when all source sentences are available.
\citet{choi2018improving} and \citet{nishimura18iwslt} proposed to use MT generated {\it synthetic sentences}, instead of a dummy token for the missing source languages. \REVISE{NMT models are used to translate sentences from the available source languages into the missing ones. When sentences for all source languages are ready, any standard multi-source approach may be used.} The fact that such a simple ``trick'' works so well means that combining all these methods allow the training of a single model with both, single as well as multi-source capabilities. \REVISE{Therefore, a single model can be used in different deployment conditions. If low-latency translation services are desired then the model can be used in single-source mode. In moderate-latency scenarios, only a few source languages could be used. In offline translation scenarios, all source languages could be used. Future work could focus on dynamically determining which source languages are useful and which are not. Furthermore, a multiway, multisource model might be an interesting challenge to tackle.}

\vspace{2 mm}
\compactpara{Post-Editing.} Instead of having a translator translate from scratch, multi-source NMT can be used to generate high quality translations. The translations can then be post-edited, a process that is less labor intensive and cheaper compared to translating from scratch. Multi-source NMT has been used for post-editing where the translated sentence is used as an additional source, leading to improvements \cite{W17-4773}. \REVISE{Multi-source NMT has also been used for system combination, which combines NMT and SMT outputs to improve translation performance \cite{zhou-etal-2017-neural}}.

\vspace{2 mm}
In general, multi-source NMT does not receive much attention because it cannot be used for real-time translation as it involves additional intermediate translation steps. However, it should be possible to develop a system that can flexibly perform multi-source, multi-target as well as single source and single target translation. \REVISE{The system by \citet{firat16b} is such a system, but the adaptability to language pairs other than European languages and multiple (more than two) source languages has not been verified.} Such a model can be flexibly used during online as well as offline translation scenarios.


\section{Multilingualism in Older Paradigms}
\label{sec:history}
One of the long term goals of the MT community has been the development of architectures that can handle more than two languages. There have been approaches to incorporate multilingualism in the older rule-based and classical SMT paradigms. This section gives a brief overview of these attempts and compares them with MNMT. 

\subsection{Rule-Based Machine Translation} 
Although rule-based systems (RBMT) do not attract research interest, we have included this short discussion for completeness with a goal to understanding the connections of MNMT with previous approaches. RBMT systems using an \textit{interlingua} were explored widely in the past. The interlingua is a symbolic semantic, language-independent representation for natural language text \cite{Sgall:1987:MTL:976858.976876}. Two popular interlinguas are UNL \cite{uchida1996unl} and AMR \cite{banarescu-EtAl:2013:LAW7-ID} Different interlinguas have been proposed in various systems like KANT \cite{nyberg1997kant}, UNL, UNITRAN \cite{dorr1987unitran} and DLT \cite{witkam2006dlt}. Language specific analyzers converted language input to interlingua representation, while language specific decoders converted the interlingua representation into another language. To achieve an unambiguous semantic representation, a lot of linguistic analysis had to be performed and many linguistic resources were required. Hence, in practice, most interlingua systems were limited to research systems or translation in specific domains and could not scale to many languages. 

\subsection{Statistical Machine Translation}
Phrase-based SMT (PBSMT) systems \cite{koehn2003statistical}, a very successful MT paradigm, were also bilingual for the most part. Compared to RBMT, PBSMT requires less linguistic resources and instead requires parallel corpora. However, like RBMT, they work with symbolic, discrete representations making multilingual representation difficult. Moreover, the central unit in PBSMT is the \textit{phrase}, an ordered sequence of words (not in the linguistic sense). Given its arbitrary structure, it is not clear how to build a common symbolic representation for phrases across languages. Nevertheless, some shallow forms of multilingualism have been explored in the context of: (a) pivot-based SMT, (b) multi-source PBSMT, and (c) SMT involving related languages. 

\vspace{2 mm}
\compactpara{Pivoting.} Popular solutions are: chaining source-pivot and pivot-target systems at decoding \cite{utiyama2007}, training a source-target system using synthetic data generated using target-pivot and pivot-source systems \cite{de2006catalanenglish}, and phrase-table triangulation pivoting source-pivot and pivot-target phrase tables \cite{utiyama2007,wu2007pivot}. 

\vspace{2 mm}
\compactpara{Multi-source.}  Typical approaches are: re-ranking outputs from independent source-target systems \cite{och2001statistical}, composing a new output from independent source-target outputs \cite{matusov2006computing}, and translating a combined input representation of multiple sources using lattice networks over multiple phrase tables \cite{schroeder2009word}. 


\vspace{2 mm}
\compactpara{Related languages.} For multilingual translation with multiple related source languages, the typical approaches involved script unification by mapping to a common script such as Devanagari \cite{banerjee2018multilingual} or transliteration \cite{nakov2009improved}. Lexical similarity was utilized using subword-level translation models \cite{vilar2007can,tiedemann2012character,kunchukuttan2016orthographic,kunchukuttan2017bpe}. Combining subword-level representation and pivoting for translation among related languages has been explored \citep{henriquez2011pivot,tiedemann2012character,kunchukuttan2017pivot}. Most of the above mentioned multilingual systems involved either decoding-time operations, chaining black-box systems or composing new phrase-tables from existing ones.

\vspace{2 mm}
\compactpara{Comparison with MNMT.}
While symbolic representations constrain a unified multilingual representation, distributed universal language representation using real-valued vector spaces makes multilingualism easier to implement in NMT. As no language specific feature engineering is required for NMT,  it is possible to easily scale to multiple languages. Neural networks provide flexibility in experimenting with a wide variety of architectures, while advances in optimization techniques and availability of deep learning toolkits make prototyping faster. 

\section{Datasets and Resources}
\label{sec:data}

MNMT requires parallel corpora in similar domains across multiple languages. The following publicly available data-sets can be used for research.

\vspace{2 mm}
\compactpara{Multiway.} Commonly used publicly available multilingual parallel corpora are the TED corpus \cite{mauro2012wit3} (the TED corpus is also available in speech \cite{gangi19}), UN Corpus \cite{ziemski2016united} and those from the European Union like Europarl, JRC-Aquis, DGT-Aquis, DGT-TM, ECDC-TM, EAC-TM \cite{steinberger2014overview}. While these sources are primarily comprised of European languages, parallel corpora for some Asian languages is accessible through the WAT shared task \cite{nakazawa2018overview}. Only small amount of parallel corpora are available for many languages, primarily from movie subtitles and software localization strings \cite{TIEDEMANN12.463}. \REVISE{Recently, the large-scale WikiMatrix corpus \cite{DBLP:journals/corr/abs-1907-05791} and the JW300 corpus covering 300 low-resource languages \cite{agic-vulic-2019-jw300} have been released, which could be good resources for multiway study.}

\vspace{2 mm}
\compactpara{Low or Zero-Resource.} For low or zero-resource NMT translation tasks, good test sets are required for evaluating translation quality. The above mentioned multilingual parallel corpora can be a source for such test sets. In addition, there are other small parallel datasets like the FLORES dataset for English-\{Nepali,Sinhala\} \cite{guzman2019flores}, the XNLI test set spanning 15 languages \cite{conneau2018xnli} and the Indic parallel corpus \cite{birch2011indic}. The WMT shared tasks \cite{bojar2018wmtfindings} also provide test sets for some low-resource language pairs. 

\vspace{2 mm}
\compactpara{Multi-Source.} The corpora for multi-source NMT have to be aligned across languages. Multi-source corpora can be extracted from some of the above mentioned sources. The following are widely used for evaluation in the literature: Europarl \cite{koehn2005epc}, TED \cite{TIEDEMANN12.463}, UN \cite{ZIEMSKI16.1195}. The Indian Language Corpora Initiative (ILCI) corpus \cite{jha2010tdil} is a 11-way parallel corpus of Indian languages along with English. The Asian Language Treebank \cite{thu2016introducing} is a 9-way
parallel corpus of South-East Asian languages along with English, Japanese and Bengali. The MMCR4NLP project \cite{dabre2017mmcr4nlp} compiles language family grouped multi-source corpora and provides standard splits.
\REVISE{The Bible corpus \cite{Christodouloupoulos2015} is a massive N-way one containing 100 languages.}

\vspace{2 mm}
\compactpara{Shared Tasks.} Recently, shared tasks with a focus on multilingual translation have been conducted at IWSLT \cite{cettolo2017overview}, WAT \cite{nakazawa2018overview} and WMT \cite{bojar2018wmtfindings}; so common benchmarks are available.




\section{Connections with Domain Adaptation}


\label{sec:domain}
High quality parallel corpora are limited to specific domains.
Both, vanilla SMT and NMT perform poorly for domain specific translation in low-resource scenarios \cite{duh-EtAl:2013:Short,koehn-knowles:2017:NMT}.
Leveraging out-of-domain parallel corpora and in-domain monolingual corpora for in-domain translation is known as domain adaptation for MT \cite{C18-1111}. 

As we can treat each domain as a language, there are many similarities and common approaches between MNMT and domain adaptation for NMT. 
Therefore, similar to MNMT, when using out-of-domain parallel corpora for domain adaptation, multi-domain NMT and transfer learning based approaches \cite{P17-2061} have been proposed for domain adaptation.
When using in-domain monolingual corpora, a typical way of doing domain adaptation is generating a pseduo-parallel corpus by back-translating target in-domain monolingual corpora \cite{sennrich-haddow-birch:2016:P16-11}, which is similar to the pseduo-parallel corpus generation in MNMT \cite{firat16b}.

There are also many differences between MNMT and domain adaptation for NMT. While pivoting is a popular approach for MNMT \cite{ijcai2017-555}, it is unsuitable for domain adaptation.
As there are always vocabulary overlaps between different domains, there are no zero-shot translation \cite{johnson17} settings in domain adaptation. In addition, it not uncommon to write domain specific sentences in different styles and so
multi-source approaches \cite{N16-1004} are not applicable either. 
On the other hand, data selection approaches in domain adaptation that select out-of-domain sentences which are similar to in-domain sentences \cite{wang-EtAl:2017:Short3} have not been applied to MNMT. In addition, instance weighting approaches \cite{iwnmt}  that interpolate in-domain and out-of-domain models have not been studied for MNMT. However, with the development of cross-lingual sentence embeddings, data selection and instance weighting approaches might be applicable for MNMT in the near future.

\REVISE{There are also studies trying to connect MNMT and domain adaptation. These studies focus on applying or improving fine-tuning for both MNMT and domain adaptation. Chu and Dabre \cite{chu:2018:NLP2018} conducted a preliminary study for this topic, where they transfer knowledge from both multilingual and out-of-domain corpora to improve in-domain translation for a particular language pair via fine-tuning. Chu and Dabre \cite{chu-dabre-2019} focused on training a single translation model for multiple domains by either learning domain specialized hidden state representations or predictor biases for each domain, and incorporate multilingualism into the domain adaptation framework. Dabre et al. \cite{dabre-etal-2019-exploiting} applied multi-state fine-tuning on multiway MNMT, which has been shown effective in domain adaptation by \cite{P17-2061}. Bapna and Firat \cite{bapna-firat-2019-simple} improved the scalability of fune-tuning for both MNMT and domain adaptation. Instead of fine-tuning the entire NMT system, they propose using light-weight adapter layers that are suitable for the target task. Given the these success of connecting MNMT and domain adaptation with fine-tuning, we believe that there is potensial to connect them with other approaches given their similarities and differences discussed above.}





\begin{table}[t]
\centering
\resizebox{\textwidth}{!}{%
\begin{tabular}{|l|l|}
\hline
\textbf{Central Goal} & \textbf{Possible Directions/Issues/Approaches}\\\hline
\textbf{\begin{tabular}[c]{@{}l@{}}Language Representation Learning\end{tabular}} & \begin{tabular}[c]{@{}l@{}}1. Balancing the sharing of representations between languages. \\ 2. Handling language divergence. \\ 3. Addressing code-switching and dialects. \\ 4. Identifying language families computationally.\end{tabular} \\ \hline
\textbf{Leveraging Pretrained Models} & \begin{tabular}[c]{@{}l@{}}1. Pretrained BERT/ GPT/ Transformer XL encoders and decoders. \\ 2. Incorporating web-level knowledge into translation process. \\ 3. Designing pre-training objectives for multilingualism. \\ 4. Dealing with large model sizes. \\ 5. Universal parent (pre-trained) models.\end{tabular} \\ \hline
\textbf{One Model For All Languages} & \begin{tabular}[c]{@{}l@{}}1. A single model for all languages, domains, dialects and code-switching. \\ 2. Possible improvement from multi-modal knowledge. \\ 3. A model to explain multilingualism. \\ 4. Handling capacity bottleneck.\end{tabular} \\ \hline
\end{tabular}%
}
\caption{An overview future research directions and issues for MNMT}
\label{tab:futureoverview}
\end{table}

\section{Future Research Directions}
\label{sec:future}


While exciting advances have been made in MNMT in recent years, there are still many interesting directions for exploration. Refer to Table~\ref{tab:futureoverview} for possible future avenues for MNMT research. The following research directions are by no means exhaustive and are to be considered as guidelines for researchers wishing to address specific important problems in MNMT. Throughout the paper we have identified three major challenges as follows: finding the balance between language agnostic and language aware representations, improving transfer learning, and developing a single NMT model for all possible language pairs. \REVISE{The following topics, starting from the most promising, should help to further tackle these aforementioned challenges.}

\vspace{2 mm}
\compactpara{Exploring Pre-trained Models.}
Pre-training embeddings, encoders and decoders have been shown to be useful for NMT \cite{ramachandran17}. Most works rely on fine-tuning but do not address techniques to maximize the impact of transfer. Furthermore, how pre-training can be incorporated into different MNMT architectures, is important as well. Recent advances in cross-lingual word \cite{klementiev12a,mikolov13a,chandar14a,artetxe16a,conneau18a,jawanpuria2018learning} and sentence embeddings\footnote{\REVISE{https://engineering.fb.com/ai-research/laser-multilingual-sentence-embeddings/}} \cite{conneau2018xnli,chen2018multilingual,artetxe2019multilingual,8070942} could provide directions for this line of investigation. Currently, transfer learning through unsupervised pre-training on extremely large corpora and unsupervised NMT are gaining momentum and we believe that investing into these two topics or a merger between them will yield powerful insights into ways to incorporate large amounts of knowledge into translation systems. \REVISE{In recent times this research topic is has shown to have a lot of promise towards improving bilingual MT quality and should be beneficial for MNMT as well.} 

\vspace{2mm}
\REVISE{\compactpara{Unseen Language Pair Translation.} Most work on unseen language pair translation has only addressed cases where the pivot language is related to or shares the same script with the source language. In many cases, the pivot language (mostly English) is unlikely to be related to the source and target languages and this scenario requires further investigation (especially for zeroshot translation). Moreover, new approaches need to be explored to significantly improve over the simple pivot baseline.}

\vspace{2 mm}
\compactpara{Related Languages, Language Registers and Dialects.}
Translation involving related languages, language registers and dialects can be further explored given the existing evidence about the importance of language relatedness for improving MNMT quality. \REVISE{For dialects, currently} the focus is on modern standard dialects of languages but most people feel comfortable speaking in their local dialect and hence it would be valuable to transfer translation knowledge obtained for modern standard languages to their dialects. In our opinion, unsupervised MT methods using monolingual data should be extremely effective at handling dialect translation.  

\vspace{2 mm}
\compactpara{Code-Mixed Language.}
Addressing intra-sentence multilingualism \textit{i.e.} code mixed input and output, creoles and pidgins is an interesting research direction as they are a variation of related languages translation. The compact MNMT models can handle code-mixed input, but code-mixed output remains an open problem \cite{johnson17}. Code-mixed languages are gradually evolving into languages with their own unique grammar and just like dialects, a vast majority of the people feel more comfortable with code-mixing. For this, we believe that researchers should first focus on identifying code-mixing phenomena (either linguistically or statistically) and then design multilingual solutions to address them.

\vspace{2 mm}
\noindent \REVISE{\compactpara{Visualization and Model Inspection.} Most works focus on an extrinsic evaluation of multilingual models. While the visualization of the embeddings of NMT models in multilingual settings has revealed some interesting behaviours \cite{johnson17}, we feel that more researchers should pursue this. Furthermore, we believe that visualization that relies on down-projecting high dimensional vectors to 2-D and 3-D might not reveal actual phenomenon and lead to certain biases that affect MNMT modeling. It will be interesting to have works that perform an analyses of high dimensional representations and reveal the impact of multilingualism.}

\vspace{2 mm}
\compactpara{Learning Effective Language Representations.}
A core question that \REVISE{still} needs further investigation is: what is the degree of language awareness that needs to be considered to learn task-effective continuous space representations? Particularly, the questions of word-order divergence between the source languages and variable length encoder representations have received little attention. Addressing this issue will eliminate the need to deal with language divergence issues for transfer learning. Most works tend to agree that language agnostic encoders and language aware decoders tend to work well but a deeper investigation is still needed.

\vspace{2 mm}
\compactpara{Multiple Target Language MNMT.} Most current efforts address multiple source languages. 
\REVISE{Multiway systems focusing on multiple low-resource target languages is important not only because it is a significantly harder challenge but because it can help shed more light on language divergence and its impact on translation quality.} The right balance between sharing representations \textit{vs.} maintaining the distinctiveness of the target language for generation needs exploring. Furthermore, depending on the similarity between languages or lack thereof, it might be important to consider a hybrid architecture (mixing RNN, CNN and FFN) for best performance.

\vspace{2 mm}
\noindent \REVISE{\compactpara{Capacity Bottleneck.} Recent works on massively multilingual NMT have shown that existing approaches are unable to successfully incorporate more than 100 language pairs and at the same time improve translation quality. This is not a problem of the number of layers, encoders or decoders in the NMT model but rather a problem of the way the MNMT model is currently designed. Current approaches focus on either separate encoders or decoders lead to a large unwieldy model or shared encoders and decoders with a large number of parameters both of which are hard to train. Works that leverage incremental learning or knowledge distillation seem to be promising and we refer interested researchers to such works.}

\vspace{2 mm}
\compactpara{Joint Multilingual and Multi-Domain NMT.}
Jointly tackling multilingual and multi-domain translation is an interesting direction with many practical use cases. When extending an NMT system to a new language, the parallel corpus in the domain of interest may not be available. Transfer learning in this case has to span languages and domains. It might  be worthwhile to explore adversarial approaches where domain and language invariant representations can be learned so that the low-resource (or zero resource) language pairs and domains can benefit vastly.

\vspace{2 mm}
\compactpara{Multilingual Speech-to-Speech NMT.}
Most work on MNMT, and NMT in general, has focussed on translating text. Recently, progress has been made in speech translation as well as in multilingual ASR and TTS. An interesting research direction would be to explore multilingual speech translation, where the ASR, translation and TTS modules can be multilingual. Interesting challenges and opportunities may arise in the quest to compose all these multilingual systems in an end-to-end method. Multilingual end-to-end speech-to-speech translation would also be a future challenging scenario. Some datasets are already available for such research \cite{gangi19}. \REVISE{We believe that this will be one of the hardest challenges in the field of multilingual translation.}



\section{Conclusion}
\label{sec:conlusion}
MNMT has made rapid progress in recent years. In this survey, we have covered literature pertaining to the major scenarios we identified for multilingual NMT: multiway, low or zero-resource (transfer learning, pivoting, and zero-shot approaches) and multi-source translation. We have systematically compiled the principal design approaches and their variants, central MNMT issues and their proposed solutions along with their strengths and weaknesses. We have put MNMT in a historical perspective with respect to work on multilingual RBMT and SMT systems. \REVISE{At the outset, we raised two important research questions which we feel can be answered to a certain extent as below:}

\begin{description}
\item[Q1.] \REVISE{\emph{Is it possible to have a one-model-for-all-languages solution to MT (NLP) applications?}}
\item[A1.] \REVISE{\emph{Yes. While it is possible to fit all language pairs into a single model, existing deep learning methodologies suffer from capacity bottlenecks and generalization capabilities which puts a limit on the gains from multilingualism on translation quality. Further research into better data selection and representation, network architectures and learning algorithms is needed.}}
\item[Q2.] \REVISE{\emph{Can shared multilingual distributed representations help MT (NLP) for low-resource languages?}}
\item[A2.] \REVISE{\emph{Yes. Low-resource language translation benefits significantly as a result of multilingually learned continuous space distributions but further research on balancing the language agnosticity and specificity of representations can help push performance even further.}}
\end{description}

It is clear that multilingualism can not only improve translation quality when leveraged but also can provide a number of insights into the relationships between languages. \REVISE{Most works lack human evaluation of multilingual models which should help in better understanding of the impact of multilingualism.} We  suggest promising and important directions for future work. We hope that this survey paper will give researchers and practitioners a deeper understanding of the MNMT landscape and enable them to choose promising topics for research. We hope that our work will help promote and accelerate MNMT research.


\bibliographystyle{ACM-Reference-Format}
\bibliography{main}


\begin{thebibliography}{156}


\ifx \showCODEN    \undefined \def \showCODEN     #1{\unskip}     \fi
\ifx \showDOI      \undefined \def \showDOI       #1{#1}\fi
\ifx \showISBNx    \undefined \def \showISBNx     #1{\unskip}     \fi
\ifx \showISBNxiii \undefined \def \showISBNxiii  #1{\unskip}     \fi
\ifx \showISSN     \undefined \def \showISSN      #1{\unskip}     \fi
\ifx \showLCCN     \undefined \def \showLCCN      #1{\unskip}     \fi
\ifx \shownote     \undefined \def \shownote      #1{#1}          \fi
\ifx \showarticletitle \undefined \def \showarticletitle #1{#1}   \fi
\ifx \showURL      \undefined \def \showURL       {\relax}        \fi
\providecommand\bibfield[2]{#2}
\providecommand\bibinfo[2]{#2}
\providecommand\natexlab[1]{#1}
\providecommand\showeprint[2][]{arXiv:#2}

\bibitem[\protect\citeauthoryear{Agi{\'c} and Vuli{\'c}}{Agi{\'c} and
  Vuli{\'c}}{2019}]%
        {agic-vulic-2019-jw300}
\bibfield{author}{\bibinfo{person}{{\v{Z}}eljko Agi{\'c}} {and}
  \bibinfo{person}{Ivan Vuli{\'c}}.} \bibinfo{year}{2019}\natexlab{}.
\newblock \showarticletitle{{JW}300: A Wide-Coverage Parallel Corpus for
  Low-Resource Languages}. In \bibinfo{booktitle}{\emph{Proceedings of the 57th
  Annual Meeting of the Association for Computational Linguistics}}.
  \bibinfo{publisher}{Association for Computational Linguistics},
  \bibinfo{address}{Florence, Italy}, \bibinfo{pages}{3204--3210}.
\newblock
\urldef\tempurl%
\url{https://doi.org/10.18653/v1/P19-1310}
\showDOI{\tempurl}


\bibitem[\protect\citeauthoryear{Aharoni, Johnson, and Firat}{Aharoni
  et~al\mbox{.}}{2019}]%
        {aharoni19}
\bibfield{author}{\bibinfo{person}{Roee Aharoni}, \bibinfo{person}{Melvin
  Johnson}, {and} \bibinfo{person}{Orhan Firat}.}
  \bibinfo{year}{2019}\natexlab{}.
\newblock \showarticletitle{Massively Multilingual Neural Machine Translation}.
  In \bibinfo{booktitle}{\emph{Proceedings of the 2019 Conference of the North
  {A}merican Chapter of the Association for Computational Linguistics: Human
  Language Technologies, Volume 1 (Long and Short Papers)}}.
  \bibinfo{publisher}{Association for Computational Linguistics},
  \bibinfo{address}{Minneapolis, Minnesota}, \bibinfo{pages}{3874--3884}.
\newblock
\urldef\tempurl%
\url{https://www.aclweb.org/anthology/N19-1388}
\showURL{%
\tempurl}


\bibitem[\protect\citeauthoryear{Al-Shedivat and Parikh}{Al-Shedivat and
  Parikh}{2019}]%
        {al-shedivat-parikh-2019-consistency}
\bibfield{author}{\bibinfo{person}{Maruan Al-Shedivat} {and}
  \bibinfo{person}{Ankur Parikh}.} \bibinfo{year}{2019}\natexlab{}.
\newblock \showarticletitle{Consistency by Agreement in Zero-Shot Neural
  Machine Translation}. In \bibinfo{booktitle}{\emph{Proceedings of the 2019
  Conference of the North {A}merican Chapter of the Association for
  Computational Linguistics: Human Language Technologies, Volume 1 (Long and
  Short Papers)}}. \bibinfo{publisher}{Association for Computational
  Linguistics}, \bibinfo{address}{Minneapolis, Minnesota},
  \bibinfo{pages}{1184--1197}.
\newblock
\urldef\tempurl%
\url{https://www.aclweb.org/anthology/N19-1121}
\showURL{%
\tempurl}


\bibitem[\protect\citeauthoryear{Arivazhagan, Bapna, Firat, Aharoni, Johnson,
  and Macherey}{Arivazhagan et~al\mbox{.}}{2019a}]%
        {arivazhagan2018missing}
\bibfield{author}{\bibinfo{person}{Naveen Arivazhagan}, \bibinfo{person}{Ankur
  Bapna}, \bibinfo{person}{Orhan Firat}, \bibinfo{person}{Roee Aharoni},
  \bibinfo{person}{Melvin Johnson}, {and} \bibinfo{person}{Wolfgang Macherey}.}
  \bibinfo{year}{2019}\natexlab{a}.
\newblock \showarticletitle{The Missing Ingredient in Zero-Shot Neural Machine
  Translation}.
\newblock \bibinfo{journal}{\emph{CoRR}}  \bibinfo{volume}{abs/1903.07091}
  (\bibinfo{year}{2019}).
\newblock
\showeprint[arxiv]{1903.07091}
\urldef\tempurl%
\url{http://arxiv.org/abs/1903.07091}
\showURL{%
\tempurl}


\bibitem[\protect\citeauthoryear{Arivazhagan, Bapna, Firat, Lepikhin, Johnson,
  Krikun, Chen, Cao, Foster, Cherry, Macherey, Chen, and Wu}{Arivazhagan
  et~al\mbox{.}}{2019b}]%
        {wild}
\bibfield{author}{\bibinfo{person}{Naveen Arivazhagan}, \bibinfo{person}{Ankur
  Bapna}, \bibinfo{person}{Orhan Firat}, \bibinfo{person}{Dmitry Lepikhin},
  \bibinfo{person}{Melvin Johnson}, \bibinfo{person}{Maxim Krikun},
  \bibinfo{person}{Mia~Xu Chen}, \bibinfo{person}{Yuan Cao},
  \bibinfo{person}{George Foster}, \bibinfo{person}{Colin Cherry},
  \bibinfo{person}{Wolfgang Macherey}, \bibinfo{person}{Zhifeng Chen}, {and}
  \bibinfo{person}{Yonghui Wu}.} \bibinfo{year}{2019}\natexlab{b}.
\newblock \showarticletitle{Massively Multilingual Neural Machine Translation
  in the Wild: Findings and Challenges}.
\newblock \bibinfo{journal}{\emph{CoRR}}  \bibinfo{volume}{abs/1907.05019}
  (\bibinfo{year}{2019}).
\newblock
\showeprint[arxiv]{1907.05019}
\urldef\tempurl%
\url{http://arxiv.org/abs/1907.05019}
\showURL{%
\tempurl}


\bibitem[\protect\citeauthoryear{Artetxe, Labaka, and Agirre}{Artetxe
  et~al\mbox{.}}{2016}]%
        {artetxe16a}
\bibfield{author}{\bibinfo{person}{Mikel Artetxe}, \bibinfo{person}{Gorka
  Labaka}, {and} \bibinfo{person}{Eneko Agirre}.}
  \bibinfo{year}{2016}\natexlab{}.
\newblock \showarticletitle{Learning principled bilingual mappings of word
  embeddings while preserving monolingual invariance}. In
  \bibinfo{booktitle}{\emph{Proceedings of the 2016 Conference on Empirical
  Methods in Natural Language Processing}}. \bibinfo{publisher}{Association for
  Computational Linguistics}, \bibinfo{address}{Austin, Texas},
  \bibinfo{pages}{2289--2294}.
\newblock
\urldef\tempurl%
\url{https://doi.org/10.18653/v1/D16-1250}
\showDOI{\tempurl}


\bibitem[\protect\citeauthoryear{Artetxe and Schwenk}{Artetxe and
  Schwenk}{2018}]%
        {artetxe2019multilingual}
\bibfield{author}{\bibinfo{person}{Mikel Artetxe} {and} \bibinfo{person}{Holger
  Schwenk}.} \bibinfo{year}{2018}\natexlab{}.
\newblock \showarticletitle{Massively Multilingual Sentence Embeddings for
  Zero-Shot Cross-Lingual Transfer and Beyond}.
\newblock \bibinfo{journal}{\emph{CoRR}}  \bibinfo{volume}{abs/1812.10464}
  (\bibinfo{year}{2018}).
\newblock
\showeprint[arxiv]{1812.10464}
\urldef\tempurl%
\url{http://arxiv.org/abs/1812.10464}
\showURL{%
\tempurl}


\bibitem[\protect\citeauthoryear{Bahdanau, Cho, and Bengio}{Bahdanau
  et~al\mbox{.}}{2015}]%
        {bahdanau15}
\bibfield{author}{\bibinfo{person}{Dzmitry Bahdanau},
  \bibinfo{person}{Kyunghyun Cho}, {and} \bibinfo{person}{Yoshua Bengio}.}
  \bibinfo{year}{2015}\natexlab{}.
\newblock \showarticletitle{Neural Machine Translation by Jointly Learning to
  Align and Translate}. In \bibinfo{booktitle}{\emph{In Proceedings of the 3rd
  International Conference on Learning Representations (ICLR 2015)}}.
  \bibinfo{publisher}{International Conference on Learning Representations},
  \bibinfo{address}{San Diego, USA}.
\newblock


\bibitem[\protect\citeauthoryear{Banarescu, Bonial, Cai, Georgescu, Griffitt,
  Hermjakob, Knight, Koehn, Palmer, and Schneider}{Banarescu
  et~al\mbox{.}}{2013}]%
        {banarescu-EtAl:2013:LAW7-ID}
\bibfield{author}{\bibinfo{person}{Laura Banarescu}, \bibinfo{person}{Claire
  Bonial}, \bibinfo{person}{Shu Cai}, \bibinfo{person}{Madalina Georgescu},
  \bibinfo{person}{Kira Griffitt}, \bibinfo{person}{Ulf Hermjakob},
  \bibinfo{person}{Kevin Knight}, \bibinfo{person}{Philipp Koehn},
  \bibinfo{person}{Martha Palmer}, {and} \bibinfo{person}{Nathan Schneider}.}
  \bibinfo{year}{2013}\natexlab{}.
\newblock \showarticletitle{Abstract Meaning Representation for Sembanking}. In
  \bibinfo{booktitle}{\emph{Proceedings of the 7th Linguistic Annotation
  Workshop and Interoperability with Discourse}}.
  \bibinfo{publisher}{Association for Computational Linguistics},
  \bibinfo{address}{Sofia, Bulgaria}, \bibinfo{pages}{178--186}.
\newblock
\urldef\tempurl%
\url{http://www.aclweb.org/anthology/W13-2322}
\showURL{%
\tempurl}


\bibitem[\protect\citeauthoryear{Banerjee, Kunchukuttan, and
  Bhattacharyya}{Banerjee et~al\mbox{.}}{2018}]%
        {banerjee2018multilingual}
\bibfield{author}{\bibinfo{person}{Tamali Banerjee}, \bibinfo{person}{Anoop
  Kunchukuttan}, {and} \bibinfo{person}{Pushpak Bhattacharyya}.}
  \bibinfo{year}{2018}\natexlab{}.
\newblock \showarticletitle{{Multilingual Indian Language Translation System at
  {WAT} 2018: Many-to-one Phrase-based SMT}}. In
  \bibinfo{booktitle}{\emph{Proceedings of the 5th Workshop on Asian Language
  Translation}}. \bibinfo{address}{Hong Kong, China}.
\newblock


\bibitem[\protect\citeauthoryear{Bapna and Firat}{Bapna and Firat}{2019}]%
        {bapna-firat-2019-simple}
\bibfield{author}{\bibinfo{person}{Ankur Bapna} {and} \bibinfo{person}{Orhan
  Firat}.} \bibinfo{year}{2019}\natexlab{}.
\newblock \showarticletitle{Simple, Scalable Adaptation for Neural Machine
  Translation}. In \bibinfo{booktitle}{\emph{Proceedings of the 2019 Conference
  on Empirical Methods in Natural Language Processing and the 9th International
  Joint Conference on Natural Language Processing (EMNLP-IJCNLP)}}.
  \bibinfo{publisher}{Association for Computational Linguistics},
  \bibinfo{address}{Hong Kong, China}, \bibinfo{pages}{1538--1548}.
\newblock
\urldef\tempurl%
\url{https://doi.org/10.18653/v1/D19-1165}
\showDOI{\tempurl}


\bibitem[\protect\citeauthoryear{Birch, Callison-Burch, Osborne, and
  Post}{Birch et~al\mbox{.}}{2011}]%
        {birch2011indic}
\bibfield{author}{\bibinfo{person}{Lexi Birch}, \bibinfo{person}{Chris
  Callison-Burch}, \bibinfo{person}{Miles Osborne}, {and} \bibinfo{person}{Matt
  Post}.} \bibinfo{year}{2011}\natexlab{}.
\newblock \bibinfo{title}{The Indic multi-parallel corpus}.
\newblock
  \bibinfo{howpublished}{\url{http://homepages.inf.ed.ac.uk/miles/babel.html}}.
\newblock


\bibitem[\protect\citeauthoryear{Blackwood, Ballesteros, and Ward}{Blackwood
  et~al\mbox{.}}{2018}]%
        {blackwood18}
\bibfield{author}{\bibinfo{person}{Graeme Blackwood}, \bibinfo{person}{Miguel
  Ballesteros}, {and} \bibinfo{person}{Todd Ward}.}
  \bibinfo{year}{2018}\natexlab{}.
\newblock \showarticletitle{Multilingual Neural Machine Translation with
  Task-Specific Attention}. In \bibinfo{booktitle}{\emph{Proceedings of the
  27th International Conference on Computational Linguistics}}.
  \bibinfo{publisher}{Association for Computational Linguistics},
  \bibinfo{address}{Santa Fe, New Mexico, USA}, \bibinfo{pages}{3112--3122}.
\newblock
\urldef\tempurl%
\url{http://aclweb.org/anthology/C18-1263}
\showURL{%
\tempurl}


\bibitem[\protect\citeauthoryear{Bojar, Chatterjee, Federmann, Graham, Haddow,
  Huang, Huck, Koehn, Liu, Logacheva, Monz, Negri, Post, Rubino, Specia, and
  Turchi}{Bojar et~al\mbox{.}}{2017}]%
        {bojar-EtAl:2017:WMT1}
\bibfield{author}{\bibinfo{person}{Ond\v{r}ej Bojar}, \bibinfo{person}{Rajen
  Chatterjee}, \bibinfo{person}{Christian Federmann}, \bibinfo{person}{Yvette
  Graham}, \bibinfo{person}{Barry Haddow}, \bibinfo{person}{Shujian Huang},
  \bibinfo{person}{Matthias Huck}, \bibinfo{person}{Philipp Koehn},
  \bibinfo{person}{Qun Liu}, \bibinfo{person}{Varvara Logacheva},
  \bibinfo{person}{Christof Monz}, \bibinfo{person}{Matteo Negri},
  \bibinfo{person}{Matt Post}, \bibinfo{person}{Raphael Rubino},
  \bibinfo{person}{Lucia Specia}, {and} \bibinfo{person}{Marco Turchi}.}
  \bibinfo{year}{2017}\natexlab{}.
\newblock \showarticletitle{Findings of the 2017 Conference on Machine
  Translation ({WMT}17)}. In \bibinfo{booktitle}{\emph{Proceedings of the
  Second Conference on Machine Translation}}. \bibinfo{publisher}{Association
  for Computational Linguistics}, \bibinfo{address}{Copenhagen, Denmark},
  \bibinfo{pages}{169--214}.
\newblock
\urldef\tempurl%
\url{http://www.aclweb.org/anthology/W17-4717}
\showURL{%
\tempurl}


\bibitem[\protect\citeauthoryear{Bojar, Federmann, Fishel, Graham, Haddow,
  Koehn, and Monz}{Bojar et~al\mbox{.}}{2018}]%
        {bojar2018wmtfindings}
\bibfield{author}{\bibinfo{person}{Ond{\v{r}}ej Bojar},
  \bibinfo{person}{Christian Federmann}, \bibinfo{person}{Mark Fishel},
  \bibinfo{person}{Yvette Graham}, \bibinfo{person}{Barry Haddow},
  \bibinfo{person}{Philipp Koehn}, {and} \bibinfo{person}{Christof Monz}.}
  \bibinfo{year}{2018}\natexlab{}.
\newblock \showarticletitle{Findings of the 2018 Conference on Machine
  Translation ({WMT}18)}. In \bibinfo{booktitle}{\emph{Proceedings of the Third
  Conference on Machine Translation: Shared Task Papers}}.
  \bibinfo{publisher}{Association for Computational Linguistics},
  \bibinfo{address}{Belgium, Brussels}, \bibinfo{pages}{272--303}.
\newblock
\urldef\tempurl%
\url{http://aclweb.org/anthology/W18-6401}
\showURL{%
\tempurl}


\bibitem[\protect\citeauthoryear{Cettolo, Federico, Bentivogli, Niehues,
  Stüker, Sudoh, Yoshino, and Federmann}{Cettolo et~al\mbox{.}}{2017}]%
        {cettolo2017overview}
\bibfield{author}{\bibinfo{person}{Mauro Cettolo}, \bibinfo{person}{Marcello
  Federico}, \bibinfo{person}{Luisa Bentivogli}, \bibinfo{person}{Jan Niehues},
  \bibinfo{person}{Sebastian Stüker}, \bibinfo{person}{Katsuhito Sudoh},
  \bibinfo{person}{Koichiro Yoshino}, {and} \bibinfo{person}{Christian
  Federmann}.} \bibinfo{year}{2017}\natexlab{}.
\newblock \showarticletitle{Overview of the {IWSLT} 2017 Evaluation Campaign}.
  In \bibinfo{booktitle}{\emph{Proceedings of the 14th International Workshop
  on Spoken Language Translation}}. \bibinfo{address}{Tokyo, Japan},
  \bibinfo{pages}{2--14}.
\newblock


\bibitem[\protect\citeauthoryear{Chandar, Lauly, Larochelle, Khapra, Ravindran,
  Raykar, and Saha}{Chandar et~al\mbox{.}}{2014}]%
        {chandar14a}
\bibfield{author}{\bibinfo{person}{Sarath Chandar}, \bibinfo{person}{Stanislas
  Lauly}, \bibinfo{person}{Hugo Larochelle}, \bibinfo{person}{Mitesh Khapra},
  \bibinfo{person}{Balaraman Ravindran}, \bibinfo{person}{Vikas~C Raykar},
  {and} \bibinfo{person}{Amrita Saha}.} \bibinfo{year}{2014}\natexlab{}.
\newblock \showarticletitle{An autoencoder approach to learning bilingual word
  representations}. In \bibinfo{booktitle}{\emph{Proceedings of the Advances in
  Neural Information Processing Systems}}. \bibinfo{address}{Montréal,
  Canada}, \bibinfo{pages}{1853--1861}.
\newblock


\bibitem[\protect\citeauthoryear{Chatterjee, Farajian, Negri, Turchi,
  Srivastava, and Pal}{Chatterjee et~al\mbox{.}}{2017}]%
        {W17-4773}
\bibfield{author}{\bibinfo{person}{Rajen Chatterjee}, \bibinfo{person}{M.~Amin
  Farajian}, \bibinfo{person}{Matteo Negri}, \bibinfo{person}{Marco Turchi},
  \bibinfo{person}{Ankit Srivastava}, {and} \bibinfo{person}{Santanu Pal}.}
  \bibinfo{year}{2017}\natexlab{}.
\newblock \showarticletitle{Multi-source Neural Automatic Post-Editing: FBK's
  participation in the {WMT} 2017 APE shared task}. In
  \bibinfo{booktitle}{\emph{Proceedings of the Second Conference on Machine
  Translation}}. \bibinfo{publisher}{Association for Computational
  Linguistics}, \bibinfo{address}{Copenhagen, Denmark},
  \bibinfo{pages}{630--638}.
\newblock
\urldef\tempurl%
\url{https://doi.org/10.18653/v1/W17-4773}
\showDOI{\tempurl}


\bibitem[\protect\citeauthoryear{Chaudhary, Dalmia, Hu, Li, Matthews, Muis,
  Otani, Rijhwani, Sheikh, Vyas, Wang, Xie, Xu, Zhou, Jansen, Yang, Levin,
  Metze, Mitamura, Mortensen, Neubig, Hovy, Black, Carbonell, Horwood,
  Tafreshi, Diab, Kayi, Farra, and McKeown}{Chaudhary et~al\mbox{.}}{2019}]%
        {DBLP:journals/corr/Chaudhary19}
\bibfield{author}{\bibinfo{person}{Aditi Chaudhary}, \bibinfo{person}{Siddharth
  Dalmia}, \bibinfo{person}{Junjie Hu}, \bibinfo{person}{Xinjian Li},
  \bibinfo{person}{Austin Matthews}, \bibinfo{person}{Aldrian~Obaja Muis},
  \bibinfo{person}{Naoki Otani}, \bibinfo{person}{Shruti Rijhwani},
  \bibinfo{person}{Zaid Sheikh}, \bibinfo{person}{Nidhi Vyas},
  \bibinfo{person}{Xinyi Wang}, \bibinfo{person}{Jiateng Xie},
  \bibinfo{person}{Ruochen Xu}, \bibinfo{person}{Chunting Zhou},
  \bibinfo{person}{Peter~J. Jansen}, \bibinfo{person}{Yiming Yang},
  \bibinfo{person}{Lori Levin}, \bibinfo{person}{Florian Metze},
  \bibinfo{person}{Teruko Mitamura}, \bibinfo{person}{David~R. Mortensen},
  \bibinfo{person}{Graham Neubig}, \bibinfo{person}{Eduard Hovy},
  \bibinfo{person}{Alan~W Black}, \bibinfo{person}{Jaime Carbonell},
  \bibinfo{person}{Graham~V. Horwood}, \bibinfo{person}{Shabnam Tafreshi},
  \bibinfo{person}{Mona Diab}, \bibinfo{person}{Efsun~S. Kayi},
  \bibinfo{person}{Noura Farra}, {and} \bibinfo{person}{Kathleen McKeown}.}
  \bibinfo{year}{2019}\natexlab{}.
\newblock \showarticletitle{{The ARIEL-CMU Systems for LoReHLT18}}.
\newblock \bibinfo{journal}{\emph{CoRR}}  \bibinfo{volume}{abs/1902.08899}
  (\bibinfo{year}{2019}).
\newblock
\showeprint[arxiv]{1902.08899}
\urldef\tempurl%
\url{https://arxiv.org/abs/1902.08899}
\showURL{%
\tempurl}


\bibitem[\protect\citeauthoryear{Chen, Awadallah, Hassan, Wang, and
  Cardie}{Chen et~al\mbox{.}}{2018a}]%
        {chen2018multilingual}
\bibfield{author}{\bibinfo{person}{Xilun Chen}, \bibinfo{person}{Ahmed~Hassan
  Awadallah}, \bibinfo{person}{Hany Hassan}, \bibinfo{person}{Wei Wang}, {and}
  \bibinfo{person}{Claire Cardie}.} \bibinfo{year}{2018}\natexlab{a}.
\newblock \showarticletitle{Zero-Resource Multilingual Model Transfer: Learning
  What to Share}.
\newblock \bibinfo{journal}{\emph{CoRR}}  \bibinfo{volume}{abs/1810.03552}
  (\bibinfo{year}{2018}).
\newblock
\showeprint[arxiv]{1810.03552}
\urldef\tempurl%
\url{http://arxiv.org/abs/1810.03552}
\showURL{%
\tempurl}


\bibitem[\protect\citeauthoryear{Chen, Liu, Cheng, and Li}{Chen
  et~al\mbox{.}}{2017}]%
        {P17-1176}
\bibfield{author}{\bibinfo{person}{Yun Chen}, \bibinfo{person}{Yang Liu},
  \bibinfo{person}{Yong Cheng}, {and} \bibinfo{person}{Victor~O.K. Li}.}
  \bibinfo{year}{2017}\natexlab{}.
\newblock \showarticletitle{A Teacher-Student Framework for Zero-Resource
  Neural Machine Translation}. In \bibinfo{booktitle}{\emph{Proceedings of the
  55th Annual Meeting of the Association for Computational Linguistics (Volume
  1: Long Papers)}}. \bibinfo{publisher}{Association for Computational
  Linguistics}, \bibinfo{address}{Vancouver, Canada},
  \bibinfo{pages}{1925--1935}.
\newblock
\urldef\tempurl%
\url{https://doi.org/10.18653/v1/P17-1176}
\showDOI{\tempurl}


\bibitem[\protect\citeauthoryear{Chen, Liu, and Li}{Chen
  et~al\mbox{.}}{2018b}]%
        {DBLP:conf/aaai/ChenLL18}
\bibfield{author}{\bibinfo{person}{Yun Chen}, \bibinfo{person}{Yang Liu}, {and}
  \bibinfo{person}{Victor O.~K. Li}.} \bibinfo{year}{2018}\natexlab{b}.
\newblock \showarticletitle{Zero-Resource Neural Machine Translation with
  Multi-Agent Communication Game}. In \bibinfo{booktitle}{\emph{Proceedings of
  the Thirty-Second AAAI Conference on Artificial Intelligence}}.
  \bibinfo{publisher}{{AAAI} Press}, \bibinfo{pages}{5086--5093}.
\newblock


\bibitem[\protect\citeauthoryear{Cheng, Yang, Liu, Sun, and Xu}{Cheng
  et~al\mbox{.}}{2017}]%
        {ijcai2017-555}
\bibfield{author}{\bibinfo{person}{Yong Cheng}, \bibinfo{person}{Qian Yang},
  \bibinfo{person}{Yang Liu}, \bibinfo{person}{Maosong Sun}, {and}
  \bibinfo{person}{Wei Xu}.} \bibinfo{year}{2017}\natexlab{}.
\newblock \showarticletitle{Joint Training for Pivot-based Neural Machine
  Translation}. In \bibinfo{booktitle}{\emph{Proceedings of the Twenty-Sixth
  International Joint Conference on Artificial Intelligence, {IJCAI-17}}}.
  \bibinfo{address}{Melbourne}, \bibinfo{pages}{3974--3980}.
\newblock
\urldef\tempurl%
\url{https://doi.org/10.24963/ijcai.2017/555}
\showDOI{\tempurl}


\bibitem[\protect\citeauthoryear{Cho, van Merrienboer, Bahdanau, and
  Bengio}{Cho et~al\mbox{.}}{2014}]%
        {DBLP:journals/corr/ChoMBB14}
\bibfield{author}{\bibinfo{person}{Kyunghyun Cho}, \bibinfo{person}{Bart van
  Merrienboer}, \bibinfo{person}{Dzmitry Bahdanau}, {and}
  \bibinfo{person}{Yoshua Bengio}.} \bibinfo{year}{2014}\natexlab{}.
\newblock \showarticletitle{On the Properties of Neural Machine Translation:
  Encoder{--}Decoder Approaches}. In \bibinfo{booktitle}{\emph{Proceedings of
  {SSST}-8, Eighth Workshop on Syntax, Semantics and Structure in Statistical
  Translation}}. \bibinfo{publisher}{Association for Computational
  Linguistics}, \bibinfo{address}{Doha, Qatar}, \bibinfo{pages}{103--111}.
\newblock
\urldef\tempurl%
\url{https://doi.org/10.3115/v1/W14-4012}
\showDOI{\tempurl}


\bibitem[\protect\citeauthoryear{Choi, Shin, and Kim}{Choi
  et~al\mbox{.}}{2018}]%
        {choi2018improving}
\bibfield{author}{\bibinfo{person}{Gyu~Hyeon Choi}, \bibinfo{person}{Jong~Hun
  Shin}, {and} \bibinfo{person}{Young~Kil Kim}.}
  \bibinfo{year}{2018}\natexlab{}.
\newblock \showarticletitle{Improving a Multi-Source Neural Machine Translation
  Model with Corpus Extension for Low-Resource Languages}. In
  \bibinfo{booktitle}{\emph{Proceedings of the Eleventh International
  Conference on Language Resources and Evaluation (LREC-2018)}}.
  \bibinfo{publisher}{European Language Resource Association},
  \bibinfo{address}{Miyazaki, Japan}, \bibinfo{pages}{900--904}.
\newblock
\urldef\tempurl%
\url{http://aclweb.org/anthology/L18-1144}
\showURL{%
\tempurl}


\bibitem[\protect\citeauthoryear{Christodouloupoulos and
  Steedman}{Christodouloupoulos and Steedman}{2015}]%
        {Christodouloupoulos2015}
\bibfield{author}{\bibinfo{person}{Christos Christodouloupoulos} {and}
  \bibinfo{person}{Mark Steedman}.} \bibinfo{year}{2015}\natexlab{}.
\newblock \showarticletitle{A massively parallel corpus: the Bible in 100
  languages}.
\newblock \bibinfo{journal}{\emph{Language Resources and Evaluation}}
  \bibinfo{volume}{49} (\bibinfo{year}{2015}), \bibinfo{pages}{375--395}.
\newblock
Issue 2.


\bibitem[\protect\citeauthoryear{Chu and Dabre}{Chu and Dabre}{2018}]%
        {chu:2018:NLP2018}
\bibfield{author}{\bibinfo{person}{Chenhui Chu} {and} \bibinfo{person}{Raj
  Dabre}.} \bibinfo{year}{2018}\natexlab{}.
\newblock \showarticletitle{Multilingual and Multi-Domain Adaptation for Neural
  Machine Translation}. In \bibinfo{booktitle}{\emph{Proceedings of the 24st
  Annual Meeting of the Association for Natural Language Processing (NLP
  2018)}}. \bibinfo{address}{Okayama, Japan}, \bibinfo{pages}{909--912}.
\newblock


\bibitem[\protect\citeauthoryear{Chu and Dabre}{Chu and Dabre}{2019}]%
        {chu-dabre-2019}
\bibfield{author}{\bibinfo{person}{Chenhui Chu} {and} \bibinfo{person}{Raj
  Dabre}.} \bibinfo{year}{2019}\natexlab{}.
\newblock \showarticletitle{Multilingual Multi-Domain Adaptation Approaches for
  Neural Machine Translation}.
\newblock \bibinfo{journal}{\emph{CoRR}}  \bibinfo{volume}{abs/1906.07978}
  (\bibinfo{year}{2019}).
\newblock
\showeprint{1906.07978}
\urldef\tempurl%
\url{https://arxiv.org/abs/1906.07978}
\showURL{%
\tempurl}


\bibitem[\protect\citeauthoryear{Chu, Dabre, and Kurohashi}{Chu
  et~al\mbox{.}}{2017}]%
        {P17-2061}
\bibfield{author}{\bibinfo{person}{Chenhui Chu}, \bibinfo{person}{Raj Dabre},
  {and} \bibinfo{person}{Sadao Kurohashi}.} \bibinfo{year}{2017}\natexlab{}.
\newblock \showarticletitle{An Empirical Comparison of Domain Adaptation
  Methods for Neural Machine Translation}. In
  \bibinfo{booktitle}{\emph{Proceedings of the 55th Annual Meeting of the
  Association for Computational Linguistics (Volume 2: Short Papers)}}.
  \bibinfo{publisher}{Association for Computational Linguistics},
  \bibinfo{address}{Vancouver, Canada}, \bibinfo{pages}{385--391}.
\newblock
\urldef\tempurl%
\url{https://doi.org/10.18653/v1/P17-2061}
\showDOI{\tempurl}


\bibitem[\protect\citeauthoryear{Chu and Wang}{Chu and Wang}{2018}]%
        {C18-1111}
\bibfield{author}{\bibinfo{person}{Chenhui Chu} {and} \bibinfo{person}{Rui
  Wang}.} \bibinfo{year}{2018}\natexlab{}.
\newblock \showarticletitle{A Survey of Domain Adaptation for Neural Machine
  Translation}. In \bibinfo{booktitle}{\emph{Proceedings of the 27th
  International Conference on Computational Linguistics}}.
  \bibinfo{publisher}{Association for Computational Linguistics},
  \bibinfo{address}{Santa Fe, New Mexico, USA}, \bibinfo{pages}{1304--1319}.
\newblock
\urldef\tempurl%
\url{http://aclweb.org/anthology/C18-1111}
\showURL{%
\tempurl}


\bibitem[\protect\citeauthoryear{Conneau, Lample, Ranzato, Denoyer, and
  J{\'e}gou}{Conneau et~al\mbox{.}}{2018a}]%
        {conneau18a}
\bibfield{author}{\bibinfo{person}{Alexis Conneau}, \bibinfo{person}{Guillaume
  Lample}, \bibinfo{person}{Marc'Aurelio Ranzato}, \bibinfo{person}{Ludovic
  Denoyer}, {and} \bibinfo{person}{Herv{\'e} J{\'e}gou}.}
  \bibinfo{year}{2018}\natexlab{a}.
\newblock \showarticletitle{Word Translation Without Parallel Data}. In
  \bibinfo{booktitle}{\emph{Proceedings of the International Conference on
  Learning Representations}}. \bibinfo{address}{Vancouver, BC, Canada}.
\newblock
\newblock
\shownote{{URL}: \url{https://github.com/facebookresearch/MUSE}.}


\bibitem[\protect\citeauthoryear{Conneau, Rinott, Lample, Williams, Bowman,
  Schwenk, and Stoyanov}{Conneau et~al\mbox{.}}{2018b}]%
        {conneau2018xnli}
\bibfield{author}{\bibinfo{person}{Alexis Conneau}, \bibinfo{person}{Ruty
  Rinott}, \bibinfo{person}{Guillaume Lample}, \bibinfo{person}{Adina
  Williams}, \bibinfo{person}{Samuel Bowman}, \bibinfo{person}{Holger Schwenk},
  {and} \bibinfo{person}{Veselin Stoyanov}.} \bibinfo{year}{2018}\natexlab{b}.
\newblock \showarticletitle{{XNLI}: Evaluating Cross-lingual Sentence
  Representations}. In \bibinfo{booktitle}{\emph{Proceedings of the 2018
  Conference on Empirical Methods in Natural Language Processing}}.
  \bibinfo{publisher}{Association for Computational Linguistics},
  \bibinfo{address}{Brussels, Belgium}, \bibinfo{pages}{2475--2485}.
\newblock
\urldef\tempurl%
\url{https://www.aclweb.org/anthology/D18-1269}
\showURL{%
\tempurl}


\bibitem[\protect\citeauthoryear{Currey and Heafield}{Currey and
  Heafield}{2019}]%
        {currey-heafield-2019-zero}
\bibfield{author}{\bibinfo{person}{Anna Currey} {and} \bibinfo{person}{Kenneth
  Heafield}.} \bibinfo{year}{2019}\natexlab{}.
\newblock \showarticletitle{Zero-Resource Neural Machine Translation with
  Monolingual Pivot Data}. In \bibinfo{booktitle}{\emph{Proceedings of the 3rd
  Workshop on Neural Generation and Translation}}.
  \bibinfo{publisher}{Association for Computational Linguistics},
  \bibinfo{address}{Hong Kong}, \bibinfo{pages}{99--107}.
\newblock
\urldef\tempurl%
\url{https://doi.org/10.18653/v1/D19-5610}
\showDOI{\tempurl}


\bibitem[\protect\citeauthoryear{Dabre, Cromieres, and Kurohashi}{Dabre
  et~al\mbox{.}}{2017a}]%
        {Dabre-MTS2017}
\bibfield{author}{\bibinfo{person}{Raj Dabre}, \bibinfo{person}{Fabien
  Cromieres}, {and} \bibinfo{person}{Sadao Kurohashi}.}
  \bibinfo{year}{2017}\natexlab{a}.
\newblock \showarticletitle{Enabling Multi-Source Neural Machine Translation By
  Concatenating Source Sentences In Multiple Languages}. In
  \bibinfo{booktitle}{\emph{Proceedings of MT Summit XVI, vol.1: Research
  Track}}. \bibinfo{address}{Nagoya, Jpaan}, \bibinfo{pages}{96--106}.
\newblock


\bibitem[\protect\citeauthoryear{Dabre, Fujita, and Chu}{Dabre
  et~al\mbox{.}}{2019}]%
        {dabre-etal-2019-exploiting}
\bibfield{author}{\bibinfo{person}{Raj Dabre}, \bibinfo{person}{Atsushi
  Fujita}, {and} \bibinfo{person}{Chenhui Chu}.}
  \bibinfo{year}{2019}\natexlab{}.
\newblock \showarticletitle{Exploiting Multilingualism through Multistage
  Fine-Tuning for Low-Resource Neural Machine Translation}. In
  \bibinfo{booktitle}{\emph{Proceedings of the 2019 Conference on Empirical
  Methods in Natural Language Processing and the 9th International Joint
  Conference on Natural Language Processing (EMNLP-IJCNLP)}}.
  \bibinfo{publisher}{Association for Computational Linguistics},
  \bibinfo{address}{Hong Kong, China}, \bibinfo{pages}{1410--1416}.
\newblock
\urldef\tempurl%
\url{https://doi.org/10.18653/v1/D19-1146}
\showDOI{\tempurl}


\bibitem[\protect\citeauthoryear{Dabre, Kunchukuttan, Fujita, and Sumita}{Dabre
  et~al\mbox{.}}{2018}]%
        {dabre18wat}
\bibfield{author}{\bibinfo{person}{Raj Dabre}, \bibinfo{person}{Anoop
  Kunchukuttan}, \bibinfo{person}{Atsushi Fujita}, {and}
  \bibinfo{person}{Eiichiro Sumita}.} \bibinfo{year}{2018}\natexlab{}.
\newblock \showarticletitle{{NICT}'s Participation in {WAT} 2018: Approaches
  Using Multilingualism and Recurrently Stacked Layers}. In
  \bibinfo{booktitle}{\emph{Proceedings of the 5th Workshop on Asian Language
  Translation}}. \bibinfo{address}{Hong Kong, China}.
\newblock


\bibitem[\protect\citeauthoryear{Dabre and Kurohashi}{Dabre and
  Kurohashi}{2017}]%
        {dabre2017mmcr4nlp}
\bibfield{author}{\bibinfo{person}{Raj Dabre} {and} \bibinfo{person}{Sadao
  Kurohashi}.} \bibinfo{year}{2017}\natexlab{}.
\newblock \showarticletitle{MMCR4NLP: Multilingual Multiway Corpora Repository
  for Natural Language Processing}.
\newblock \bibinfo{journal}{\emph{arXiv preprint arXiv:1710.01025}}
  (\bibinfo{year}{2017}).
\newblock


\bibitem[\protect\citeauthoryear{Dabre, Nakagawa, and Kazawa}{Dabre
  et~al\mbox{.}}{2017b}]%
        {Y17-1038}
\bibfield{author}{\bibinfo{person}{Raj Dabre}, \bibinfo{person}{Tetsuji
  Nakagawa}, {and} \bibinfo{person}{Hideto Kazawa}.}
  \bibinfo{year}{2017}\natexlab{b}.
\newblock \showarticletitle{An Empirical Study of Language Relatedness for
  Transfer Learning in Neural Machine Translation}. In
  \bibinfo{booktitle}{\emph{Proceedings of the 31st Pacific Asia Conference on
  Language, Information and Computation}}. \bibinfo{publisher}{The National
  University (Phillippines)}, \bibinfo{pages}{282--286}.
\newblock
\urldef\tempurl%
\url{http://aclweb.org/anthology/Y17-1038}
\showURL{%
\tempurl}


\bibitem[\protect\citeauthoryear{Di~Gangi, Cattoni, Bentivogli, Negri, and
  Turchi}{Di~Gangi et~al\mbox{.}}{2019}]%
        {gangi19}
\bibfield{author}{\bibinfo{person}{Mattia~A. Di~Gangi},
  \bibinfo{person}{Roldano Cattoni}, \bibinfo{person}{Luisa Bentivogli},
  \bibinfo{person}{Matteo Negri}, {and} \bibinfo{person}{Marco Turchi}.}
  \bibinfo{year}{2019}\natexlab{}.
\newblock \showarticletitle{{M}u{ST}-{C}: a {M}ultilingual {S}peech
  {T}ranslation {C}orpus}. In \bibinfo{booktitle}{\emph{Proceedings of the 2019
  Conference of the North {A}merican Chapter of the Association for
  Computational Linguistics: Human Language Technologies, Volume 1 (Long and
  Short Papers)}}. \bibinfo{publisher}{Association for Computational
  Linguistics}, \bibinfo{address}{Minneapolis, Minnesota},
  \bibinfo{pages}{2012--2017}.
\newblock
\urldef\tempurl%
\url{https://www.aclweb.org/anthology/N19-1202}
\showURL{%
\tempurl}


\bibitem[\protect\citeauthoryear{Dong, Wu, He, Yu, and Wang}{Dong
  et~al\mbox{.}}{2015}]%
        {dong15}
\bibfield{author}{\bibinfo{person}{Daxiang Dong}, \bibinfo{person}{Hua Wu},
  \bibinfo{person}{Wei He}, \bibinfo{person}{Dianhai Yu}, {and}
  \bibinfo{person}{Haifeng Wang}.} \bibinfo{year}{2015}\natexlab{}.
\newblock \showarticletitle{Multi-Task Learning for Multiple Language
  Translation}. In \bibinfo{booktitle}{\emph{Proceedings of the 53rd Annual
  Meeting of the Association for Computational Linguistics and the 7th
  International Joint Conference on Natural Language Processing (Volume 1: Long
  Papers)}}. \bibinfo{publisher}{Association for Computational Linguistics},
  \bibinfo{address}{Beijing, China}, \bibinfo{pages}{1723--1732}.
\newblock
\urldef\tempurl%
\url{https://doi.org/10.3115/v1/P15-1166}
\showDOI{\tempurl}


\bibitem[\protect\citeauthoryear{Dorr}{Dorr}{1987}]%
        {dorr1987unitran}
\bibfield{author}{\bibinfo{person}{Bonnie~J. Dorr}.}
  \bibinfo{year}{1987}\natexlab{}.
\newblock \showarticletitle{{UNITRAN: An Interlingua Approach to Machine
  Translation}}. In \bibinfo{booktitle}{\emph{Proceedings of the 6th Conference
  of the American Association of Artificial Intelligence}}.
  \bibinfo{address}{Seattle}.
\newblock


\bibitem[\protect\citeauthoryear{Duh, Neubig, Sudoh, and Tsukada}{Duh
  et~al\mbox{.}}{2013}]%
        {duh-EtAl:2013:Short}
\bibfield{author}{\bibinfo{person}{Kevin Duh}, \bibinfo{person}{Graham Neubig},
  \bibinfo{person}{Katsuhito Sudoh}, {and} \bibinfo{person}{Hajime Tsukada}.}
  \bibinfo{year}{2013}\natexlab{}.
\newblock \showarticletitle{Adaptation Data Selection using Neural Language
  Models: Experiments in Machine Translation}. In
  \bibinfo{booktitle}{\emph{Proceedings of the 51st Annual Meeting of the
  Association for Computational Linguistics (Volume 2: Short Papers)}}.
  \bibinfo{address}{Sofia, Bulgaria}, \bibinfo{pages}{678--683}.
\newblock
\urldef\tempurl%
\url{http://www.aclweb.org/anthology/P13-2119}
\showURL{%
\tempurl}


\bibitem[\protect\citeauthoryear{Escolano, Costa-jussà, and
  Fonollosa}{Escolano et~al\mbox{.}}{2019}]%
        {escolano19}
\bibfield{author}{\bibinfo{person}{Carlos Escolano}, \bibinfo{person}{Marta~R.
  Costa-jussà}, {and} \bibinfo{person}{José A.~R. Fonollosa}.}
  \bibinfo{year}{2019}\natexlab{}.
\newblock \showarticletitle{From Bilingual to Multilingual Neural Machine
  Translation by Incremental Training}. In
  \bibinfo{booktitle}{\emph{Proceedings of the 57th Annual Meeting of the
  Association for Computational Linguistics}}.
\newblock


\bibitem[\protect\citeauthoryear{{España-Bonet}, {Varga}, {Barrón-Cedeño},
  and {van Genabith}}{{España-Bonet} et~al\mbox{.}}{2017}]%
        {8070942}
\bibfield{author}{\bibinfo{person}{C. {España-Bonet}}, \bibinfo{person}{Á.~C.
  {Varga}}, \bibinfo{person}{A. {Barrón-Cedeño}}, {and} \bibinfo{person}{J.
  {van Genabith}}.} \bibinfo{year}{2017}\natexlab{}.
\newblock \showarticletitle{An Empirical Analysis of NMT-Derived Interlingual
  Embeddings and Their Use in Parallel Sentence Identification}.
\newblock \bibinfo{journal}{\emph{IEEE Journal of Selected Topics in Signal
  Processing}} \bibinfo{volume}{11}, \bibinfo{number}{8} (\bibinfo{date}{Dec}
  \bibinfo{year}{2017}), \bibinfo{pages}{1340--1350}.
\newblock
\showISSN{1941-0484}
\urldef\tempurl%
\url{https://doi.org/10.1109/JSTSP.2017.2764273}
\showDOI{\tempurl}


\bibitem[\protect\citeauthoryear{Finn, Abbeel, and Levine}{Finn
  et~al\mbox{.}}{2017}]%
        {finn17}
\bibfield{author}{\bibinfo{person}{Chelsea Finn}, \bibinfo{person}{Pieter
  Abbeel}, {and} \bibinfo{person}{Sergey Levine}.}
  \bibinfo{year}{2017}\natexlab{}.
\newblock \showarticletitle{Model-Agnostic Meta-Learning for Fast Adaptation of
  Deep Networks}. In \bibinfo{booktitle}{\emph{Proceedings of the 34th
  International Conference on Machine Learning}}
  \emph{(\bibinfo{series}{Proceedings of Machine Learning Research})},
  \bibfield{editor}{\bibinfo{person}{Doina Precup} {and}
  \bibinfo{person}{Yee~Whye Teh}} (Eds.), Vol.~\bibinfo{volume}{70}.
  \bibinfo{publisher}{PMLR}, \bibinfo{address}{International Convention Centre,
  Sydney, Australia}, \bibinfo{pages}{1126--1135}.
\newblock
\urldef\tempurl%
\url{http://proceedings.mlr.press/v70/finn17a.html}
\showURL{%
\tempurl}


\bibitem[\protect\citeauthoryear{Firat, Cho, and Bengio}{Firat
  et~al\mbox{.}}{2016a}]%
        {firat16}
\bibfield{author}{\bibinfo{person}{Orhan Firat}, \bibinfo{person}{Kyunghyun
  Cho}, {and} \bibinfo{person}{Yoshua Bengio}.}
  \bibinfo{year}{2016}\natexlab{a}.
\newblock \showarticletitle{Multi-Way, Multilingual Neural Machine Translation
  with a Shared Attention Mechanism}. In \bibinfo{booktitle}{\emph{Proceedings
  of the 2016 Conference of the North American Chapter of the Association for
  Computational Linguistics: Human Language Technologies}}.
  \bibinfo{publisher}{Association for Computational Linguistics},
  \bibinfo{address}{San Diego, California}, \bibinfo{pages}{866--875}.
\newblock
\urldef\tempurl%
\url{https://doi.org/10.18653/v1/N16-1101}
\showDOI{\tempurl}


\bibitem[\protect\citeauthoryear{Firat, Sankaran, Al-Onaizan, Yarman~Vural, and
  Cho}{Firat et~al\mbox{.}}{2016b}]%
        {firat16b}
\bibfield{author}{\bibinfo{person}{Orhan Firat}, \bibinfo{person}{Baskaran
  Sankaran}, \bibinfo{person}{Yaser Al-Onaizan}, \bibinfo{person}{Fatos~T.
  Yarman~Vural}, {and} \bibinfo{person}{Kyunghyun Cho}.}
  \bibinfo{year}{2016}\natexlab{b}.
\newblock \showarticletitle{Zero-Resource Translation with Multi-Lingual Neural
  Machine Translation}. In \bibinfo{booktitle}{\emph{Proceedings of the 2016
  Conference on Empirical Methods in Natural Language Processing}}.
  \bibinfo{publisher}{Association for Computational Linguistics},
  \bibinfo{address}{Austin, Texas}, \bibinfo{pages}{268--277}.
\newblock
\urldef\tempurl%
\url{https://doi.org/10.18653/v1/D16-1026}
\showDOI{\tempurl}


\bibitem[\protect\citeauthoryear{Ganin, Ustinova, Ajakan, Germain, Larochelle,
  Laviolette, Marchand, and Lempitsky}{Ganin et~al\mbox{.}}{2016}]%
        {ganin2016domain}
\bibfield{author}{\bibinfo{person}{Yaroslav Ganin}, \bibinfo{person}{Evgeniya
  Ustinova}, \bibinfo{person}{Hana Ajakan}, \bibinfo{person}{Pascal Germain},
  \bibinfo{person}{Hugo Larochelle}, \bibinfo{person}{Fran{\c{c}}ois
  Laviolette}, \bibinfo{person}{Mario Marchand}, {and} \bibinfo{person}{Victor
  Lempitsky}.} \bibinfo{year}{2016}\natexlab{}.
\newblock \showarticletitle{Domain-adversarial training of neural networks}.
\newblock \bibinfo{journal}{\emph{The Journal of Machine Learning Research}}
  \bibinfo{volume}{17}, \bibinfo{number}{1} (\bibinfo{year}{2016}),
  \bibinfo{pages}{2096--2030}.
\newblock


\bibitem[\protect\citeauthoryear{Garmash and Monz}{Garmash and Monz}{2016}]%
        {C16-1133}
\bibfield{author}{\bibinfo{person}{Ekaterina Garmash} {and}
  \bibinfo{person}{Christof Monz}.} \bibinfo{year}{2016}\natexlab{}.
\newblock \showarticletitle{Ensemble Learning for Multi-Source Neural Machine
  Translation}. In \bibinfo{booktitle}{\emph{Proceedings of COLING 2016, the
  26th International Conference on Computational Linguistics: Technical
  Papers}}. \bibinfo{publisher}{The COLING 2016 Organizing Committee},
  \bibinfo{address}{Osaka, Japan}, \bibinfo{pages}{1409--1418}.
\newblock
\urldef\tempurl%
\url{http://aclweb.org/anthology/C16-1133}
\showURL{%
\tempurl}


\bibitem[\protect\citeauthoryear{Gehring, Auli, Grangier, Yarats, and
  Dauphin}{Gehring et~al\mbox{.}}{2017}]%
        {gehring17}
\bibfield{author}{\bibinfo{person}{Jonas Gehring}, \bibinfo{person}{Michael
  Auli}, \bibinfo{person}{David Grangier}, \bibinfo{person}{Denis Yarats},
  {and} \bibinfo{person}{Yann~N. Dauphin}.} \bibinfo{year}{2017}\natexlab{}.
\newblock \showarticletitle{Convolutional Sequence to Sequence Learning}. In
  \bibinfo{booktitle}{\emph{Proceedings of the 34th International Conference on
  Machine Learning}} \emph{(\bibinfo{series}{Proceedings of Machine Learning
  Research})}, \bibfield{editor}{\bibinfo{person}{Doina Precup} {and}
  \bibinfo{person}{Yee~Whye Teh}} (Eds.), Vol.~\bibinfo{volume}{70}.
  \bibinfo{publisher}{PMLR}, \bibinfo{address}{International Convention Centre,
  Sydney, Australia}, \bibinfo{pages}{1243--1252}.
\newblock
\urldef\tempurl%
\url{http://proceedings.mlr.press/v70/gehring17a.html}
\showURL{%
\tempurl}


\bibitem[\protect\citeauthoryear{Gispert and Marino}{Gispert and
  Marino}{2006}]%
        {de2006catalanenglish}
\bibfield{author}{\bibinfo{person}{Adri`a~De Gispert} {and}
  \bibinfo{person}{Jose~B Marino}.} \bibinfo{year}{2006}\natexlab{}.
\newblock \showarticletitle{{{Catalan-English statistical machine translation
  without parallel corpus: bridging through Spanish}}}. In
  \bibinfo{booktitle}{\emph{Proceedings of the 5th International Conference on
  Language Resources and Evaluation (LREC)}}. \bibinfo{address}{Genoa, Italy},
  \bibinfo{pages}{65--68}.
\newblock


\bibitem[\protect\citeauthoryear{Gu, Hassan, Devlin, and Li}{Gu
  et~al\mbox{.}}{2018a}]%
        {gu18}
\bibfield{author}{\bibinfo{person}{Jiatao Gu}, \bibinfo{person}{Hany Hassan},
  \bibinfo{person}{Jacob Devlin}, {and} \bibinfo{person}{Victor~O.K. Li}.}
  \bibinfo{year}{2018}\natexlab{a}.
\newblock \showarticletitle{Universal Neural Machine Translation for Extremely
  Low Resource Languages}. In \bibinfo{booktitle}{\emph{Proceedings of the 2018
  Conference of the North American Chapter of the Association for Computational
  Linguistics: Human Language Technologies, Volume 1 (Long Papers)}}.
  \bibinfo{publisher}{Association for Computational Linguistics},
  \bibinfo{address}{New Orleans, Louisiana}, \bibinfo{pages}{344--354}.
\newblock
\urldef\tempurl%
\url{https://doi.org/10.18653/v1/N18-1032}
\showDOI{\tempurl}


\bibitem[\protect\citeauthoryear{Gu, Wang, Chen, Li, and Cho}{Gu
  et~al\mbox{.}}{2018b}]%
        {gu18b}
\bibfield{author}{\bibinfo{person}{Jiatao Gu}, \bibinfo{person}{Yong Wang},
  \bibinfo{person}{Yun Chen}, \bibinfo{person}{Victor O.~K. Li}, {and}
  \bibinfo{person}{Kyunghyun Cho}.} \bibinfo{year}{2018}\natexlab{b}.
\newblock \showarticletitle{Meta-Learning for Low-Resource Neural Machine
  Translation}. In \bibinfo{booktitle}{\emph{Proceedings of the 2018 Conference
  on Empirical Methods in Natural Language Processing}}.
  \bibinfo{publisher}{Association for Computational Linguistics},
  \bibinfo{address}{Brussels, Belgium}, \bibinfo{pages}{3622--3631}.
\newblock
\urldef\tempurl%
\url{http://aclweb.org/anthology/D18-1398}
\showURL{%
\tempurl}


\bibitem[\protect\citeauthoryear{Gu, Wang, Cho, and Li}{Gu
  et~al\mbox{.}}{2019}]%
        {gu-etal-2019-improved}
\bibfield{author}{\bibinfo{person}{Jiatao Gu}, \bibinfo{person}{Yong Wang},
  \bibinfo{person}{Kyunghyun Cho}, {and} \bibinfo{person}{Victor~O.K. Li}.}
  \bibinfo{year}{2019}\natexlab{}.
\newblock \showarticletitle{Improved Zero-shot Neural Machine Translation via
  Ignoring Spurious Correlations}. In \bibinfo{booktitle}{\emph{Proceedings of
  the 57th Annual Meeting of the Association for Computational Linguistics}}.
  \bibinfo{publisher}{Association for Computational Linguistics},
  \bibinfo{address}{Florence, Italy}, \bibinfo{pages}{1258--1268}.
\newblock
\urldef\tempurl%
\url{https://doi.org/10.18653/v1/P19-1121}
\showDOI{\tempurl}


\bibitem[\protect\citeauthoryear{Guzm\'{a}n, Chen, Ott, Pino, Lample, Koehn,
  Chaudhary, and Ranzato}{Guzm\'{a}n et~al\mbox{.}}{2019}]%
        {guzman2019flores}
\bibfield{author}{\bibinfo{person}{Francisco Guzm\'{a}n},
  \bibinfo{person}{Peng-Jen Chen}, \bibinfo{person}{Myle Ott},
  \bibinfo{person}{Juan Pino}, \bibinfo{person}{Guillaume Lample},
  \bibinfo{person}{Philipp Koehn}, \bibinfo{person}{Vishrav Chaudhary}, {and}
  \bibinfo{person}{Marc'Aurelio Ranzato}.} \bibinfo{year}{2019}\natexlab{}.
\newblock \showarticletitle{{Two New Evaluation Datasets for Low-Resource
  Machine Translation: Nepali-English and Sinhala-English}}.
\newblock \bibinfo{journal}{\emph{arXiv preprint arXiv:1902.01382}}
  (\bibinfo{year}{2019}).
\newblock


\bibitem[\protect\citeauthoryear{Ha, Niehues, and Waibel}{Ha
  et~al\mbox{.}}{2016}]%
        {ha16}
\bibfield{author}{\bibinfo{person}{Thanh{-}Le Ha}, \bibinfo{person}{Jan
  Niehues}, {and} \bibinfo{person}{Alexander~H. Waibel}.}
  \bibinfo{year}{2016}\natexlab{}.
\newblock \showarticletitle{Toward Multilingual Neural Machine Translation with
  Universal Encoder and Decoder}. In \bibinfo{booktitle}{\emph{Proceedings of
  the 13th International Workshop on Spoken Language Translation}}.
  \bibinfo{address}{Seattle}, \bibinfo{pages}{1--7}.
\newblock


\bibitem[\protect\citeauthoryear{Ha, Niehues, and Waibel}{Ha
  et~al\mbox{.}}{2017}]%
        {DBLP:journals/corr/abs-1711-07893}
\bibfield{author}{\bibinfo{person}{Thanh{-}Le Ha}, \bibinfo{person}{Jan
  Niehues}, {and} \bibinfo{person}{Alexander~H. Waibel}.}
  \bibinfo{year}{2017}\natexlab{}.
\newblock \showarticletitle{Effective Strategies in Zero-Shot Neural Machine
  Translation}. In \bibinfo{booktitle}{\emph{Proceedings of the 14th
  International Workshop on Spoken Language Translation}}.
  \bibinfo{address}{Tokyo, Japan}, \bibinfo{pages}{105--112}.
\newblock


\bibitem[\protect\citeauthoryear{He, Gu, Shen, and Ranzato}{He
  et~al\mbox{.}}{2020}]%
        {he2020revisiting}
\bibfield{author}{\bibinfo{person}{Junxian He}, \bibinfo{person}{Jiatao Gu},
  \bibinfo{person}{Jiajun Shen}, {and} \bibinfo{person}{Marc'Aurelio Ranzato}.}
  \bibinfo{year}{2020}\natexlab{}.
\newblock \showarticletitle{Revisiting self-training for neural sequence
  generation}. In \bibinfo{booktitle}{\emph{ICLR}}.
\newblock


\bibitem[\protect\citeauthoryear{Henr\'iquez, Costa-juss\'a, Banchs, Formiga,
  and no}{Henr\'iquez et~al\mbox{.}}{2011}]%
        {henriquez2011pivot}
\bibfield{author}{\bibinfo{person}{Carlos Henr\'iquez},
  \bibinfo{person}{Marta~R. Costa-juss\'a}, \bibinfo{person}{Rafael~E. Banchs},
  \bibinfo{person}{Lluis Formiga}, {and} \bibinfo{person}{Jos\'e B.~Mari\ no}.}
  \bibinfo{year}{2011}\natexlab{}.
\newblock \showarticletitle{{Pivot Strategies as an Alternative for Statistical
  Machine Translation Tasks Involving Iberian Languages}}. In
  \bibinfo{booktitle}{\emph{Proceedings of the Workshop on Iberian
  Cross-Language Natural Language Processing Tasks (ICL 2011)}}.
  \bibinfo{address}{Huelva, Spain}, \bibinfo{pages}{22--27}.
\newblock


\bibitem[\protect\citeauthoryear{Hinton, Vinyals, and Dean}{Hinton
  et~al\mbox{.}}{2015}]%
        {HinVin15Distilling}
\bibfield{author}{\bibinfo{person}{Geoffrey Hinton}, \bibinfo{person}{Oriol
  Vinyals}, {and} \bibinfo{person}{Jeff Dean}.}
  \bibinfo{year}{2015}\natexlab{}.
\newblock \showarticletitle{Distilling the knowledge in a neural network}.
\newblock \bibinfo{journal}{\emph{arXiv preprint arXiv:1503.02531}}
  (\bibinfo{year}{2015}).
\newblock
\urldef\tempurl%
\url{https://arxiv.org/abs/1503.02531v1}
\showURL{%
\tempurl}


\bibitem[\protect\citeauthoryear{Hokamp, Glover, and
  Gholipour~Ghalandari}{Hokamp et~al\mbox{.}}{2019}]%
        {hokamp-etal-2019-evaluating}
\bibfield{author}{\bibinfo{person}{Chris Hokamp}, \bibinfo{person}{John
  Glover}, {and} \bibinfo{person}{Demian Gholipour~Ghalandari}.}
  \bibinfo{year}{2019}\natexlab{}.
\newblock \showarticletitle{Evaluating the Supervised and Zero-shot Performance
  of Multi-lingual Translation Models}. In
  \bibinfo{booktitle}{\emph{Proceedings of the Fourth Conference on Machine
  Translation (Volume 2: Shared Task Papers, Day 1)}}.
  \bibinfo{publisher}{Association for Computational Linguistics},
  \bibinfo{address}{Florence, Italy}, \bibinfo{pages}{209--217}.
\newblock
\urldef\tempurl%
\url{https://doi.org/10.18653/v1/W19-5319}
\showDOI{\tempurl}


\bibitem[\protect\citeauthoryear{Jawanpuria, Balgovind, Kunchukuttan, and
  Mishra}{Jawanpuria et~al\mbox{.}}{2019}]%
        {jawanpuria2018learning}
\bibfield{author}{\bibinfo{person}{Pratik Jawanpuria}, \bibinfo{person}{Arjun
  Balgovind}, \bibinfo{person}{Anoop Kunchukuttan}, {and}
  \bibinfo{person}{Bamdev Mishra}.} \bibinfo{year}{2019}\natexlab{}.
\newblock \showarticletitle{Learning Multilingual Word Embeddings in Latent
  Metric Space: A Geometric Approach}.
\newblock \bibinfo{journal}{\emph{Transactions of the Association for
  Computational Linguistics}}  \bibinfo{volume}{7} (\bibinfo{year}{2019}),
  \bibinfo{pages}{107--120}.
\newblock
\urldef\tempurl%
\url{https://doi.org/10.1162/tacl_a_00257}
\showDOI{\tempurl}


\bibitem[\protect\citeauthoryear{Jean, Firat, and Johnson}{Jean
  et~al\mbox{.}}{2019}]%
        {Jean2019AdaptiveSF}
\bibfield{author}{\bibinfo{person}{S{\'e}bastien Jean}, \bibinfo{person}{Orhan
  Firat}, {and} \bibinfo{person}{Melvin Johnson}.}
  \bibinfo{year}{2019}\natexlab{}.
\newblock \showarticletitle{Adaptive Scheduling for Multi-Task Learning}.
\newblock \bibinfo{journal}{\emph{ArXiv}}  \bibinfo{volume}{abs/1909.06434}
  (\bibinfo{year}{2019}).
\newblock


\bibitem[\protect\citeauthoryear{Jha}{Jha}{2010}]%
        {jha2010tdil}
\bibfield{author}{\bibinfo{person}{Girish~Nath Jha}.}
  \bibinfo{year}{2010}\natexlab{}.
\newblock \showarticletitle{The {TDIL} Program and the {I}ndian Langauge
  Corpora Intitiative ({ILCI})}. In \bibinfo{booktitle}{\emph{Proceedings of
  the Seventh conference on International Language Resources and Evaluation
  ({LREC}{'}10)}}. \bibinfo{publisher}{European Languages Resources Association
  (ELRA)}, \bibinfo{address}{Valletta, Malta}.
\newblock
\urldef\tempurl%
\url{http://www.lrec-conf.org/proceedings/lrec2010/pdf/874_Paper.pdf}
\showURL{%
\tempurl}


\bibitem[\protect\citeauthoryear{Ji, Zhang, Duan, Zhang, Chen, and Luo}{Ji
  et~al\mbox{.}}{2020}]%
        {ji2020cross}
\bibfield{author}{\bibinfo{person}{Baijun Ji}, \bibinfo{person}{Zhirui Zhang},
  \bibinfo{person}{Xiangyu Duan}, \bibinfo{person}{Min Zhang},
  \bibinfo{person}{Boxing Chen}, {and} \bibinfo{person}{Weihua Luo}.}
  \bibinfo{year}{2020}\natexlab{}.
\newblock \showarticletitle{Cross-lingual Pre-training Based Transfer for
  Zero-shot Neural Machine Translation}. In
  \bibinfo{booktitle}{\emph{Proceedings of the Thirty-Fourth AAAI Conference on
  Artificial Intelligence}}.
\newblock


\bibitem[\protect\citeauthoryear{Johnson, Schuster, Le, Krikun, Wu, Chen,
  Thorat, Vi{\'e}gas, Wattenberg, Corrado, Hughes, and Dean}{Johnson
  et~al\mbox{.}}{2017}]%
        {johnson17}
\bibfield{author}{\bibinfo{person}{Melvin Johnson}, \bibinfo{person}{Mike
  Schuster}, \bibinfo{person}{Quoc~V. Le}, \bibinfo{person}{Maxim Krikun},
  \bibinfo{person}{Yonghui Wu}, \bibinfo{person}{Zhifeng Chen},
  \bibinfo{person}{Nikhil Thorat}, \bibinfo{person}{Fernanda Vi{\'e}gas},
  \bibinfo{person}{Martin Wattenberg}, \bibinfo{person}{Greg Corrado},
  \bibinfo{person}{Macduff Hughes}, {and} \bibinfo{person}{Jeffrey Dean}.}
  \bibinfo{year}{2017}\natexlab{}.
\newblock \showarticletitle{Google's Multilingual Neural Machine Translation
  System: Enabling Zero-Shot Translation}.
\newblock \bibinfo{journal}{\emph{Transactions of the Association for
  Computational Linguistics}}  \bibinfo{volume}{5} (\bibinfo{year}{2017}),
  \bibinfo{pages}{339--351}.
\newblock
\urldef\tempurl%
\url{http://aclweb.org/anthology/Q17-1024}
\showURL{%
\tempurl}


\bibitem[\protect\citeauthoryear{Kim, Gao, and Ney}{Kim et~al\mbox{.}}{2019a}]%
        {kim19}
\bibfield{author}{\bibinfo{person}{Yunsu Kim}, \bibinfo{person}{Yingbo Gao},
  {and} \bibinfo{person}{Hermann Ney}.} \bibinfo{year}{2019}\natexlab{a}.
\newblock \showarticletitle{Effective Cross-lingual Transfer of Neural Machine
  Translation Models without Shared Vocabularies}. In
  \bibinfo{booktitle}{\emph{Proceedings of the 57th Annual Meeting of the
  Association for Computational Linguistics}}.
\newblock


\bibitem[\protect\citeauthoryear{Kim, Petrov, Petrushkov, Khadivi, and Ney}{Kim
  et~al\mbox{.}}{2019b}]%
        {kim-etal-2019-pivot}
\bibfield{author}{\bibinfo{person}{Yunsu Kim}, \bibinfo{person}{Petre Petrov},
  \bibinfo{person}{Pavel Petrushkov}, \bibinfo{person}{Shahram Khadivi}, {and}
  \bibinfo{person}{Hermann Ney}.} \bibinfo{year}{2019}\natexlab{b}.
\newblock \showarticletitle{Pivot-based Transfer Learning for Neural Machine
  Translation between Non-{E}nglish Languages}. In
  \bibinfo{booktitle}{\emph{Proceedings of the 2019 Conference on Empirical
  Methods in Natural Language Processing and the 9th International Joint
  Conference on Natural Language Processing (EMNLP-IJCNLP)}}.
  \bibinfo{publisher}{Association for Computational Linguistics},
  \bibinfo{address}{Hong Kong, China}, \bibinfo{pages}{866--876}.
\newblock
\urldef\tempurl%
\url{https://doi.org/10.18653/v1/D19-1080}
\showDOI{\tempurl}


\bibitem[\protect\citeauthoryear{Kim and Rush}{Kim and Rush}{2016}]%
        {kim-rush-2016-sequence}
\bibfield{author}{\bibinfo{person}{Yoon Kim} {and}
  \bibinfo{person}{Alexander~M. Rush}.} \bibinfo{year}{2016}\natexlab{}.
\newblock \showarticletitle{Sequence-Level Knowledge Distillation}. In
  \bibinfo{booktitle}{\emph{Proceedings of the 2016 Conference on Empirical
  Methods in Natural Language Processing}}. \bibinfo{publisher}{Association for
  Computational Linguistics}, \bibinfo{address}{Austin, Texas},
  \bibinfo{pages}{1317--1327}.
\newblock
\urldef\tempurl%
\url{https://doi.org/10.18653/v1/D16-1139}
\showDOI{\tempurl}


\bibitem[\protect\citeauthoryear{Kiperwasser and Ballesteros}{Kiperwasser and
  Ballesteros}{2018}]%
        {kiperwasser-ballesteros-2018-scheduled}
\bibfield{author}{\bibinfo{person}{Eliyahu Kiperwasser} {and}
  \bibinfo{person}{Miguel Ballesteros}.} \bibinfo{year}{2018}\natexlab{}.
\newblock \showarticletitle{Scheduled Multi-Task Learning: From Syntax to
  Translation}.
\newblock \bibinfo{journal}{\emph{Transactions of the Association for
  Computational Linguistics}}  \bibinfo{volume}{6} (\bibinfo{year}{2018}),
  \bibinfo{pages}{225--240}.
\newblock
\urldef\tempurl%
\url{https://doi.org/10.1162/tacl_a_00017}
\showDOI{\tempurl}


\bibitem[\protect\citeauthoryear{Klementiev, Titov, and Bhattarai}{Klementiev
  et~al\mbox{.}}{2012}]%
        {klementiev12a}
\bibfield{author}{\bibinfo{person}{Alexandre Klementiev}, \bibinfo{person}{Ivan
  Titov}, {and} \bibinfo{person}{Binod Bhattarai}.}
  \bibinfo{year}{2012}\natexlab{}.
\newblock \showarticletitle{Inducing crosslingual distributed representations
  of words}. In \bibinfo{booktitle}{\emph{Proceedings of the International
  Conference on Computational Linguistics: Technical Papers}}.
  \bibinfo{address}{Mumbai, India}, \bibinfo{pages}{1459--1474}.
\newblock


\bibitem[\protect\citeauthoryear{Kocmi and Bojar}{Kocmi and Bojar}{2018}]%
        {kocmi-bojar:2018:WMT}
\bibfield{author}{\bibinfo{person}{Tom Kocmi} {and} \bibinfo{person}{OndÅ™ej
  Bojar}.} \bibinfo{year}{2018}\natexlab{}.
\newblock \showarticletitle{Trivial Transfer Learning for Low-Resource Neural
  Machine Translation}. In \bibinfo{booktitle}{\emph{Proceedings of the Third
  Conference on Machine Translation, Volume 1: Research Papers}}.
  \bibinfo{publisher}{Association for Computational Linguistics},
  \bibinfo{address}{Belgium, Brussels}, \bibinfo{pages}{244--252}.
\newblock
\urldef\tempurl%
\url{http://www.aclweb.org/anthology/W18-6325}
\showURL{%
\tempurl}


\bibitem[\protect\citeauthoryear{Koehn}{Koehn}{2005}]%
        {koehn2005epc}
\bibfield{author}{\bibinfo{person}{Philipp Koehn}.}
  \bibinfo{year}{2005}\natexlab{}.
\newblock \showarticletitle{{Europarl: A Parallel Corpus for Statistical
  Machine Translation}}. In \bibinfo{booktitle}{\emph{{Conference Proceedings:
  the tenth Machine Translation Summit}}}. AAMT, \bibinfo{publisher}{AAMT},
  \bibinfo{address}{Phuket, Thailand}, \bibinfo{pages}{79--86}.
\newblock
\urldef\tempurl%
\url{http://mt-archive.info/MTS-2005-Koehn.pdf}
\showURL{%
\tempurl}


\bibitem[\protect\citeauthoryear{Koehn}{Koehn}{2017}]%
        {koehn17nmt}
\bibfield{author}{\bibinfo{person}{Philipp Koehn}.}
  \bibinfo{year}{2017}\natexlab{}.
\newblock \showarticletitle{Neural Machine Translation}.
\newblock \bibinfo{journal}{\emph{CoRR}}  \bibinfo{volume}{abs/1709.07809}
  (\bibinfo{year}{2017}).
\newblock
\showeprint[arxiv]{1709.07809}
\urldef\tempurl%
\url{http://arxiv.org/abs/1709.07809}
\showURL{%
\tempurl}


\bibitem[\protect\citeauthoryear{Koehn, Hoang, Birch, Callison-Burch, Federico,
  Bertoldi, Cowan, Shen, Moran, Zens, Dyer, Bojar, Constantin, and
  Herbst}{Koehn et~al\mbox{.}}{2007}]%
        {koehn-EtAl:2007:PosterDemo}
\bibfield{author}{\bibinfo{person}{Philipp Koehn}, \bibinfo{person}{Hieu
  Hoang}, \bibinfo{person}{Alexandra Birch}, \bibinfo{person}{Chris
  Callison-Burch}, \bibinfo{person}{Marcello Federico}, \bibinfo{person}{Nicola
  Bertoldi}, \bibinfo{person}{Brooke Cowan}, \bibinfo{person}{Wade Shen},
  \bibinfo{person}{Christine Moran}, \bibinfo{person}{Richard Zens},
  \bibinfo{person}{Chris Dyer}, \bibinfo{person}{Ondrej Bojar},
  \bibinfo{person}{Alexandra Constantin}, {and} \bibinfo{person}{Evan Herbst}.}
  \bibinfo{year}{2007}\natexlab{}.
\newblock \showarticletitle{Moses: Open Source Toolkit for Statistical Machine
  Translation}. In \bibinfo{booktitle}{\emph{Proceedings of the 45th Annual
  Meeting of the Association for Computational Linguistics Companion Volume
  Proceedings of the Demo and Poster Sessions}}.
  \bibinfo{publisher}{Association for Computational Linguistics},
  \bibinfo{address}{Prague, Czech Republic}, \bibinfo{pages}{177--180}.
\newblock
\urldef\tempurl%
\url{http://www.aclweb.org/anthology/P/P07/P07-2045}
\showURL{%
\tempurl}


\bibitem[\protect\citeauthoryear{Koehn and Knowles}{Koehn and Knowles}{2017}]%
        {koehn-knowles:2017:NMT}
\bibfield{author}{\bibinfo{person}{Philipp Koehn} {and}
  \bibinfo{person}{Rebecca Knowles}.} \bibinfo{year}{2017}\natexlab{}.
\newblock \showarticletitle{Six Challenges for Neural Machine Translation}. In
  \bibinfo{booktitle}{\emph{Proceedings of the First Workshop on Neural Machine
  Translation}}. \bibinfo{publisher}{Association for Computational
  Linguistics}, \bibinfo{address}{Vancouver}, \bibinfo{pages}{28--39}.
\newblock
\urldef\tempurl%
\url{http://www.aclweb.org/anthology/W17-3204}
\showURL{%
\tempurl}


\bibitem[\protect\citeauthoryear{Koehn, Och, and Marcu}{Koehn
  et~al\mbox{.}}{2003}]%
        {koehn2003statistical}
\bibfield{author}{\bibinfo{person}{Philipp Koehn}, \bibinfo{person}{Franz~J.
  Och}, {and} \bibinfo{person}{Daniel Marcu}.} \bibinfo{year}{2003}\natexlab{}.
\newblock \showarticletitle{Statistical Phrase-Based Translation}. In
  \bibinfo{booktitle}{\emph{Proceedings of the 2003 Human Language Technology
  Conference of the North {A}merican Chapter of the Association for
  Computational Linguistics}}. \bibinfo{pages}{127--133}.
\newblock
\urldef\tempurl%
\url{https://www.aclweb.org/anthology/N03-1017}
\showURL{%
\tempurl}


\bibitem[\protect\citeauthoryear{Kudo and Richardson}{Kudo and
  Richardson}{2018}]%
        {kudo-richardson-2018-sentencepiece}
\bibfield{author}{\bibinfo{person}{Taku Kudo} {and} \bibinfo{person}{John
  Richardson}.} \bibinfo{year}{2018}\natexlab{}.
\newblock \showarticletitle{{S}entence{P}iece: A simple and language
  independent subword tokenizer and detokenizer for Neural Text Processing}. In
  \bibinfo{booktitle}{\emph{Proceedings of the 2018 Conference on Empirical
  Methods in Natural Language Processing: System Demonstrations}}.
  \bibinfo{publisher}{Association for Computational Linguistics},
  \bibinfo{address}{Brussels, Belgium}, \bibinfo{pages}{66--71}.
\newblock
\urldef\tempurl%
\url{https://doi.org/10.18653/v1/D18-2012}
\showDOI{\tempurl}


\bibitem[\protect\citeauthoryear{Kudugunta, Bapna, Caswell, and
  Firat}{Kudugunta et~al\mbox{.}}{2019}]%
        {kudugunta-etal-2019-investigating}
\bibfield{author}{\bibinfo{person}{Sneha Kudugunta}, \bibinfo{person}{Ankur
  Bapna}, \bibinfo{person}{Isaac Caswell}, {and} \bibinfo{person}{Orhan
  Firat}.} \bibinfo{year}{2019}\natexlab{}.
\newblock \showarticletitle{Investigating Multilingual {NMT} Representations at
  Scale}. In \bibinfo{booktitle}{\emph{Proceedings of the 2019 Conference on
  Empirical Methods in Natural Language Processing and the 9th International
  Joint Conference on Natural Language Processing (EMNLP-IJCNLP)}}.
  \bibinfo{publisher}{Association for Computational Linguistics},
  \bibinfo{address}{Hong Kong, China}, \bibinfo{pages}{1565--1575}.
\newblock
\urldef\tempurl%
\url{https://doi.org/10.18653/v1/D19-1167}
\showDOI{\tempurl}


\bibitem[\protect\citeauthoryear{Kunchukuttan and Bhattacharyya}{Kunchukuttan
  and Bhattacharyya}{2016}]%
        {kunchukuttan2016orthographic}
\bibfield{author}{\bibinfo{person}{Anoop Kunchukuttan} {and}
  \bibinfo{person}{Pushpak Bhattacharyya}.} \bibinfo{year}{2016}\natexlab{}.
\newblock \showarticletitle{Orthographic Syllable as basic unit for {SMT}
  between Related Languages}. In \bibinfo{booktitle}{\emph{Proceedings of the
  2016 Conference on Empirical Methods in Natural Language Processing}}.
  \bibinfo{publisher}{Association for Computational Linguistics},
  \bibinfo{address}{Austin, Texas}, \bibinfo{pages}{1912--1917}.
\newblock
\urldef\tempurl%
\url{https://doi.org/10.18653/v1/D16-1196}
\showDOI{\tempurl}


\bibitem[\protect\citeauthoryear{Kunchukuttan and Bhattacharyya}{Kunchukuttan
  and Bhattacharyya}{2017}]%
        {kunchukuttan2017bpe}
\bibfield{author}{\bibinfo{person}{Anoop Kunchukuttan} {and}
  \bibinfo{person}{Pushpak Bhattacharyya}.} \bibinfo{year}{2017}\natexlab{}.
\newblock \showarticletitle{Learning variable length units for {SMT} between
  related languages via Byte Pair Encoding}. In
  \bibinfo{booktitle}{\emph{Proceedings of the First Workshop on Subword and
  Character Level Models in {NLP}}}. \bibinfo{publisher}{Association for
  Computational Linguistics}, \bibinfo{address}{Copenhagen, Denmark},
  \bibinfo{pages}{14--24}.
\newblock
\urldef\tempurl%
\url{https://doi.org/10.18653/v1/W17-4102}
\showDOI{\tempurl}


\bibitem[\protect\citeauthoryear{Kunchukuttan, Shah, Prakash, and
  Bhattacharyya}{Kunchukuttan et~al\mbox{.}}{2017}]%
        {kunchukuttan2017pivot}
\bibfield{author}{\bibinfo{person}{Anoop Kunchukuttan}, \bibinfo{person}{Maulik
  Shah}, \bibinfo{person}{Pradyot Prakash}, {and} \bibinfo{person}{Pushpak
  Bhattacharyya}.} \bibinfo{year}{2017}\natexlab{}.
\newblock \showarticletitle{Utilizing Lexical Similarity between Related,
  Low-resource Languages for Pivot-based SMT}. In
  \bibinfo{booktitle}{\emph{Proceedings of the Eighth International Joint
  Conference on Natural Language Processing (Volume 2: Short Papers)}}.
  \bibinfo{publisher}{Asian Federation of Natural Language Processing},
  \bibinfo{address}{Taipei, Taiwan}, \bibinfo{pages}{283--289}.
\newblock
\urldef\tempurl%
\url{http://aclweb.org/anthology/I17-2048}
\showURL{%
\tempurl}


\bibitem[\protect\citeauthoryear{Lakew, Cettolo, and Federico}{Lakew
  et~al\mbox{.}}{2018a}]%
        {lakew18}
\bibfield{author}{\bibinfo{person}{Surafel~Melaku Lakew},
  \bibinfo{person}{Mauro Cettolo}, {and} \bibinfo{person}{Marcello Federico}.}
  \bibinfo{year}{2018}\natexlab{a}.
\newblock \showarticletitle{A Comparison of Transformer and Recurrent Neural
  Networks on Multilingual Neural Machine Translation}. In
  \bibinfo{booktitle}{\emph{Proceedings of the 27th International Conference on
  Computational Linguistics}}. \bibinfo{publisher}{Association for
  Computational Linguistics}, \bibinfo{address}{Santa Fe, New Mexico, USA},
  \bibinfo{pages}{641--652}.
\newblock
\urldef\tempurl%
\url{http://aclweb.org/anthology/C18-1054}
\showURL{%
\tempurl}


\bibitem[\protect\citeauthoryear{Lakew, Erofeeva, Negri, Federico, and
  Turchi}{Lakew et~al\mbox{.}}{2018b}]%
        {DBLP:journals/corr/abs-1811-01137}
\bibfield{author}{\bibinfo{person}{Surafel~Melaku Lakew},
  \bibinfo{person}{Aliia Erofeeva}, \bibinfo{person}{Matteo Negri},
  \bibinfo{person}{Marcello Federico}, {and} \bibinfo{person}{Marco Turchi}.}
  \bibinfo{year}{2018}\natexlab{b}.
\newblock \showarticletitle{Transfer Learning in Multilingual Neural Machine
  Translation with Dynamic Vocabulary}. In
  \bibinfo{booktitle}{\emph{Proceedings of the 15th International Workshop on
  Spoken Language Translation ({IWSLT})}}. \bibinfo{pages}{54--61}.
\newblock


\bibitem[\protect\citeauthoryear{Lakew, Lotito, Negri, Turchi, and
  Federico}{Lakew et~al\mbox{.}}{2017a}]%
        {lakew-etal-2017-zeroshot}
\bibfield{author}{\bibinfo{person}{Surafel~Melaku Lakew},
  \bibinfo{person}{Quintino~F. Lotito}, \bibinfo{person}{Matteo Negri},
  \bibinfo{person}{Marco Turchi}, {and} \bibinfo{person}{Marcello Federico}.}
  \bibinfo{year}{2017}\natexlab{a}.
\newblock \showarticletitle{Improving Zero-Shot Translation of Low-Resource
  Languages}. In \bibinfo{booktitle}{\emph{International Workshop on Spoken
  Language Translation}}.
\newblock


\bibitem[\protect\citeauthoryear{Lakew, Lotito, Negri, Turchi, and
  Federico}{Lakew et~al\mbox{.}}{2017b}]%
        {DBLP:journals/corr/abs-1811-01389}
\bibfield{author}{\bibinfo{person}{Surafel~Melaku Lakew},
  \bibinfo{person}{Quintino~F. Lotito}, \bibinfo{person}{Matteo Negri},
  \bibinfo{person}{Marco Turchi}, {and} \bibinfo{person}{Marcello Federico}.}
  \bibinfo{year}{2017}\natexlab{b}.
\newblock \showarticletitle{Improving Zero-Shot Translation of Low-Resource
  Languages}. In \bibinfo{booktitle}{\emph{Proceedings of the 14th
  International Workshop on Spoken Language Translation}}.
  \bibinfo{address}{Tokyo, Japan}, \bibinfo{pages}{113--119}.
\newblock


\bibitem[\protect\citeauthoryear{Lample and Conneau}{Lample and
  Conneau}{2019}]%
        {lample2019crosslingual}
\bibfield{author}{\bibinfo{person}{Guillaume Lample} {and}
  \bibinfo{person}{Alexis Conneau}.} \bibinfo{year}{2019}\natexlab{}.
\newblock \showarticletitle{Cross-lingual Language Model Pretraining}.
\newblock \bibinfo{journal}{\emph{CoRR}}  \bibinfo{volume}{abs/1901.07291}
  (\bibinfo{year}{2019}).
\newblock
\showeprint[arxiv]{1901.07291}
\urldef\tempurl%
\url{http://arxiv.org/abs/1901.07291}
\showURL{%
\tempurl}


\bibitem[\protect\citeauthoryear{Lample, Conneau, Denoyer, and Ranzato}{Lample
  et~al\mbox{.}}{2018}]%
        {lample2018unsupervised}
\bibfield{author}{\bibinfo{person}{Guillaume Lample}, \bibinfo{person}{Alexis
  Conneau}, \bibinfo{person}{Ludovic Denoyer}, {and}
  \bibinfo{person}{Marc'Aurelio Ranzato}.} \bibinfo{year}{2018}\natexlab{}.
\newblock \showarticletitle{Unsupervised Machine Translation Using Monolingual
  Corpora Only}. In \bibinfo{booktitle}{\emph{Proceedings of International
  Conference on Learning Representations}}. \bibinfo{address}{Vancouver, BC,
  Canada}.
\newblock
\urldef\tempurl%
\url{https://openreview.net/forum?id=rkYTTf-AZ}
\showURL{%
\tempurl}


\bibitem[\protect\citeauthoryear{Lee, Cho, and Hofmann}{Lee
  et~al\mbox{.}}{2017}]%
        {lee17}
\bibfield{author}{\bibinfo{person}{Jason Lee}, \bibinfo{person}{Kyunghyun Cho},
  {and} \bibinfo{person}{Thomas Hofmann}.} \bibinfo{year}{2017}\natexlab{}.
\newblock \showarticletitle{Fully Character-Level Neural Machine Translation
  without Explicit Segmentation}.
\newblock \bibinfo{journal}{\emph{Transactions of the Association for
  Computational Linguistics}}  \bibinfo{volume}{5} (\bibinfo{year}{2017}),
  \bibinfo{pages}{365--378}.
\newblock
\urldef\tempurl%
\url{http://aclweb.org/anthology/Q17-1026}
\showURL{%
\tempurl}


\bibitem[\protect\citeauthoryear{Lu, Keung, Ladhak, Bhardwaj, Zhang, and
  Sun}{Lu et~al\mbox{.}}{2018}]%
        {lu18}
\bibfield{author}{\bibinfo{person}{Yichao Lu}, \bibinfo{person}{Phillip Keung},
  \bibinfo{person}{Faisal Ladhak}, \bibinfo{person}{Vikas Bhardwaj},
  \bibinfo{person}{Shaonan Zhang}, {and} \bibinfo{person}{Jason Sun}.}
  \bibinfo{year}{2018}\natexlab{}.
\newblock \showarticletitle{A neural interlingua for multilingual machine
  translation}. In \bibinfo{booktitle}{\emph{Proceedings of the Third
  Conference on Machine Translation: Research Papers}}.
  \bibinfo{publisher}{Association for Computational Linguistics},
  \bibinfo{address}{Belgium, Brussels}, \bibinfo{pages}{84--92}.
\newblock
\urldef\tempurl%
\url{http://aclweb.org/anthology/W18-6309}
\showURL{%
\tempurl}


\bibitem[\protect\citeauthoryear{Maimaiti, Liu, Luan, and Sun}{Maimaiti
  et~al\mbox{.}}{2019}]%
        {Maimaiti:2019:MTL:3327969.3314945}
\bibfield{author}{\bibinfo{person}{Mieradilijiang Maimaiti},
  \bibinfo{person}{Yang Liu}, \bibinfo{person}{Huanbo Luan}, {and}
  \bibinfo{person}{Maosong Sun}.} \bibinfo{year}{2019}\natexlab{}.
\newblock \showarticletitle{Multi-Round Transfer Learning for Low-Resource NMT
  Using Multiple High-Resource Languages}.
\newblock \bibinfo{journal}{\emph{ACM Trans. Asian Low-Resour. Lang. Inf.
  Process.}} \bibinfo{volume}{18}, \bibinfo{number}{4}, Article
  \bibinfo{articleno}{38} (\bibinfo{date}{May} \bibinfo{year}{2019}),
  \bibinfo{numpages}{26}~pages.
\newblock
\showISSN{2375-4699}
\urldef\tempurl%
\url{https://doi.org/10.1145/3314945}
\showDOI{\tempurl}


\bibitem[\protect\citeauthoryear{Mattoni, Nagle, Collantes, and
  Shterionov}{Mattoni et~al\mbox{.}}{2017}]%
        {Giulia-MTS2017}
\bibfield{author}{\bibinfo{person}{Giulia Mattoni}, \bibinfo{person}{Pat
  Nagle}, \bibinfo{person}{Carlos Collantes}, {and} \bibinfo{person}{Dimitar
  Shterionov}.} \bibinfo{year}{2017}\natexlab{}.
\newblock \showarticletitle{Zero-Shot Translation for Indian Languages with
  Sparse Data}. In \bibinfo{booktitle}{\emph{Proceedings of MT Summit XVI,
  Vol.2: Users and Translators Track}}. \bibinfo{address}{Nagoya, Jpaan},
  \bibinfo{pages}{1--10}.
\newblock


\bibitem[\protect\citeauthoryear{Matusov, Ueffing, and Ney}{Matusov
  et~al\mbox{.}}{2006}]%
        {matusov2006computing}
\bibfield{author}{\bibinfo{person}{Evgeny Matusov}, \bibinfo{person}{Nicola
  Ueffing}, {and} \bibinfo{person}{Hermann Ney}.}
  \bibinfo{year}{2006}\natexlab{}.
\newblock \showarticletitle{Computing Consensus Translation for Multiple
  Machine Translation Systems Using Enhanced Hypothesis Alignment}. In
  \bibinfo{booktitle}{\emph{11th Conference of the {E}uropean Chapter of the
  Association for Computational Linguistics}}. \bibinfo{pages}{33--40}.
\newblock
\urldef\tempurl%
\url{https://www.aclweb.org/anthology/E06-1005}
\showURL{%
\tempurl}


\bibitem[\protect\citeauthoryear{Mauro, Christian, and Marcello}{Mauro
  et~al\mbox{.}}{2012}]%
        {mauro2012wit3}
\bibfield{author}{\bibinfo{person}{Cettolo Mauro}, \bibinfo{person}{Girardi
  Christian}, {and} \bibinfo{person}{Federico Marcello}.}
  \bibinfo{year}{2012}\natexlab{}.
\newblock \showarticletitle{Wit3: Web inventory of transcribed and translated
  talks}. In \bibinfo{booktitle}{\emph{Proceedings of the 16th Conference of
  European Association for Machine Translation}}. \bibinfo{address}{Trento,
  Italy}, \bibinfo{pages}{261--268}.
\newblock


\bibitem[\protect\citeauthoryear{Mikolov, Le, and Sutskever}{Mikolov
  et~al\mbox{.}}{2013}]%
        {mikolov13a}
\bibfield{author}{\bibinfo{person}{Tomas Mikolov}, \bibinfo{person}{Quoc~V.
  Le}, {and} \bibinfo{person}{Ilya Sutskever}.}
  \bibinfo{year}{2013}\natexlab{}.
\newblock \showarticletitle{Exploiting Similarities among Languages for Machine
  Translation}.
\newblock \bibinfo{journal}{\emph{CoRR}}  \bibinfo{volume}{abs/1309.4168}
  (\bibinfo{year}{2013}).
\newblock
\showeprint[arxiv]{1309.4168}
\urldef\tempurl%
\url{http://arxiv.org/abs/1309.4168}
\showURL{%
\tempurl}


\bibitem[\protect\citeauthoryear{Murthy, Kunchukuttan, and
  Bhattacharyya}{Murthy et~al\mbox{.}}{2019}]%
        {rudramurthy19}
\bibfield{author}{\bibinfo{person}{Rudra Murthy}, \bibinfo{person}{Anoop
  Kunchukuttan}, {and} \bibinfo{person}{Pushpak Bhattacharyya}.}
  \bibinfo{year}{2019}\natexlab{}.
\newblock \showarticletitle{Addressing word-order Divergence in Multilingual
  Neural Machine Translation for extremely Low Resource Languages}. In
  \bibinfo{booktitle}{\emph{Proceedings of the 2019 Conference of the North
  {A}merican Chapter of the Association for Computational Linguistics: Human
  Language Technologies, Volume 1 (Long and Short Papers)}}.
  \bibinfo{publisher}{Association for Computational Linguistics},
  \bibinfo{address}{Minneapolis, Minnesota}, \bibinfo{pages}{3868--3873}.
\newblock
\urldef\tempurl%
\url{https://www.aclweb.org/anthology/N19-1387}
\showURL{%
\tempurl}


\bibitem[\protect\citeauthoryear{Nakazawa, Higashiyama, Ding, Dabre,
  Kunchukuttan, Pa, Goto, Mino, Sudoh, and Kurohashi}{Nakazawa
  et~al\mbox{.}}{2018}]%
        {nakazawa2018overview}
\bibfield{author}{\bibinfo{person}{Toshiaki Nakazawa}, \bibinfo{person}{Shohei
  Higashiyama}, \bibinfo{person}{Chenchen Ding}, \bibinfo{person}{Raj Dabre},
  \bibinfo{person}{Anoop Kunchukuttan}, \bibinfo{person}{Win~Pa Pa},
  \bibinfo{person}{Isao Goto}, \bibinfo{person}{Hideya Mino},
  \bibinfo{person}{Katsuhito Sudoh}, {and} \bibinfo{person}{Sadao Kurohashi}.}
  \bibinfo{year}{2018}\natexlab{}.
\newblock \showarticletitle{Overview of the 5th Workshop on Asian Translation}.
  In \bibinfo{booktitle}{\emph{Proceedings of the 5th Workshop on Asian
  Translation ({WAT}2018)}}. \bibinfo{address}{Hong Kong, China},
  \bibinfo{pages}{1--41}.
\newblock


\bibitem[\protect\citeauthoryear{Nakov and Ng}{Nakov and Ng}{2009}]%
        {nakov2009improved}
\bibfield{author}{\bibinfo{person}{Preslav Nakov} {and}
  \bibinfo{person}{Hwee~Tou Ng}.} \bibinfo{year}{2009}\natexlab{}.
\newblock \showarticletitle{Improved Statistical Machine Translation for
  Resource-Poor Languages Using Related Resource-Rich Languages}. In
  \bibinfo{booktitle}{\emph{Proceedings of the 2009 Conference on Empirical
  Methods in Natural Language Processing}}. \bibinfo{publisher}{Association for
  Computational Linguistics}, \bibinfo{address}{Singapore},
  \bibinfo{pages}{1358--1367}.
\newblock
\urldef\tempurl%
\url{https://www.aclweb.org/anthology/D09-1141}
\showURL{%
\tempurl}


\bibitem[\protect\citeauthoryear{Neubig}{Neubig}{2017}]%
        {neubig17nmt}
\bibfield{author}{\bibinfo{person}{Graham Neubig}.}
  \bibinfo{year}{2017}\natexlab{}.
\newblock \showarticletitle{Neural Machine Translation and Sequence-to-sequence
  Models: {A} Tutorial}.
\newblock \bibinfo{journal}{\emph{CoRR}}  \bibinfo{volume}{abs/1703.01619}
  (\bibinfo{year}{2017}).
\newblock
\showeprint[arxiv]{1703.01619}
\urldef\tempurl%
\url{http://arxiv.org/abs/1703.01619}
\showURL{%
\tempurl}


\bibitem[\protect\citeauthoryear{Neubig and Hu}{Neubig and Hu}{2018}]%
        {D18-1103}
\bibfield{author}{\bibinfo{person}{Graham Neubig} {and} \bibinfo{person}{Junjie
  Hu}.} \bibinfo{year}{2018}\natexlab{}.
\newblock \showarticletitle{Rapid Adaptation of Neural Machine Translation to
  New Languages}. In \bibinfo{booktitle}{\emph{Proceedings of the 2018
  Conference on Empirical Methods in Natural Language Processing}}.
  \bibinfo{publisher}{Association for Computational Linguistics},
  \bibinfo{address}{Brussels, Belgium}, \bibinfo{pages}{875--880}.
\newblock
\urldef\tempurl%
\url{http://aclweb.org/anthology/D18-1103}
\showURL{%
\tempurl}


\bibitem[\protect\citeauthoryear{Nguyen and Chiang}{Nguyen and Chiang}{2017}]%
        {nguyen17}
\bibfield{author}{\bibinfo{person}{Toan~Q. Nguyen} {and} \bibinfo{person}{David
  Chiang}.} \bibinfo{year}{2017}\natexlab{}.
\newblock \showarticletitle{Transfer Learning across Low-Resource, Related
  Languages for Neural Machine Translation}. In
  \bibinfo{booktitle}{\emph{Proceedings of the Eighth International Joint
  Conference on Natural Language Processing (Volume 2: Short Papers)}}.
  \bibinfo{publisher}{Asian Federation of Natural Language Processing},
  \bibinfo{address}{Taipei, Taiwan}, \bibinfo{pages}{296--301}.
\newblock
\urldef\tempurl%
\url{http://aclweb.org/anthology/I17-2050}
\showURL{%
\tempurl}


\bibitem[\protect\citeauthoryear{Nishimura, Sudoh, Neubig, and
  Nakamura}{Nishimura et~al\mbox{.}}{2018a}]%
        {nishimura18iwslt}
\bibfield{author}{\bibinfo{person}{Yuta Nishimura}, \bibinfo{person}{Katsuhito
  Sudoh}, \bibinfo{person}{Graham Neubig}, {and} \bibinfo{person}{Satoshi
  Nakamura}.} \bibinfo{year}{2018}\natexlab{a}.
\newblock \showarticletitle{Multi-Source Neural Machine Translation with Data
  Augmentation}. In \bibinfo{booktitle}{\emph{Proceedings of the 15th
  International Workshop on Spoken Language Translation ({IWSLT})}}.
  \bibinfo{address}{Brussels, Belgium}, \bibinfo{pages}{48--53}.
\newblock
\urldef\tempurl%
\url{https://arxiv.org/abs/1810.06826}
\showURL{%
\tempurl}


\bibitem[\protect\citeauthoryear{Nishimura, Sudoh, Neubig, and
  Nakamura}{Nishimura et~al\mbox{.}}{2018b}]%
        {W18-2711}
\bibfield{author}{\bibinfo{person}{Yuta Nishimura}, \bibinfo{person}{Katsuhito
  Sudoh}, \bibinfo{person}{Graham Neubig}, {and} \bibinfo{person}{Satoshi
  Nakamura}.} \bibinfo{year}{2018}\natexlab{b}.
\newblock \showarticletitle{Multi-Source Neural Machine Translation with
  Missing Data}. In \bibinfo{booktitle}{\emph{Proceedings of the 2nd Workshop
  on Neural Machine Translation and Generation}}.
  \bibinfo{publisher}{Association for Computational Linguistics},
  \bibinfo{address}{Melbourne, Australia}, \bibinfo{pages}{92--99}.
\newblock
\urldef\tempurl%
\url{http://aclweb.org/anthology/W18-2711}
\showURL{%
\tempurl}


\bibitem[\protect\citeauthoryear{Nyberg, Mitamura, and Carbonell}{Nyberg
  et~al\mbox{.}}{1997}]%
        {nyberg1997kant}
\bibfield{author}{\bibinfo{person}{Eric Nyberg}, \bibinfo{person}{Teruko
  Mitamura}, {and} \bibinfo{person}{Jaime Carbonell}.}
  \bibinfo{year}{1997}\natexlab{}.
\newblock \showarticletitle{The KANT Machine Translation System: From R\&D to
  Initial Deployment}. In \bibinfo{booktitle}{\emph{Proceedings of LISA
  Workshop on Integrating Advanced Translation Technology}}.
  \bibinfo{address}{Washington, D.C.}, \bibinfo{pages}{1--7}.
\newblock


\bibitem[\protect\citeauthoryear{Och and Ney}{Och and Ney}{2001}]%
        {och2001statistical}
\bibfield{author}{\bibinfo{person}{Franz~Josef Och} {and}
  \bibinfo{person}{Hermann Ney}.} \bibinfo{year}{2001}\natexlab{}.
\newblock \showarticletitle{Statistical multi-source translation}. In
  \bibinfo{booktitle}{\emph{Proceedings of MT Summit}},
  Vol.~\bibinfo{volume}{8}. \bibinfo{pages}{253--258}.
\newblock


\bibitem[\protect\citeauthoryear{Pan and Yang}{Pan and Yang}{2010}]%
        {Pan:2010:STL:1850483.1850545}
\bibfield{author}{\bibinfo{person}{Sinno~Jialin Pan} {and}
  \bibinfo{person}{Qiang Yang}.} \bibinfo{year}{2010}\natexlab{}.
\newblock \showarticletitle{A Survey on Transfer Learning}.
\newblock \bibinfo{journal}{\emph{IEEE Transactions on Knowledge and Data
  Engineering}} \bibinfo{volume}{22}, \bibinfo{number}{10}
  (\bibinfo{date}{Oct.} \bibinfo{year}{2010}), \bibinfo{pages}{1345--1359}.
\newblock
\showISSN{1041-4347}
\urldef\tempurl%
\url{https://doi.org/10.1109/TKDE.2009.191}
\showDOI{\tempurl}


\bibitem[\protect\citeauthoryear{Pham, Niehues, Ha, and Waibel}{Pham
  et~al\mbox{.}}{2019}]%
        {pham-etal-2019-improving}
\bibfield{author}{\bibinfo{person}{Ngoc-Quan Pham}, \bibinfo{person}{Jan
  Niehues}, \bibinfo{person}{Thanh-Le Ha}, {and} \bibinfo{person}{Alexander
  Waibel}.} \bibinfo{year}{2019}\natexlab{}.
\newblock \showarticletitle{Improving Zero-shot Translation with
  Language-Independent Constraints}. In \bibinfo{booktitle}{\emph{Proceedings
  of the Fourth Conference on Machine Translation (Volume 1: Research
  Papers)}}. \bibinfo{publisher}{Association for Computational Linguistics},
  \bibinfo{address}{Florence, Italy}, \bibinfo{pages}{13--23}.
\newblock
\urldef\tempurl%
\url{https://doi.org/10.18653/v1/W19-5202}
\showDOI{\tempurl}


\bibitem[\protect\citeauthoryear{Pires, Schlinger, and Garrette}{Pires
  et~al\mbox{.}}{2019}]%
        {pires-etal-2019-multilingual}
\bibfield{author}{\bibinfo{person}{Telmo Pires}, \bibinfo{person}{Eva
  Schlinger}, {and} \bibinfo{person}{Dan Garrette}.}
  \bibinfo{year}{2019}\natexlab{}.
\newblock \showarticletitle{How Multilingual is Multilingual {BERT}?}. In
  \bibinfo{booktitle}{\emph{Proceedings of the 57th Annual Meeting of the
  Association for Computational Linguistics}}.
\newblock


\bibitem[\protect\citeauthoryear{Platanios, Sachan, Neubig, and
  Mitchell}{Platanios et~al\mbox{.}}{2018}]%
        {platanios18}
\bibfield{author}{\bibinfo{person}{Emmanouil~Antonios Platanios},
  \bibinfo{person}{Mrinmaya Sachan}, \bibinfo{person}{Graham Neubig}, {and}
  \bibinfo{person}{Tom Mitchell}.} \bibinfo{year}{2018}\natexlab{}.
\newblock \showarticletitle{Contextual Parameter Generation for Universal
  Neural Machine Translation}. In \bibinfo{booktitle}{\emph{Proceedings of the
  2018 Conference on Empirical Methods in Natural Language Processing}}.
  \bibinfo{publisher}{Association for Computational Linguistics},
  \bibinfo{address}{Brussels, Belgium}, \bibinfo{pages}{425--435}.
\newblock
\urldef\tempurl%
\url{http://aclweb.org/anthology/D18-1039}
\showURL{%
\tempurl}


\bibitem[\protect\citeauthoryear{Prasanna}{Prasanna}{2018}]%
        {PraR:2018}
\bibfield{author}{\bibinfo{person}{Raj Noel~Dabre Prasanna}.}
  \bibinfo{year}{2018}\natexlab{}.
\newblock \emph{\bibinfo{title}{Exploiting Multilingualism and Transfer
  Learning for Low Resource Machine Translation}}.
\newblock \bibinfo{thesistype}{Ph.D. Dissertation}. \bibinfo{school}{Kyoto
  University}.
\newblock
\urldef\tempurl%
\url{http://hdl.handle.net/2433/232411}
\showURL{%
\tempurl}


\bibitem[\protect\citeauthoryear{Raghu, Gilmer, Yosinski, and
  Sohl-Dickstein}{Raghu et~al\mbox{.}}{2017}]%
        {raghu2017svcca}
\bibfield{author}{\bibinfo{person}{Maithra Raghu}, \bibinfo{person}{Justin
  Gilmer}, \bibinfo{person}{Jason Yosinski}, {and} \bibinfo{person}{Jascha
  Sohl-Dickstein}.} \bibinfo{year}{2017}\natexlab{}.
\newblock \showarticletitle{SVCCA: Singular Vector Canonical Correlation
  Analysis for Deep Learning Dynamics and Interpretability}.
\newblock In \bibinfo{booktitle}{\emph{Advances in Neural Information
  Processing Systems 30}}, \bibfield{editor}{\bibinfo{person}{I.~Guyon},
  \bibinfo{person}{U.~V. Luxburg}, \bibinfo{person}{S.~Bengio},
  \bibinfo{person}{H.~Wallach}, \bibinfo{person}{R.~Fergus},
  \bibinfo{person}{S.~Vishwanathan}, {and} \bibinfo{person}{R.~Garnett}}
  (Eds.). \bibinfo{publisher}{Curran Associates, Inc.},
  \bibinfo{pages}{6076--6085}.
\newblock
\urldef\tempurl%
\url{http://papers.nips.cc/paper/7188-svcca-singular-vector-canonical-correlation-analysis-for-deep-learning-dynamics-and-interpretability.pdf}
\showURL{%
\tempurl}


\bibitem[\protect\citeauthoryear{Ramachandran, Liu, and Le}{Ramachandran
  et~al\mbox{.}}{2017}]%
        {ramachandran17}
\bibfield{author}{\bibinfo{person}{Prajit Ramachandran}, \bibinfo{person}{Peter
  Liu}, {and} \bibinfo{person}{Quoc Le}.} \bibinfo{year}{2017}\natexlab{}.
\newblock \showarticletitle{Unsupervised Pretraining for Sequence to Sequence
  Learning}. In \bibinfo{booktitle}{\emph{Proceedings of the 2017 Conference on
  Empirical Methods in Natural Language Processing}}.
  \bibinfo{publisher}{Association for Computational Linguistics},
  \bibinfo{address}{Copenhagen, Denmark}, \bibinfo{pages}{383--391}.
\newblock
\urldef\tempurl%
\url{https://doi.org/10.18653/v1/D17-1039}
\showDOI{\tempurl}


\bibitem[\protect\citeauthoryear{Rikters, Pinnis, and Krišlauks}{Rikters
  et~al\mbox{.}}{2018}]%
        {rikters18}
\bibfield{author}{\bibinfo{person}{Matīss Rikters}, \bibinfo{person}{Mārcis
  Pinnis}, {and} \bibinfo{person}{Rihards Krišlauks}.}
  \bibinfo{year}{2018}\natexlab{}.
\newblock \showarticletitle{{Training and Adapting Multilingual NMT for
  Less-resourced and Morphologically Rich Languages}}. In
  \bibinfo{booktitle}{\emph{Proceedings of the Eleventh International
  Conference on Language Resources and Evaluation (LREC 2018)}}.
  \bibinfo{publisher}{European Language Resources Association (ELRA)},
  \bibinfo{address}{Miyazaki, Japan}, \bibinfo{pages}{3766--3773}.
\newblock
\showISBNx{979-10-95546-00-9}


\bibitem[\protect\citeauthoryear{Ruder}{Ruder}{2016}]%
        {DBLP:journals/corr/Ruder16}
\bibfield{author}{\bibinfo{person}{Sebastian Ruder}.}
  \bibinfo{year}{2016}\natexlab{}.
\newblock \showarticletitle{An overview of gradient descent optimization
  algorithms}.
\newblock \bibinfo{journal}{\emph{CoRR}}  \bibinfo{volume}{abs/1609.04747}
  (\bibinfo{year}{2016}).
\newblock
\showeprint[arxiv]{1609.04747}
\urldef\tempurl%
\url{http://arxiv.org/abs/1609.04747}
\showURL{%
\tempurl}


\bibitem[\protect\citeauthoryear{Sachan and Neubig}{Sachan and Neubig}{2018}]%
        {sachan18}
\bibfield{author}{\bibinfo{person}{Devendra Sachan} {and}
  \bibinfo{person}{Graham Neubig}.} \bibinfo{year}{2018}\natexlab{}.
\newblock \showarticletitle{Parameter Sharing Methods for Multilingual
  Self-Attentional Translation Models}. In
  \bibinfo{booktitle}{\emph{Proceedings of the Third Conference on Machine
  Translation: Research Papers}}. \bibinfo{publisher}{Association for
  Computational Linguistics}, \bibinfo{address}{Belgium, Brussels},
  \bibinfo{pages}{261--271}.
\newblock
\urldef\tempurl%
\url{http://aclweb.org/anthology/W18-6327}
\showURL{%
\tempurl}


\bibitem[\protect\citeauthoryear{Saha, Khapra, Chandar, Rajendran, and
  Cho}{Saha et~al\mbox{.}}{2016}]%
        {saha2016correlational}
\bibfield{author}{\bibinfo{person}{Amrita Saha}, \bibinfo{person}{Mitesh~M.
  Khapra}, \bibinfo{person}{Sarath Chandar}, \bibinfo{person}{Janarthanan
  Rajendran}, {and} \bibinfo{person}{Kyunghyun Cho}.}
  \bibinfo{year}{2016}\natexlab{}.
\newblock \showarticletitle{A Correlational Encoder Decoder Architecture for
  Pivot Based Sequence Generation}. In \bibinfo{booktitle}{\emph{Proceedings of
  {COLING} 2016, the 26th International Conference on Computational
  Linguistics: Technical Papers}}. \bibinfo{publisher}{The COLING 2016
  Organizing Committee}, \bibinfo{address}{Osaka, Japan},
  \bibinfo{pages}{109--118}.
\newblock
\urldef\tempurl%
\url{https://www.aclweb.org/anthology/C16-1011}
\showURL{%
\tempurl}


\bibitem[\protect\citeauthoryear{Sch{\"o}nemann}{Sch{\"o}nemann}{1966}]%
        {schonemann1966generalized}
\bibfield{author}{\bibinfo{person}{Peter~H Sch{\"o}nemann}.}
  \bibinfo{year}{1966}\natexlab{}.
\newblock \showarticletitle{A generalized solution of the orthogonal procrustes
  problem}.
\newblock \bibinfo{journal}{\emph{Psychometrika}} \bibinfo{volume}{31},
  \bibinfo{number}{1} (\bibinfo{year}{1966}), \bibinfo{pages}{1--10}.
\newblock


\bibitem[\protect\citeauthoryear{Schroeder, Cohn, and Koehn}{Schroeder
  et~al\mbox{.}}{2009}]%
        {schroeder2009word}
\bibfield{author}{\bibinfo{person}{Josh Schroeder}, \bibinfo{person}{Trevor
  Cohn}, {and} \bibinfo{person}{Philipp Koehn}.}
  \bibinfo{year}{2009}\natexlab{}.
\newblock \showarticletitle{Word Lattices for Multi-Source Translation}. In
  \bibinfo{booktitle}{\emph{Proceedings of the 12th Conference of the
  {E}uropean Chapter of the {ACL} ({EACL} 2009)}}.
  \bibinfo{publisher}{Association for Computational Linguistics},
  \bibinfo{address}{Athens, Greece}, \bibinfo{pages}{719--727}.
\newblock
\urldef\tempurl%
\url{https://www.aclweb.org/anthology/E09-1082}
\showURL{%
\tempurl}


\bibitem[\protect\citeauthoryear{Schuster and Nakajima}{Schuster and
  Nakajima}{2012}]%
        {conf/icassp/SchusterN12}
\bibfield{author}{\bibinfo{person}{Mike Schuster} {and}
  \bibinfo{person}{Kaisuke Nakajima}.} \bibinfo{year}{2012}\natexlab{}.
\newblock \showarticletitle{Japanese and Korean voice search.}. In
  \bibinfo{booktitle}{\emph{ICASSP}}. \bibinfo{publisher}{IEEE},
  \bibinfo{pages}{5149--5152}.
\newblock
\showISBNx{978-1-4673-0046-9}
\urldef\tempurl%
\url{http://dblp.uni-trier.de/db/conf/icassp/icassp2012.html#SchusterN12}
\showURL{%
\tempurl}


\bibitem[\protect\citeauthoryear{Schwenk, Chaudhary, Sun, Gong, and
  Guzm{\'{a}}n}{Schwenk et~al\mbox{.}}{2019}]%
        {DBLP:journals/corr/abs-1907-05791}
\bibfield{author}{\bibinfo{person}{Holger Schwenk}, \bibinfo{person}{Vishrav
  Chaudhary}, \bibinfo{person}{Shuo Sun}, \bibinfo{person}{Hongyu Gong}, {and}
  \bibinfo{person}{Francisco Guzm{\'{a}}n}.} \bibinfo{year}{2019}\natexlab{}.
\newblock \showarticletitle{WikiMatrix: Mining 135M Parallel Sentences in 1620
  Language Pairs from Wikipedia}.
\newblock \bibinfo{journal}{\emph{CoRR}}  \bibinfo{volume}{abs/1907.05791}
  (\bibinfo{year}{2019}).
\newblock
\showeprint[arxiv]{1907.05791}
\urldef\tempurl%
\url{http://arxiv.org/abs/1907.05791}
\showURL{%
\tempurl}


\bibitem[\protect\citeauthoryear{Sennrich, Haddow, and Birch}{Sennrich
  et~al\mbox{.}}{2016a}]%
        {sennrich-haddow-birch:2016:P16-11}
\bibfield{author}{\bibinfo{person}{Rico Sennrich}, \bibinfo{person}{Barry
  Haddow}, {and} \bibinfo{person}{Alexandra Birch}.}
  \bibinfo{year}{2016}\natexlab{a}.
\newblock \showarticletitle{Improving Neural Machine Translation Models with
  Monolingual Data}. In \bibinfo{booktitle}{\emph{Proceedings of the 54th
  Annual Meeting of the Association for Computational Linguistics (Volume 1:
  Long Papers)}}. \bibinfo{publisher}{Association for Computational
  Linguistics}, \bibinfo{address}{Berlin, Germany}, \bibinfo{pages}{86--96}.
\newblock
\urldef\tempurl%
\url{http://www.aclweb.org/anthology/P16-1009}
\showURL{%
\tempurl}


\bibitem[\protect\citeauthoryear{Sennrich, Haddow, and Birch}{Sennrich
  et~al\mbox{.}}{2016b}]%
        {DBLP:journals/corr/SennrichHB15}
\bibfield{author}{\bibinfo{person}{Rico Sennrich}, \bibinfo{person}{Barry
  Haddow}, {and} \bibinfo{person}{Alexandra Birch}.}
  \bibinfo{year}{2016}\natexlab{b}.
\newblock \showarticletitle{Neural Machine Translation of Rare Words with
  Subword Units}. In \bibinfo{booktitle}{\emph{Proceedings of the 54th Annual
  Meeting of the Association for Computational Linguistics (Volume 1: Long
  Papers)}}. \bibinfo{publisher}{Association for Computational Linguistics},
  \bibinfo{address}{Berlin, Germany}, \bibinfo{pages}{1715--1725}.
\newblock
\urldef\tempurl%
\url{http://www.aclweb.org/anthology/P16-1162}
\showURL{%
\tempurl}


\bibitem[\protect\citeauthoryear{Sestorain, Ciaramita, Buck, and
  Hofmann}{Sestorain et~al\mbox{.}}{2018}]%
        {DBLP:journals/corr/abs-1805-10338}
\bibfield{author}{\bibinfo{person}{Lierni Sestorain},
  \bibinfo{person}{Massimiliano Ciaramita}, \bibinfo{person}{Christian Buck},
  {and} \bibinfo{person}{Thomas Hofmann}.} \bibinfo{year}{2018}\natexlab{}.
\newblock \showarticletitle{Zero-Shot Dual Machine Translation}.
\newblock \bibinfo{journal}{\emph{CoRR}}  \bibinfo{volume}{abs/1805.10338}
  (\bibinfo{year}{2018}).
\newblock
\showeprint[arxiv]{1805.10338}
\urldef\tempurl%
\url{http://arxiv.org/abs/1805.10338}
\showURL{%
\tempurl}


\bibitem[\protect\citeauthoryear{Sgall and Panevov\'{a}}{Sgall and
  Panevov\'{a}}{1987}]%
        {Sgall:1987:MTL:976858.976876}
\bibfield{author}{\bibinfo{person}{Petr Sgall} {and} \bibinfo{person}{Jarmila
  Panevov\'{a}}.} \bibinfo{year}{1987}\natexlab{}.
\newblock \showarticletitle{Machine Translation, Linguistics, and Interlingua}.
  In \bibinfo{booktitle}{\emph{Proceedings of the Third Conference on European
  Chapter of the Association for Computational Linguistics}}
  \emph{(\bibinfo{series}{EACL '87})}. \bibinfo{publisher}{Association for
  Computational Linguistics}, \bibinfo{address}{Copenhagen, Denmark},
  \bibinfo{pages}{99--103}.
\newblock
\urldef\tempurl%
\url{https://doi.org/10.3115/976858.976876}
\showDOI{\tempurl}


\bibitem[\protect\citeauthoryear{Siddhant, Johnson, Tsai, Arivazhagan, Riesa,
  Bapna, Firat, and Raman}{Siddhant et~al\mbox{.}}{2020}]%
        {Siddhant-AAAI2020}
\bibfield{author}{\bibinfo{person}{Aditya Siddhant}, \bibinfo{person}{Melvin
  Johnson}, \bibinfo{person}{Henry Tsai}, \bibinfo{person}{Naveen Arivazhagan},
  \bibinfo{person}{Jason Riesa}, \bibinfo{person}{Ankur Bapna},
  \bibinfo{person}{Orhan Firat}, {and} \bibinfo{person}{Karthik Raman}.}
  \bibinfo{year}{2020}\natexlab{}.
\newblock \showarticletitle{Evaluating the Cross-Lingual Effectiveness of
  Massively Multilingual Neural Machine Translation}. In
  \bibinfo{booktitle}{\emph{AAAI}}.
\newblock


\bibitem[\protect\citeauthoryear{{S{\o}gaard}, {Ruder}, and
  {Vuli{\'c}}}{{S{\o}gaard} et~al\mbox{.}}{2018}]%
        {sogaard2018limitations}
\bibfield{author}{\bibinfo{person}{Anders {S{\o}gaard}},
  \bibinfo{person}{Sebastian {Ruder}}, {and} \bibinfo{person}{Ivan
  {Vuli{\'c}}}.} \bibinfo{year}{2018}\natexlab{}.
\newblock \showarticletitle{{On the Limitations of Unsupervised Bilingual
  Dictionary Induction}}. In \bibinfo{booktitle}{\emph{ACL}}.
\newblock


\bibitem[\protect\citeauthoryear{Song, Zhang, Yu, Luo, Wang, and Zhang}{Song
  et~al\mbox{.}}{2019}]%
        {song-etal-2019-code}
\bibfield{author}{\bibinfo{person}{Kai Song}, \bibinfo{person}{Yue Zhang},
  \bibinfo{person}{Heng Yu}, \bibinfo{person}{Weihua Luo}, \bibinfo{person}{Kun
  Wang}, {and} \bibinfo{person}{Min Zhang}.} \bibinfo{year}{2019}\natexlab{}.
\newblock \showarticletitle{Code-Switching for Enhancing {NMT} with
  Pre-Specified Translation}. In \bibinfo{booktitle}{\emph{Proceedings of the
  2019 Conference of the North {A}merican Chapter of the Association for
  Computational Linguistics: Human Language Technologies, Volume 1 (Long and
  Short Papers)}}. \bibinfo{publisher}{Association for Computational
  Linguistics}, \bibinfo{address}{Minneapolis, Minnesota},
  \bibinfo{pages}{449--459}.
\newblock
\urldef\tempurl%
\url{https://www.aclweb.org/anthology/N19-1044}
\showURL{%
\tempurl}


\bibitem[\protect\citeauthoryear{Steinberger, Ebrahim, Poulis,
  Carrasco-Benitez, Schl{\"u}ter, Przybyszewski, and Gilbro}{Steinberger
  et~al\mbox{.}}{2014}]%
        {steinberger2014overview}
\bibfield{author}{\bibinfo{person}{Ralf Steinberger}, \bibinfo{person}{Mohamed
  Ebrahim}, \bibinfo{person}{Alexandros Poulis}, \bibinfo{person}{Manuel
  Carrasco-Benitez}, \bibinfo{person}{Patrick Schl{\"u}ter},
  \bibinfo{person}{Marek Przybyszewski}, {and} \bibinfo{person}{Signe Gilbro}.}
  \bibinfo{year}{2014}\natexlab{}.
\newblock \showarticletitle{{An overview of the European Union’s highly
  multilingual parallel corpora}}.
\newblock \bibinfo{journal}{\emph{Language Resources and Evaluation}}
  \bibinfo{volume}{48}, \bibinfo{number}{4} (\bibinfo{year}{2014}),
  \bibinfo{pages}{679--707}.
\newblock


\bibitem[\protect\citeauthoryear{Sutskever, Vinyals, and Le}{Sutskever
  et~al\mbox{.}}{2014}]%
        {DBLP:journals/corr/SutskeverVL14}
\bibfield{author}{\bibinfo{person}{Ilya Sutskever}, \bibinfo{person}{Oriol
  Vinyals}, {and} \bibinfo{person}{Quoc~V. Le}.}
  \bibinfo{year}{2014}\natexlab{}.
\newblock \showarticletitle{Sequence to Sequence Learning with Neural
  Networks}. In \bibinfo{booktitle}{\emph{Proceedings of the 27th International
  Conference on Neural Information Processing Systems}}
  \emph{(\bibinfo{series}{NIPS'14})}. \bibinfo{publisher}{MIT Press},
  \bibinfo{address}{Montreal, Canada}, \bibinfo{pages}{3104--3112}.
\newblock
\urldef\tempurl%
\url{http://dl.acm.org/citation.cfm?id=2969033.2969173}
\showURL{%
\tempurl}


\bibitem[\protect\citeauthoryear{Tan, Chen, He, Xia, Qin, and Liu}{Tan
  et~al\mbox{.}}{2019a}]%
        {tan-etal-2019-multilingual}
\bibfield{author}{\bibinfo{person}{Xu Tan}, \bibinfo{person}{Jiale Chen},
  \bibinfo{person}{Di He}, \bibinfo{person}{Yingce Xia}, \bibinfo{person}{Tao
  Qin}, {and} \bibinfo{person}{Tie-Yan Liu}.} \bibinfo{year}{2019}\natexlab{a}.
\newblock \showarticletitle{Multilingual Neural Machine Translation with
  Language Clustering}. In \bibinfo{booktitle}{\emph{Proceedings of the 2019
  Conference on Empirical Methods in Natural Language Processing and the 9th
  International Joint Conference on Natural Language Processing
  (EMNLP-IJCNLP)}}. \bibinfo{publisher}{Association for Computational
  Linguistics}, \bibinfo{address}{Hong Kong, China}, \bibinfo{pages}{963--973}.
\newblock
\urldef\tempurl%
\url{https://doi.org/10.18653/v1/D19-1089}
\showDOI{\tempurl}


\bibitem[\protect\citeauthoryear{Tan, Ren, He, Qin, and Liu}{Tan
  et~al\mbox{.}}{2019b}]%
        {tan18}
\bibfield{author}{\bibinfo{person}{Xu Tan}, \bibinfo{person}{Yi Ren},
  \bibinfo{person}{Di He}, \bibinfo{person}{Tao Qin}, {and}
  \bibinfo{person}{Tie-Yan Liu}.} \bibinfo{year}{2019}\natexlab{b}.
\newblock \showarticletitle{Multilingual Neural Machine Translation with
  Knowledge Distillation}. In \bibinfo{booktitle}{\emph{Proceedings of
  International Conference on Learning Representations}}. \bibinfo{address}{New
  Orleans}.
\newblock


\bibitem[\protect\citeauthoryear{Thu, Pa, Utiyama, Finch, and Sumita}{Thu
  et~al\mbox{.}}{2016}]%
        {thu2016introducing}
\bibfield{author}{\bibinfo{person}{Ye~Kyaw Thu}, \bibinfo{person}{Win~Pa Pa},
  \bibinfo{person}{Masao Utiyama}, \bibinfo{person}{Andrew~M Finch}, {and}
  \bibinfo{person}{Eiichiro Sumita}.} \bibinfo{year}{2016}\natexlab{}.
\newblock \showarticletitle{Introducing the Asian Language Treebank ({ALT}).}.
  In \bibinfo{booktitle}{\emph{Proceedings of the Tenth International
  Conference on Language Resources and Evaluation (LREC 2016)}} (23-28).
  \bibinfo{publisher}{European Language Resources Association (ELRA)},
  \bibinfo{address}{Portorož, Slovenia}, \bibinfo{pages}{1574--1578}.
\newblock


\bibitem[\protect\citeauthoryear{Tiedemann}{Tiedemann}{2012a}]%
        {tiedemann2012character}
\bibfield{author}{\bibinfo{person}{J{\"o}rg Tiedemann}.}
  \bibinfo{year}{2012}\natexlab{a}.
\newblock \showarticletitle{Character-Based Pivot Translation for
  Under-Resourced Languages and Domains}. In
  \bibinfo{booktitle}{\emph{Proceedings of the 13th Conference of the
  {E}uropean Chapter of the Association for Computational Linguistics}}.
  \bibinfo{publisher}{Association for Computational Linguistics},
  \bibinfo{address}{Avignon, France}, \bibinfo{pages}{141--151}.
\newblock
\urldef\tempurl%
\url{https://www.aclweb.org/anthology/E12-1015}
\showURL{%
\tempurl}


\bibitem[\protect\citeauthoryear{Tiedemann}{Tiedemann}{2012b}]%
        {TIEDEMANN12.463}
\bibfield{author}{\bibinfo{person}{J{\"o}rg Tiedemann}.}
  \bibinfo{year}{2012}\natexlab{b}.
\newblock \showarticletitle{Parallel Data, Tools and Interfaces in OPUS}. In
  \bibinfo{booktitle}{\emph{Proceedings of the Eight International Conference
  on Language Resources and Evaluation (LREC'12)}} (23-25).
  \bibinfo{publisher}{European Language Resources Association (ELRA)},
  \bibinfo{address}{Istanbul, Turkey}.
\newblock
\showISBNx{978-2-9517408-7-7}


\bibitem[\protect\citeauthoryear{Uchida}{Uchida}{1996}]%
        {uchida1996unl}
\bibfield{author}{\bibinfo{person}{Hiroshi Uchida}.}
  \bibinfo{year}{1996}\natexlab{}.
\newblock \showarticletitle{{UNL: Universal Networking Language – An
  Electronic Language for Communication, Understanding, and Collaboration}}. In
  \bibinfo{booktitle}{\emph{UNU/IAS/UNL Center}}.
\newblock


\bibitem[\protect\citeauthoryear{Utiyama and Isahara}{Utiyama and
  Isahara}{2007}]%
        {utiyama2007}
\bibfield{author}{\bibinfo{person}{Masao Utiyama} {and}
  \bibinfo{person}{Hitoshi Isahara}.} \bibinfo{year}{2007}\natexlab{}.
\newblock \showarticletitle{A Comparison of Pivot Methods for Phrase-Based
  Statistical Machine Translation}. In \bibinfo{booktitle}{\emph{Human Language
  Technologies 2007: The Conference of the North {A}merican Chapter of the
  Association for Computational Linguistics; Proceedings of the Main
  Conference}}. \bibinfo{publisher}{Association for Computational Linguistics},
  \bibinfo{address}{Rochester, New York}, \bibinfo{pages}{484--491}.
\newblock
\urldef\tempurl%
\url{https://www.aclweb.org/anthology/N07-1061}
\showURL{%
\tempurl}


\bibitem[\protect\citeauthoryear{Vaswani, Shazeer, Parmar, Uszkoreit, Jones,
  Gomez, Kaiser, and Polosukhin}{Vaswani et~al\mbox{.}}{2017}]%
        {NIPS2017_7181}
\bibfield{author}{\bibinfo{person}{Ashish Vaswani}, \bibinfo{person}{Noam
  Shazeer}, \bibinfo{person}{Niki Parmar}, \bibinfo{person}{Jakob Uszkoreit},
  \bibinfo{person}{Llion Jones}, \bibinfo{person}{Aidan~N Gomez},
  \bibinfo{person}{\L~ukasz Kaiser}, {and} \bibinfo{person}{Illia Polosukhin}.}
  \bibinfo{year}{2017}\natexlab{}.
\newblock \showarticletitle{Attention is All you Need}.
\newblock In \bibinfo{booktitle}{\emph{Proceedings of the Advances in Neural
  Information Processing Systems 30}},
  \bibfield{editor}{\bibinfo{person}{I.~Guyon}, \bibinfo{person}{U.~V.
  Luxburg}, \bibinfo{person}{S.~Bengio}, \bibinfo{person}{H.~Wallach},
  \bibinfo{person}{R.~Fergus}, \bibinfo{person}{S.~Vishwanathan}, {and}
  \bibinfo{person}{R.~Garnett}} (Eds.). \bibinfo{publisher}{Curran Associates,
  Inc.}, \bibinfo{pages}{5998--6008}.
\newblock
\urldef\tempurl%
\url{http://papers.nips.cc/paper/7181-attention-is-all-you-need.pdf}
\showURL{%
\tempurl}


\bibitem[\protect\citeauthoryear{V{\'{a}}zquez, Raganato, Tiedemann, and
  Creutz}{V{\'{a}}zquez et~al\mbox{.}}{2018}]%
        {vazquez18}
\bibfield{author}{\bibinfo{person}{Ra{\'{u}}l V{\'{a}}zquez},
  \bibinfo{person}{Alessandro Raganato}, \bibinfo{person}{J{\"{o}}rg
  Tiedemann}, {and} \bibinfo{person}{Mathias Creutz}.}
  \bibinfo{year}{2018}\natexlab{}.
\newblock \showarticletitle{Multilingual {NMT} with a language-independent
  attention bridge}.
\newblock \bibinfo{journal}{\emph{CoRR}}  \bibinfo{volume}{abs/1811.00498}
  (\bibinfo{year}{2018}).
\newblock
\showeprint[arxiv]{1811.00498}
\urldef\tempurl%
\url{http://arxiv.org/abs/1811.00498}
\showURL{%
\tempurl}


\bibitem[\protect\citeauthoryear{Vilar, Peter, and Ney}{Vilar
  et~al\mbox{.}}{2007}]%
        {vilar2007can}
\bibfield{author}{\bibinfo{person}{David Vilar}, \bibinfo{person}{Jan-Thorsten
  Peter}, {and} \bibinfo{person}{Hermann Ney}.}
  \bibinfo{year}{2007}\natexlab{}.
\newblock \showarticletitle{Can We Translate Letters?}. In
  \bibinfo{booktitle}{\emph{Proceedings of the Second Workshop on Statistical
  Machine Translation}}. \bibinfo{publisher}{Association for Computational
  Linguistics}, \bibinfo{address}{Prague, Czech Republic},
  \bibinfo{pages}{33--39}.
\newblock
\urldef\tempurl%
\url{https://www.aclweb.org/anthology/W07-0705}
\showURL{%
\tempurl}


\bibitem[\protect\citeauthoryear{Wang, Finch, Utiyama, and Sumita}{Wang
  et~al\mbox{.}}{2017a}]%
        {wang-EtAl:2017:Short3}
\bibfield{author}{\bibinfo{person}{Rui Wang}, \bibinfo{person}{Andrew Finch},
  \bibinfo{person}{Masao Utiyama}, {and} \bibinfo{person}{Eiichiro Sumita}.}
  \bibinfo{year}{2017}\natexlab{a}.
\newblock \showarticletitle{Sentence Embedding for Neural Machine Translation
  Domain Adaptation}. In \bibinfo{booktitle}{\emph{Proceedings of the 55th
  Annual Meeting of the Association for Computational Linguistics (Volume 2:
  Short Papers)}}. \bibinfo{publisher}{Association for Computational
  Linguistics}, \bibinfo{address}{Vancouver, Canada},
  \bibinfo{pages}{560--566}.
\newblock
\urldef\tempurl%
\url{http://aclweb.org/anthology/P17-2089}
\showURL{%
\tempurl}


\bibitem[\protect\citeauthoryear{Wang, Utiyama, Liu, Chen, and Sumita}{Wang
  et~al\mbox{.}}{2017b}]%
        {iwnmt}
\bibfield{author}{\bibinfo{person}{Rui Wang}, \bibinfo{person}{Masao Utiyama},
  \bibinfo{person}{Lemao Liu}, \bibinfo{person}{Kehai Chen}, {and}
  \bibinfo{person}{Eiichiro Sumita}.} \bibinfo{year}{2017}\natexlab{b}.
\newblock \showarticletitle{Instance Weighting for Neural Machine Translation
  Domain Adaptation}. In \bibinfo{booktitle}{\emph{Proceedings of the 2017
  Conference on Empirical Methods in Natural Language Processing}}.
  \bibinfo{address}{Copenhagen, Denmark}, \bibinfo{pages}{1482--1488}.
\newblock


\bibitem[\protect\citeauthoryear{Wang and Neubig}{Wang and Neubig}{2019}]%
        {wang19}
\bibfield{author}{\bibinfo{person}{Xinyi Wang} {and} \bibinfo{person}{Graham
  Neubig}.} \bibinfo{year}{2019}\natexlab{}.
\newblock \showarticletitle{Target Conditioned Sampling: Optimizing Data
  Selection for Multilingual Neural Machine Translation}. In
  \bibinfo{booktitle}{\emph{Proceedings of the 57th Annual Meeting of the
  Association for Computational Linguistics}}.
\newblock


\bibitem[\protect\citeauthoryear{Wang, Pham, Arthur, and Neubig}{Wang
  et~al\mbox{.}}{2019a}]%
        {wang-ICLR2019}
\bibfield{author}{\bibinfo{person}{Xinyi Wang}, \bibinfo{person}{Hieu Pham},
  \bibinfo{person}{Philip Arthur}, {and} \bibinfo{person}{Graham Neubig}.}
  \bibinfo{year}{2019}\natexlab{a}.
\newblock \showarticletitle{Multilingual Neural Machine Translation With Soft
  Decoupled Encoding}. In \bibinfo{booktitle}{\emph{Proceedings of
  International Conference on Learning Representations}}. \bibinfo{address}{New
  Orleans}.
\newblock
\urldef\tempurl%
\url{https://openreview.net/forum?id=Skeke3C5Fm}
\showURL{%
\tempurl}


\bibitem[\protect\citeauthoryear{Wang, Zhang, Zhai, Xu, and Zong}{Wang
  et~al\mbox{.}}{2018}]%
        {wang18}
\bibfield{author}{\bibinfo{person}{Yining Wang}, \bibinfo{person}{Jiajun
  Zhang}, \bibinfo{person}{Feifei Zhai}, \bibinfo{person}{Jingfang Xu}, {and}
  \bibinfo{person}{Chengqing Zong}.} \bibinfo{year}{2018}\natexlab{}.
\newblock \showarticletitle{Three Strategies to Improve One-to-Many
  Multilingual Translation}. In \bibinfo{booktitle}{\emph{Proceedings of the
  2018 Conference on Empirical Methods in Natural Language Processing}}.
  \bibinfo{publisher}{Association for Computational Linguistics},
  \bibinfo{address}{Brussels, Belgium}, \bibinfo{pages}{2955--2960}.
\newblock
\urldef\tempurl%
\url{http://aclweb.org/anthology/D18-1326}
\showURL{%
\tempurl}


\bibitem[\protect\citeauthoryear{Wang, Zhou, Zhang, Zhai, Xu, and Zong}{Wang
  et~al\mbox{.}}{2019b}]%
        {wang-etal-2019-compact}
\bibfield{author}{\bibinfo{person}{Yining Wang}, \bibinfo{person}{Long Zhou},
  \bibinfo{person}{Jiajun Zhang}, \bibinfo{person}{Feifei Zhai},
  \bibinfo{person}{Jingfang Xu}, {and} \bibinfo{person}{Chengqing Zong}.}
  \bibinfo{year}{2019}\natexlab{b}.
\newblock \showarticletitle{A Compact and Language-Sensitive Multilingual
  Translation Method}. In \bibinfo{booktitle}{\emph{Proceedings of the 57th
  Annual Meeting of the Association for Computational Linguistics}}.
  \bibinfo{publisher}{Association for Computational Linguistics},
  \bibinfo{address}{Florence, Italy}, \bibinfo{pages}{1213--1223}.
\newblock
\urldef\tempurl%
\url{https://doi.org/10.18653/v1/P19-1117}
\showDOI{\tempurl}


\bibitem[\protect\citeauthoryear{Witkam}{Witkam}{2006}]%
        {witkam2006dlt}
\bibfield{author}{\bibinfo{person}{Toon Witkam}.}
  \bibinfo{year}{2006}\natexlab{}.
\newblock \showarticletitle{{History and Heritage of the DLT (Distributed
  Language Translation) project}}. In \bibinfo{booktitle}{\emph{Utrecht, The
  Netherlands: private publication}}. \bibinfo{pages}{1--11}.
\newblock


\bibitem[\protect\citeauthoryear{Wu and Wang}{Wu and Wang}{2007}]%
        {wu2007pivot}
\bibfield{author}{\bibinfo{person}{Hua Wu} {and} \bibinfo{person}{Haifeng
  Wang}.} \bibinfo{year}{2007}\natexlab{}.
\newblock \showarticletitle{{Pivot language approach for phrase-based
  statistical machine translation}}.
\newblock \bibinfo{journal}{\emph{Machine Translation}} \bibinfo{volume}{21},
  \bibinfo{number}{3} (\bibinfo{year}{2007}), \bibinfo{pages}{165--181}.
\newblock


\bibitem[\protect\citeauthoryear{Wu and Wang}{Wu and Wang}{2009}]%
        {wuandwang2009}
\bibfield{author}{\bibinfo{person}{Hua Wu} {and} \bibinfo{person}{Haifeng
  Wang}.} \bibinfo{year}{2009}\natexlab{}.
\newblock \showarticletitle{Revisiting Pivot Language Approach for Machine
  Translation}. In \bibinfo{booktitle}{\emph{Proceedings of the Joint
  Conference of the 47th Annual Meeting of the {ACL} and the 4th International
  Joint Conference on Natural Language Processing of the {AFNLP}}}.
  \bibinfo{publisher}{Association for Computational Linguistics},
  \bibinfo{address}{Suntec, Singapore}, \bibinfo{pages}{154--162}.
\newblock
\urldef\tempurl%
\url{https://www.aclweb.org/anthology/P09-1018}
\showURL{%
\tempurl}


\bibitem[\protect\citeauthoryear{Wu, Schuster, Chen, Le, Norouzi, Macherey,
  Krikun, Cao, Gao, Macherey, Klingner, Shah, Johnson, Liu, Kaiser, Gouws,
  Kato, Kudo, Kazawa, Stevens, Kurian, Patil, Wang, Young, Smith, Riesa,
  Rudnick, Vinyals, Corrado, Hughes, and Dean}{Wu et~al\mbox{.}}{2016}]%
        {DBLP:journals/corr/WuSCLNMKCGMKSJL16}
\bibfield{author}{\bibinfo{person}{Yonghui Wu}, \bibinfo{person}{Mike
  Schuster}, \bibinfo{person}{Zhifeng Chen}, \bibinfo{person}{Quoc~V. Le},
  \bibinfo{person}{Mohammad Norouzi}, \bibinfo{person}{Wolfgang Macherey},
  \bibinfo{person}{Maxim Krikun}, \bibinfo{person}{Yuan Cao},
  \bibinfo{person}{Qin Gao}, \bibinfo{person}{Klaus Macherey},
  \bibinfo{person}{Jeff Klingner}, \bibinfo{person}{Apurva Shah},
  \bibinfo{person}{Melvin Johnson}, \bibinfo{person}{Xiaobing Liu},
  \bibinfo{person}{Lukasz Kaiser}, \bibinfo{person}{Stephan Gouws},
  \bibinfo{person}{Yoshikiyo Kato}, \bibinfo{person}{Taku Kudo},
  \bibinfo{person}{Hideto Kazawa}, \bibinfo{person}{Keith Stevens},
  \bibinfo{person}{George Kurian}, \bibinfo{person}{Nishant Patil},
  \bibinfo{person}{Wei Wang}, \bibinfo{person}{Cliff Young},
  \bibinfo{person}{Jason Smith}, \bibinfo{person}{Jason Riesa},
  \bibinfo{person}{Alex Rudnick}, \bibinfo{person}{Oriol Vinyals},
  \bibinfo{person}{Greg Corrado}, \bibinfo{person}{Macduff Hughes}, {and}
  \bibinfo{person}{Jeffrey Dean}.} \bibinfo{year}{2016}\natexlab{}.
\newblock \showarticletitle{Google's Neural Machine Translation System:
  Bridging the Gap between Human and Machine Translation}.
\newblock \bibinfo{journal}{\emph{CoRR}}  \bibinfo{volume}{abs/1609.08144}
  (\bibinfo{year}{2016}).
\newblock
\urldef\tempurl%
\url{http://arxiv.org/abs/1609.08144}
\showURL{%
\tempurl}


\bibitem[\protect\citeauthoryear{Zaremoodi, Buntine, and Haffari}{Zaremoodi
  et~al\mbox{.}}{2018}]%
        {Zaremoodi-ACL2018}
\bibfield{author}{\bibinfo{person}{Poorya Zaremoodi}, \bibinfo{person}{Wray
  Buntine}, {and} \bibinfo{person}{Gholamreza Haffari}.}
  \bibinfo{year}{2018}\natexlab{}.
\newblock \showarticletitle{Adaptive Knowledge Sharing in Multi-Task Learning:
  Improving Low-Resource Neural Machine Translation}. In
  \bibinfo{booktitle}{\emph{Proceedings of the 56th Annual Meeting of the
  Association for Computational Linguistics (Volume 2: Short Papers)}}.
  \bibinfo{publisher}{Association for Computational Linguistics},
  \bibinfo{address}{Melbourne, Australia}, \bibinfo{pages}{656--661}.
\newblock
\urldef\tempurl%
\url{http://aclweb.org/anthology/P18-2104}
\showURL{%
\tempurl}


\bibitem[\protect\citeauthoryear{Zhou, Hu, Zhang, and Zong}{Zhou
  et~al\mbox{.}}{2017}]%
        {zhou-etal-2017-neural}
\bibfield{author}{\bibinfo{person}{Long Zhou}, \bibinfo{person}{Wenpeng Hu},
  \bibinfo{person}{Jiajun Zhang}, {and} \bibinfo{person}{Chengqing Zong}.}
  \bibinfo{year}{2017}\natexlab{}.
\newblock \showarticletitle{Neural System Combination for Machine Translation}.
  In \bibinfo{booktitle}{\emph{Proceedings of the 55th Annual Meeting of the
  Association for Computational Linguistics (Volume 2: Short Papers)}}.
  \bibinfo{publisher}{Association for Computational Linguistics},
  \bibinfo{address}{Vancouver, Canada}, \bibinfo{pages}{378--384}.
\newblock
\urldef\tempurl%
\url{https://doi.org/10.18653/v1/P17-2060}
\showDOI{\tempurl}


\bibitem[\protect\citeauthoryear{Ziemski, Junczys-Dowmunt, and
  Pouliquen}{Ziemski et~al\mbox{.}}{2016a}]%
        {ZIEMSKI16.1195}
\bibfield{author}{\bibinfo{person}{Michał Ziemski}, \bibinfo{person}{Marcin
  Junczys-Dowmunt}, {and} \bibinfo{person}{Bruno Pouliquen}.}
  \bibinfo{year}{2016}\natexlab{a}.
\newblock \showarticletitle{{The United Nations Parallel Corpus v1.0}}. In
  \bibinfo{booktitle}{\emph{Proceedings of the Tenth International Conference
  on Language Resources and Evaluation (LREC 2016)}} (23-28).
  \bibinfo{publisher}{European Language Resources Association (ELRA)},
  \bibinfo{address}{Portorož, Slovenia}.
\newblock
\showISBNx{978-2-9517408-9-1}


\bibitem[\protect\citeauthoryear{Ziemski, Junczys-Dowmunt, and
  Pouliquen}{Ziemski et~al\mbox{.}}{2016b}]%
        {ziemski2016united}
\bibfield{author}{\bibinfo{person}{Micha{\l} Ziemski}, \bibinfo{person}{Marcin
  Junczys-Dowmunt}, {and} \bibinfo{person}{Bruno Pouliquen}.}
  \bibinfo{year}{2016}\natexlab{b}.
\newblock \showarticletitle{The United Nations Parallel Corpus v1.0}. In
  \bibinfo{booktitle}{\emph{Proceedings of the Tenth International Conference
  on Language Resources and Evaluation ({LREC} 2016)}}.
  \bibinfo{publisher}{European Language Resources Association (ELRA)},
  \bibinfo{address}{Portoro{\v{z}}, Slovenia}, \bibinfo{pages}{3530--3534}.
\newblock
\urldef\tempurl%
\url{https://www.aclweb.org/anthology/L16-1561}
\showURL{%
\tempurl}


\bibitem[\protect\citeauthoryear{Zoph and Knight}{Zoph and Knight}{2016}]%
        {N16-1004}
\bibfield{author}{\bibinfo{person}{Barret Zoph} {and} \bibinfo{person}{Kevin
  Knight}.} \bibinfo{year}{2016}\natexlab{}.
\newblock \showarticletitle{Multi-Source Neural Translation}. In
  \bibinfo{booktitle}{\emph{Proceedings of the 2016 Conference of the North
  American Chapter of the Association for Computational Linguistics: Human
  Language Technologies}}. \bibinfo{publisher}{Association for Computational
  Linguistics}, \bibinfo{address}{San Diego, California},
  \bibinfo{pages}{30--34}.
\newblock
\urldef\tempurl%
\url{https://doi.org/10.18653/v1/N16-1004}
\showDOI{\tempurl}


\bibitem[\protect\citeauthoryear{Zoph and Le}{Zoph and Le}{2017}]%
        {45826}
\bibfield{author}{\bibinfo{person}{Barret Zoph} {and} \bibinfo{person}{Quoc~V.
  Le}.} \bibinfo{year}{2017}\natexlab{}.
\newblock \showarticletitle{Neural Architecture Search with Reinforcement
  Learning}.
\newblock
\urldef\tempurl%
\url{https://arxiv.org/abs/1611.01578}
\showURL{%
\tempurl}


\bibitem[\protect\citeauthoryear{Zoph, Yuret, May, and Knight}{Zoph
  et~al\mbox{.}}{2016}]%
        {DBLP:conf/emnlp/ZophYMK16:original}
\bibfield{author}{\bibinfo{person}{Barret Zoph}, \bibinfo{person}{Deniz Yuret},
  \bibinfo{person}{Jonathan May}, {and} \bibinfo{person}{Kevin Knight}.}
  \bibinfo{year}{2016}\natexlab{}.
\newblock \showarticletitle{Transfer Learning for Low-Resource Neural Machine
  Translation}. In \bibinfo{booktitle}{\emph{Proceedings of the 2016 Conference
  on Empirical Methods in Natural Language Processing}}.
  \bibinfo{publisher}{Association for Computational Linguistics},
  \bibinfo{address}{Austin, Texas}, \bibinfo{pages}{1568--1575}.
\newblock
\urldef\tempurl%
\url{https://doi.org/10.18653/v1/D16-1163}
\showDOI{\tempurl}


\end{thebibliography}

\appendix

\end{document}